\pdfoutput=1 

\documentclass{egpubl}
\usepackage{pg2020}
 
\SpecialIssuePaper         

\usepackage[T1]{fontenc}
\usepackage{dfadobe}  

\usepackage{cite}  
\BibtexOrBiblatex
\electronicVersion
\PrintedOrElectronic

\ifpdf \usepackage[pdftex]{graphicx} \pdfcompresslevel=9
\else \usepackage[dvips]{graphicx} \fi

\usepackage{egweblnk} 
\usepackage{booktabs}
\usepackage{colortbl}
\usepackage{amssymb}

\definecolor{mygray}{gray}{.9}
\definecolor{mypink}{rgb}{.99,.91,.95}
\definecolor{mycyan}{cmyk}{.3,0,0,0}
\definecolor{myblue}{rgb}{0,0,0.9}

\definecolor{tabcolor}{rgb}{.5,.5,.5}

\title[Multi-scale Information Assembly for Image Matting]%
{Multi-scale Information Assembly for Image Matting}

\author[Qiao \& Liu \& Zhu et al.]
{\parbox{\textwidth}{\centering Yu Qiao\footnotemark[1], Yuhao Liu\footnotemark[1], Qiang Zhu, Xin Yang\footnotemark[2], Yuxin Wang, Qiang Zhang, and Xiaopeng Wei \\
	}
	\\
	{\parbox{\textwidth}{\centering College of Computer Science, Dalian University of Technology
		}
	}
}

\begin{document}
	
	\teaser{
		\setlength{\tabcolsep}{1pt}\small{
			\begin{tabular}{cccc}
				\includegraphics[scale=0.06665]{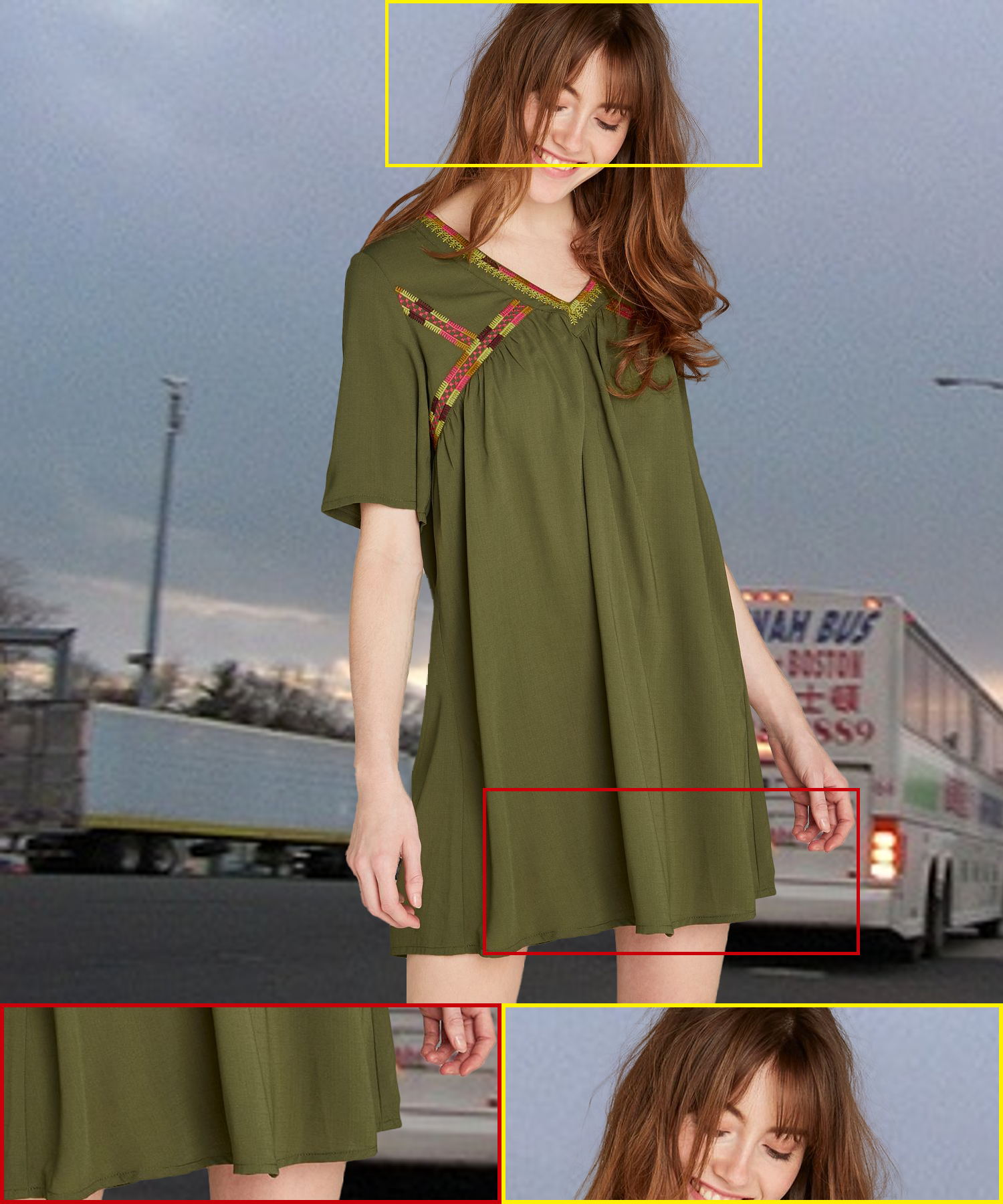} &
				\includegraphics[scale=0.06665]{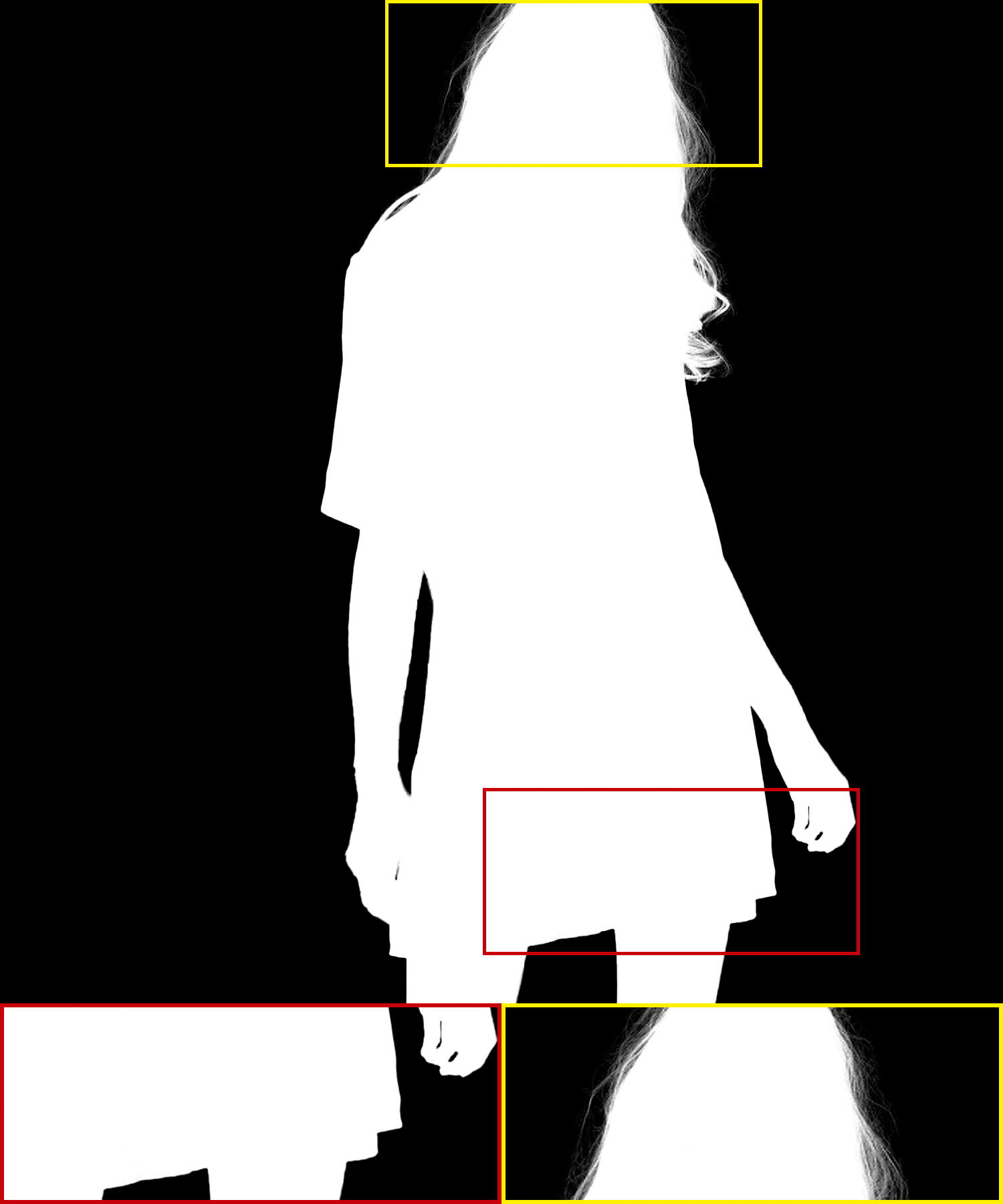} &
				\includegraphics[scale=0.25407]{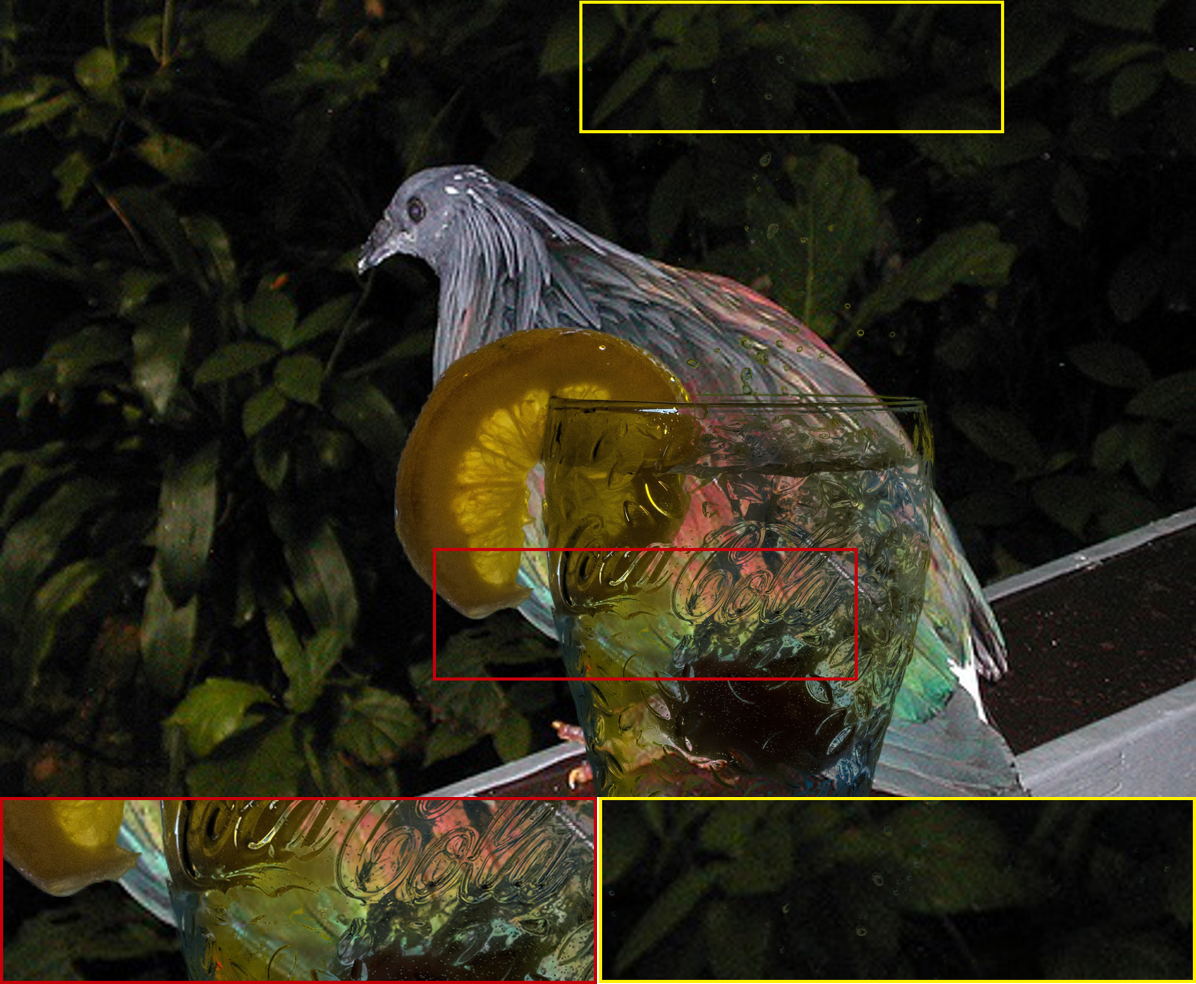} &
				\includegraphics[scale=0.273224]{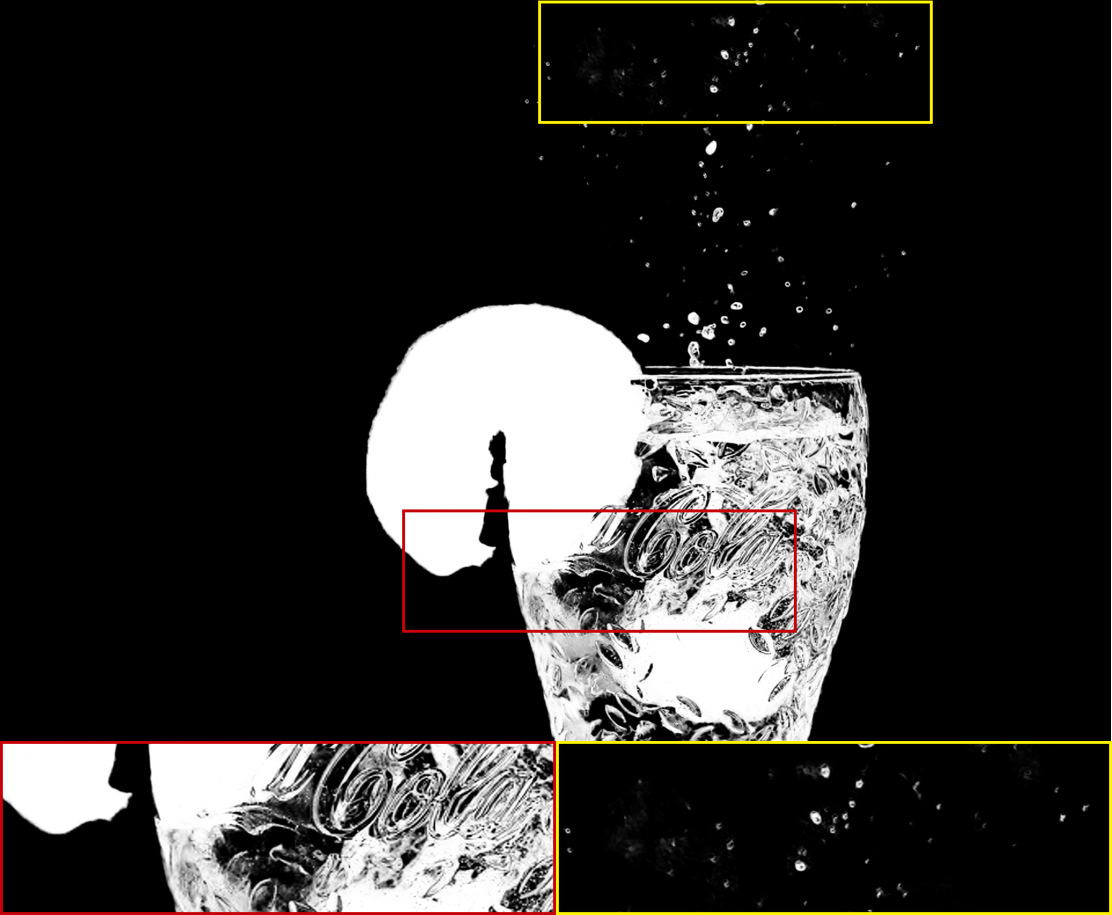} \\
				
				Input Image & Alpha Matte & Input Image & Alpha Matte \\
		\end{tabular}}
		\caption{The alpha mattes generated by our proposed MSIA-matte. Some details are zoomed in and placed under the images. }
		\label{fig:teaser}
	}
	
	\maketitle
	\begin{abstract}
		Image matting is a long-standing problem in computer graphics and vision, mostly identified as the accurate estimation of the foreground in input images. We argue that the foreground objects can be represented by different-level information, including the central bodies, large-grained boundaries, refined details, etc. Based on this observation, in this paper, we propose a multi-scale information assembly framework (MSIA-matte) to pull out high-quality alpha mattes from single RGB images. Technically speaking, given an input image, we extract advanced semantics as our subject content and retain initial CNN features to encode different-level foreground expression, then combine them by our well-designed information assembly strategy. Extensive experiments can prove the effectiveness of the proposed MSIA-matte, and we can achieve state-of-the-art performance compared to most existing matting networks.

		\ccsdesc[500]{Computing methodologies~Image segmentation}
		\ccsdesc[500]{Computing methodologies~Image representations}

		\printccsdesc   
	\end{abstract}  
	
	\footnotetext[1]{Joint first authors. $^{\ddagger}$Corresponding author and he led this project. }
	
	\section{Introduction}
	\label{sec:introduction}
	
	Natural images are composed of different kinds of objects, and we usually consider the regions of interest as foreground. Accurate prediction and separation of the foreground in the input image (noted as image matting) is a significant problem in industry and academia. Image matting has a wide range of applications, film production, live streaming, online image editing, etc. We can define image matting from the perspective of image synthesis as follows:
	\begin{equation}
	\label{eq:composition}
	I_{i}=\alpha_{i}F_{i}+(1-\alpha_{i})B_{i},
	\end{equation}
	where $I$ represents the input image, $i$ refers to the pixel location of $I$, $F$ and $B$ refer to the foreground and background layers separately. $\alpha$ denotes the alpha matte, where $\alpha_i$ varies in the range of [0,1] to suggest the opacity of the foreground. We can summarize from Equation~(\ref{eq:composition}) that a relatively complete foreground object can be represented by different-level information. They could be subject content ($\alpha_{i} = 1$), clear borders or coarse textures ($\alpha_{i}$ are very large, even close to 1), semi-transparent areas or fine-grained details ($\alpha_{i}$ are relatively small).
	
	Given only an input image, solving for other variables from Equation~(\ref{eq:composition}) can be intractable. Therefore, most existing matting methods import trimaps to confine the foreground and background. The trimap consists of three categories to indicate the foreground ($\alpha = 1$), background ($\alpha = 0$), and transition region ($\alpha \in (0,1)$). With explicit foreground and background as the reference, traditional matting methods resort to sampling or affinity to estimate the alpha values in the transition region. However, they mostly consider color distribution to predict alpha mattes and ignore the semantic information of the foreground object, which can result in poor performance when the foreground and background share similar colors.
	
	Deep learning has contributed a lot to the development of computer vision, as well as image matting. \cite{Cho2016Natural} et al. combined RGB images with the alpha mattes from~\cite{Levin2007A} and~\cite{Cho2016Natural} to predict refined results via deep convolutional neural networks. The concatenation of RGB images and trimaps as input was first introduced by~\cite{Xu2017Deep}, and the expression ability of neural networks for the foreground structure has been fully demonstrated. Many subsequent matting networks~\cite{hou2019context, cai2019disentangled} follow this principle to optimize their results, using trimaps to confine the input images and designing comprehensive network architectures to present foreground information. However, generating a well-defined trimap from a natural image is fussy and difficult for novice users, limiting the application of the above methods in practice to some extent.
	
	In recent years, some matting methods can also achieve alpha mattes without trimaps, and their motivations are mainly relying on semantic segmentation or saliency detection. \cite{Zhang2019CVPR} et al. proposed a fusion network to blend the foreground and background weight maps, both of which are derived from a segmentation network. The attention mechanism proposed in~\cite{Qiao_2020_CVPR} is more of a saliency-dominated model: the advanced semantics from the backbone network are applied to the improvement of the low-level features. Almost all trimap-free methods involve low-level features in their later fusion or refinement stage, which is the main reason that they can restore fine-grained boundary details in alpha mattes. However, many methods exploit low-level CNN features for image matting without considering their different-opacity foreground information.
	
	In this paper, we argue that the foreground expression from low-level features still possesses different-level information. For a natural image, there is usually an explicit subject content of foreground object (it could be a human, a dog, or a glass), and there are also diverse boundaries or details (such as hands, hairs, branches, leaves), denoted as superficial traces in this paper. These superficial traces obviously require different alpha values indicated by Equation~(\ref{eq:composition}). Therefore, we propose a multi-scale information assembly network (\emph{MSIA-matte}) to extract different-level foreground information for predicting alpha mattes. Specifically, we harness a backbone network to capture high-level semantics and design a superficial traces branch to extract different-level initial features. In the decoder part, we assemble multi-scale features by our proposed information assembly module to estimate alpha mattes. The feature maps from the superficial traces branch can complement the advanced semantics in multi-layered textures and details, and their integration can guarantee relatively complete and concrete foreground expression. Extensive experiments can prove the effectiveness of our proposed network, and we can achieve state-of-the-art performance compared to the existing matting methods with/without trimaps.
	
	We list our contributions as follow:
	\begin{itemize}
		\item We propose a multi-scale information assembly network (MSIA-matte) to predict alpha mattes, which can combine different-level foreground expression to refine the textures and details in results.
		\item We design a novel extraction strategy, which can adapt to various opacity in the foreground and provide high-level semantics and superficial traces for information assembly.
		\item We conduct extensive experiments to validate our model, and the alpha mattes produced by our network can achieve state-of-the-art performance.
	\end{itemize}
	
	\section{Related Work}
	\label{sec:related_work}
	
	In this paper, we propose a multi-scale integration model to generate alpha mattes without trimaps. In this section, we briefly review some matting approaches from three directions: traditional matting methods, trimap-based deep-learning networks, and trimap-free matting architectures.
	
	\textbf{Traditional approaches.} Most traditional approaches resort to scribbles or trimaps to improve their sampling or propagation process. The trimap is generally a gray-scale image with three colors: white, black, and gray. The white indicates the foreground in the corresponding input RGB image, and the black and gray can suggest the background and transition region, respectively. Most of the foreground and background are recommended by trimap, and the core of the matting is to solve the transition region according to explicit parts. The theory of scribbles is similar, using simple scribbles to provide some explicit foreground and background information, which is user-friendly, while the produced results are inferior to trimap-based methods due to the limited available reference regions. According to the different ways of exploring explicit information, traditional methods can be divided into two directions: sampling-based methods and affinity-based methods.
	
	Sampling methods~\cite{Chuang2003A,Feng2016A,Karacan2015Image,Rhemann2011A,Shahrian2013Improving,Wang2007Optimized} collect sampling pixels from explicit parts and predict the pixels in transition regions by sampled ones. Affinity methods~\cite{Aksoy2017Designing,Chen2013KNN,Grady2005Random,Lee2011Nonlocal,Levin2007A,Levin2008Spectral,Sun2004Poisson} propagate explicit pixels to the transition region to estimate the alpha values of the whole image, which is more robust than sampling methods. However, the above methods mostly exploit color or texture distribution to achieve the regression of transition regions, ignoring the high-level semantics of the foreground, which can represent the structure of the input images~\cite{Xu2017Deep}.
	
	\begin{figure*}[t]
		\centering
		\includegraphics[scale=.36]{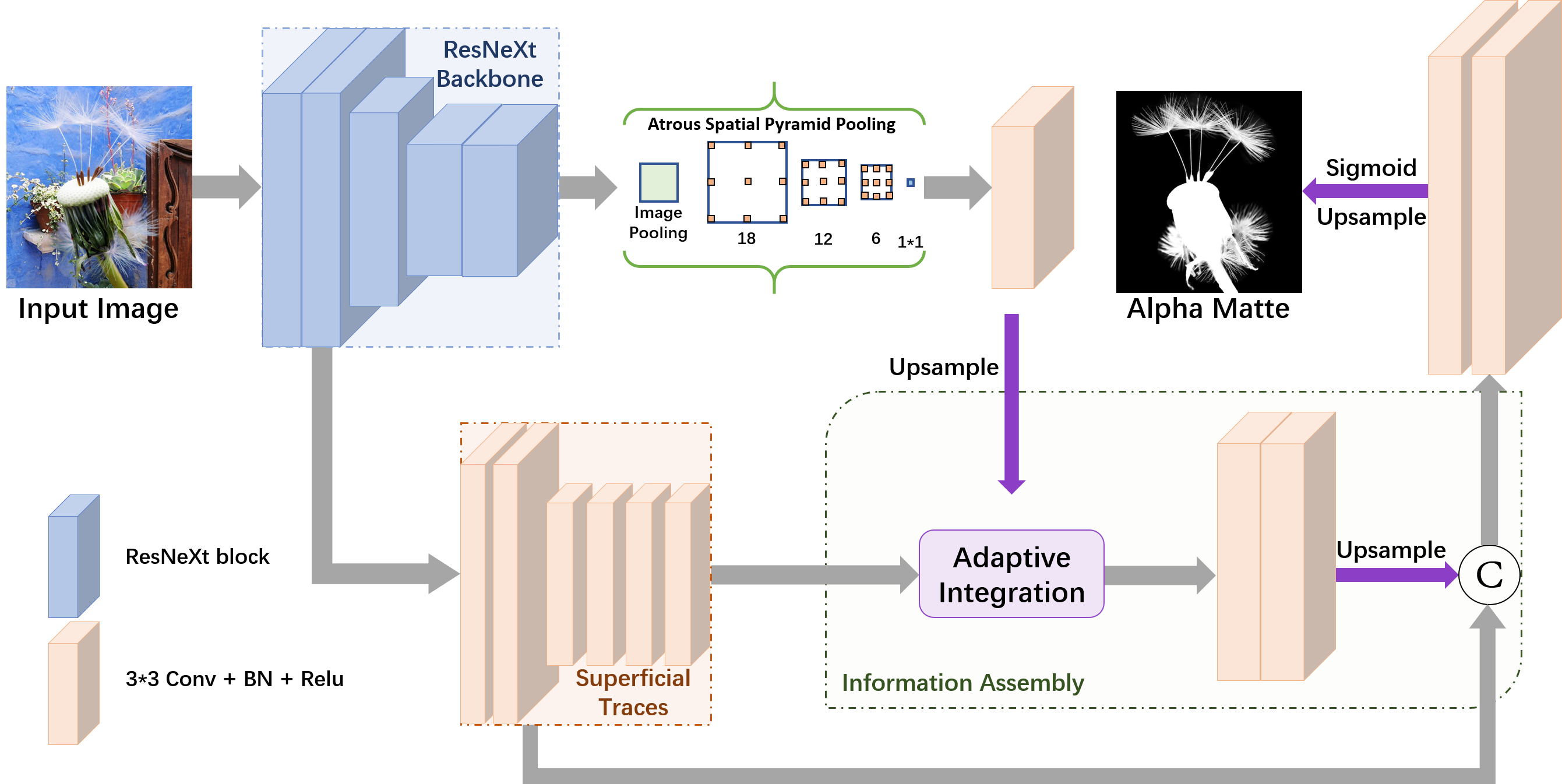}
		\caption{The pipeline of the proposed MSIA-matte. We employ ResNeXt as our backbone to extract advanced semantics, which can represent the subject content of the foreground. Synchronously, we preserve initial features to infer different-level superficial traces, which contain foreground details and textures with distinguished opacity. In the information assembly decoder stage, we employ an adaptive strategy to assemble multi-scale foreground expression to predict an alpha matte. }
		\label{fig:pipeline}
	\end{figure*}
	
	\textbf{Deep-learning matting with trimaps.} As pixel-wise dense annotations, trimaps can provide sufficient guidance for alpha perception. Therefore, many deep-learning-based matting methods utilize trimaps as additional inputs to improve the final results. The deep architecture in Cho et al.~\cite{Cho2016Natural} combined RGB images with the results from~\cite{Levin2007A} and~\cite{Chen2013KNN} as input and produce better alpha mattes. Xu et al.~\cite{Xu2017Deep} proposed deep image matting (DIM), designing an encoder-decoder network to extract advanced semantics from RGB images and trimaps, and the importance and quality of the trimaps for DIM have been proved in~\cite{Zhang2019CVPR}. Lutz et al.~\cite{lutz2018alphagan} adopted generative adversarial networks~\cite{Goodfellow2014Generative} to enhance the matte prediction, which demonstrated the ability of discriminator for pixel-wise visual quality, and the later~\cite{Qiao_2020_CVPR} also imported a discriminator to strengthen the visual effects. Sampling operation was integrated with neural networks in Tang et al.~\cite{Tang_2019_CVPR} for foreground opacity estimation. Cai et al.~\cite{cai2019disentangled} and Hou et al.~\cite{hou2019context} both resorted to another branch to improve the alpha prediction, which exhibited the joint promotion of two branches. Hao et al.~\cite{hao2019indexnet} embedded the index function into their encoder-decoder network to preserve more effective parameters. Li et al.~\cite{li2020natural} utilized guided contextual attention to improve the transmission of the opacity information. However, the above methods require carefully defined trimaps as assistant input, which partly restrict their practical values in real life.
	
	\textbf{Trimap-free matting networks.} This solution is mostly the production of semantic segmentation or saliency detection, using intermediate results as the reference to confine the final alpha mattes. Some matting frameworks~\cite{Chen2018SHM, Cho2016Automatic} employed segmentation models to produce coarse trimaps, and the results based on such trimaps can usually lose information in the foreground objects or edges. The reinforcement model proposed in~\cite{Yang2018Active} can promote alpha mattes progressively, while their feedback stage will trouble users a lot. The computation and time-consuming in~\cite{sss} were very expensive, although they can produce comparable alpha mattes. Zhang et al.~\cite{Zhang2019CVPR} used a segmentation variant to represent the foreground and background of the input image, while subtle segmentation errors can discount their results. The attention mechanism in~\cite{Qiao_2020_CVPR} can effectively aggregate the pyramidal features and appearance cues, while they use high-level semantics as primary guidance for attention, more like a saliency-dominated matting. In this paper, we investigate different-level superficial traces to complement the semantic extraction of the backbone, and the novel information assembly strategy can integrate multi-scale features effectively.
	
	\section{Methodology}
	\label{sec:methodology}
	
	In this section, we formally interpret our multi-scale information assembly network (MISA-matte). For a familiar natural image, there may be different kinds of objects, and we usually define the one of interest as the foreground. Existing matting methods mostly exploit complex models to extract the foreground semantics and adopt later refinement or fusion~\cite{Xu2017Deep, cai2019disentangled, Zhang2019CVPR} after the decoder to restore the details and edges in alpha mattes. However, there is various opacity in the foreground (like the hair, body, and hands in a portrait, the ears, legs, and hairs of a cat), and such relatively direct refinement or fusion is inapplicable to integrate different-level foreground expression, which can lead to some information absence when the opacity of the foreground or background shares a continuous variation. The motivation of our model is to capture multi-scale foreground information to represent manifold opacity, then integrate them to predict a relatively complete foreground.
	
	\subsection{Overview}
	\label{ssec:overview}
	
	The pipeline of the proposed MSIA-matte is unfolded in Figure~\ref{fig:pipeline}. We adopt ResNeXt~\cite{Xie2017Aggregated} as our backbone to extract advanced semantics, which can represent the subject content of the foreground objects. Then we use the Atrous Spatial Pyramidal Pooling module~\cite{Chen2018DeepLab} to further capture different-level semantic information. Similar to~\cite{Zhang2019CVPR,hou2019context,Qiao_2020_CVPR}, we first intend to combine low-level CNN features in the decoder parts. However, based on different-level opacity observations, we establish a superficial traces branch to encode low-level CNN features. The original textures, borders, or details can be converted to different-level foreground expression. Then we can integrate multi-scale expression by well-designed information assembly module, including upsampling, convolution, and concatenation operation in the decoder stage. The whole network architecture is optimized by a blended loss function.
	
	\subsection{Superficial traces branch}
	\label{ssec:st_branch}
	
	High-level semantic feature extraction has been widely developed by convolutional neural networks~\cite{simonyan2014deep, Khe2016Resnet}, and such multi-layer semantics can be combined with upsampled features in the decoder stage~\cite{JLong2015fcnn}. Inspired by this multi-scale semantics fusion strategy, we propose our superficial traces branch to further analyze low-level CNN features from ResNeXt block1. Low-level CNN features contain many complex image textures, which can be integrated into alpha perception to complement boundary details or semi-transparent regions~\cite{Zhang2019CVPR,Qiao_2020_CVPR}. The motivation of our superficial branch is to extract different-level foreground traces. Thus we can restore various opacity in final alpha mattes.
	
	We perform a feature transform module on the low-level features to extract superficial traces. Specifically, we first employ two convolutional layers with $3*3$ kernel size and $64$ channels to process initial low-level features (the output features here are denoted as $\mathcal F_{ini}$). This operation can preliminarily filter scrambled image textures and preserve sufficient fine-grained foreground information. After this, we use a downsampled convolution on $\mathcal F_{ini}$ with stride=2, then three cascaded convolutional layers with $3*3$ kernel size and $256$ channels are used to extract relatively deep superficial traces, recorded as $\mathcal F_{sed}$. We utilize such continuous convolutions to condense some coarse-grained borders or textures of the foreground, which belong to superficial traces and are defined in low-level CNN features but contain subtle semantics or clear transition (such as the hands, leaves, claws). Such filtration and concentration operations on superficial traces can mostly retain initial image textures and capture different-level foreground information, and their integration with high-level semantics can jointly regress alpha values.
	
	\subsection{Information assembly decoder}
	\label{ssec:ia_decoder}
	
	We assemble different-level superficial traces with high-level semantics during our upsampling integration stage to restore the foreground profiles in alpha mattes. According to our previous analysis, the secondary superficial traces $\mathcal F_{sed}$ from 4 cascaded convolutional layers contain foreground expression above the initial image textures. Nevertheless, such expression may be inapplicable for all kinds of matting images: the input image must include boundary details, but some examples may contain no middle-level superficial traces (hands, semi-transparent regions). Therefore, we first involve an information assembly to dynamically integrate secondary superficial traces $\mathcal F_{sed}$ with semantic features $\mathcal F_{aspp}$ from Atrous Spatial Pyramidal Pooling (ASPP) module. The assembled foreground information $\mathcal F_{IA}$ is encoded by two $3*3$ convolutional layers and upsampled by a factor of $2$, then integrated with preliminarily processed low-level features $\mathcal F_{ini}$ by a concatenation operation, the output of the concatenation is denoted as $\mathcal F_{cat}$. The final alpha mattes can be generated from the concatenated features $\mathcal F_{cat}$ through two $3*3$ convolutional layers, a sigmoid activation function, and the upsampling operation. The detailed upsampling process and convolutional layers in the decoder stage are all demonstrated in Figure~\ref{fig:pipeline}.
	
	Our information assembly module and concatenation operation can be formally defined as follows:
	\begin{equation}
	\label{eq:weight1_sum}
	\mathcal W_{aspp}=\mathcal W_{aspp}/(\mathcal W_{aspp}+\mathcal W_{sed})+\epsilon,
	\end{equation}
	\begin{equation}
	\label{eq:weight2_sum}
	\mathcal W_{sed}=\mathcal W_{sed}/(\mathcal W_{aspp}+\mathcal W_{sed})+\epsilon,
	\end{equation}
	\begin{equation}
	\label{eq:weight_IA}
	\mathcal F_{IA}=\mathcal{C}at(\mathcal W_{aspp}*\mathcal F_{aspp}, \mathcal W_{sed}*\mathcal F_{sed}),
	\end{equation}
	\begin{equation}
	\label{eq:weight_cat}
	\mathcal F_{cat} = \mathcal{C}at[\mathcal{U}p(\mathcal{C}onv(\mathcal{C}onv(\mathcal F_{IA}))), \mathcal F_{ini})],
	\end{equation}
	where $\mathcal W_{aspp}$ and $\mathcal W_{sed}$ represent the assembly coefficient scalars which can balance dynamic weights between the ASPP features and the secondary superficial traces, the parameter $\epsilon$ is to prevent the denominator from being zero. The feature maps $\mathcal F_{aspp}$ and $\mathcal F_{sed}$ can be adaptively integrated with this dynamic weights adjustment mode. The final concatenated features $\mathcal W_{cat}$ can be obtained by Equation~(\ref{eq:weight_cat}), and here the $\mathcal{U}p$ and $\mathcal{C}onv$ denote the upsampling and convolution operation, respectively.
	
	Such an information assembly strategy can effectively integrate multi-scale foreground expression to achieve high-quality alpha mattes. The high-level features from the ASPP module can produce foreground semantics, suggesting the subject content of the foreground. In contrast, the superficial traces from low-level CNN features contain different-level foreground details or textures which may share various opacity about the foreground boundaries. Their assembly can restore foreground information through dynamically adapted features integration. We conduct some experiments to evaluate the proposed MSIA-matte model, and the relevant ablation study can also prove the effectiveness of the proposed superficial traces branch and information assembly strategy.
	
	\subsection{Loss Function}
	\label{ssec:loss}
	
	We use a blended loss function to optimize our MSIA-matte network model, which combines $\mathcal L_{1}$ and $\mathcal L_{SSIM}$ to supervise the network training. The $\mathcal L_{1}$ is defined as follows:
	\begin{equation}
	\label{eq:L1_loss}
	\mathcal L_{1}=\sum^{\Omega}_{k}|\alpha^{k}_{m}-\alpha^{k}_{t}|,\quad\alpha^{k}_{m},\alpha^{k}_{t}\in[0,1],
	\end{equation}
	where $\Omega$ represents pixels set, and the value $k$ denotes the pixel index. $\alpha^{k}_{m}$ is the alpha value of pixel $k$ in the regressed alpha matte, while $\alpha^{k}_{t}$ represents the corresponding ground-truth value in the same pixel location. Compared to $\mathcal L_{2}$, $\mathcal L_{1}$ is more robust for noticeable alpha value differences (the foreground is $1$, while the background is $0$). With $\mathcal L_{1}$ in the optimization, the model can pay more attention to absolute alpha values, which can promote the evacuation of different-level foreground information. The $\mathcal L_{SSIM}$ expression is listed as:
	\begin{equation}
	\label{eq:ssim_loss}
	\mathcal L_{SSIM}=1-\frac{(2\mu_{m}\mu_{t}+c_{1})(2\sigma_{mt}+c_{2})}{(\mu^{2}_{m}+\mu^{2}_{t}+c_{1})(\sigma^{2}_{m}+\sigma^{2}_{t}+c_{2})},
	\end{equation}
	where $\mu_{m}$, $\mu_{t}$, and $\sigma_m$, $\sigma_t$ are the mean and standard deviations of $\alpha^{k}_{m}$ and $\alpha^{k}_{t}$, according to~\cite{Zhou2004Image}. The effectiveness of $\mathcal L_{SSIM}$ for image matting has been proved in~\cite{Qiao_2020_CVPR}. The $\mathcal L_{SSIM}$ can improve the foreground integrity and enhance the correlation between local regions, while the $\mathcal L_{1}$ can provide pixel-wise accuracy. Thus our total loss function can be defined as:
	\begin{equation}
	\label{eq:total_loss}
	\mathcal L_{total}=\lambda_{1}\mathcal L_{1} + \lambda_{2}\mathcal L_{SSIM},
	\end{equation}
	where $\lambda_{1}$ and $\lambda_{2}$ are coefficients to balance these two losses. In the training procedure, $\lambda_{1}$ and $\lambda_{2}$ are set as 1 and 0.1 for the first epoch and modified to 1 and 0.025 in the following epochs.
	
	\section{Experiments}
	\label{sec:experiments}
	
	We conduct our experiments on two large-scale matting datasets: Composition-1K and Distinctions-646. The Composition-1K dataset is from~\cite{Xu2017Deep}, which contains $431$ available alpha mattes for training and $50$ alpha mattes for testing. The Distinctions-646, provided by~\cite{Qiao_2020_CVPR}, consists of $596$ training alpha mattes and $50$ testing ones. We follow the rules in~\cite{Xu2017Deep} to produce $100$ composite examples for each training alpha matte and $20$ examples for each alpha matte in the testing set. We train our MSIA-matte on both datasets and evaluate the test and analysis of two trained models. We also perform an ablation study to verify different modules and finally demonstrate our results on natural images.
	
	\subsection{Implementation and comparison details}
	\label{ssec:implementation}
	
	\textbf{Implementation environment and parameters.} We establish our MSIA-matte model by PyTorch and use $3$ Tesla P100 graphics to train and test the model parallelly. For training, we use the pre-trained ResNeXt-101 network~\cite{Xie2017Aggregated} to initialize the weights in our backbone and the other weights are assigned with a random Gaussian distribution. All input images during training are randomly cropped to 512 $\times$ 512, 640 $\times$ 640 or 800 $\times$ 800, then resized to a resolution of 512 $\times$ 512 and augmented by horizontal random flipping. We use the stochastic gradient descent (\emph{SGD}) optimizer with a momentum of 0.9 and a weight decay of 0.0005 to optimize our network parameters. The learning rate is initialized to $0.01$, adjusted by the "poly" policy~\cite{Wei2015ParseNet} with the power of $0.9$ for $20$ epochs. During training, the batch size is set as $4$, which requires $20$ hours to finish $20$ epochs, and the test on $1000$ images only needs several minutes.
	
	\textbf{Experiment metrics for evaluation.} For quantitative comparisons with the state-of-the-art methods, we refer to the four standard metrics in~\cite{rhemann2009perceptually}: the sum of absolute differences (SAD), mean square error (MSE), the gradient, and connectivity. To better display the visual quality, we zoom in some critical details of the alpha mattes and place them under the images to show the differences more concretely.
	
	\begin{figure*}[htbp]
		\arrayrulecolor{tabcolor}
		\centering
		\setlength{\tabcolsep}{1pt}\small{
			\begin{tabular}{c|c|c|c|c}	
				\includegraphics[scale=0.1588]{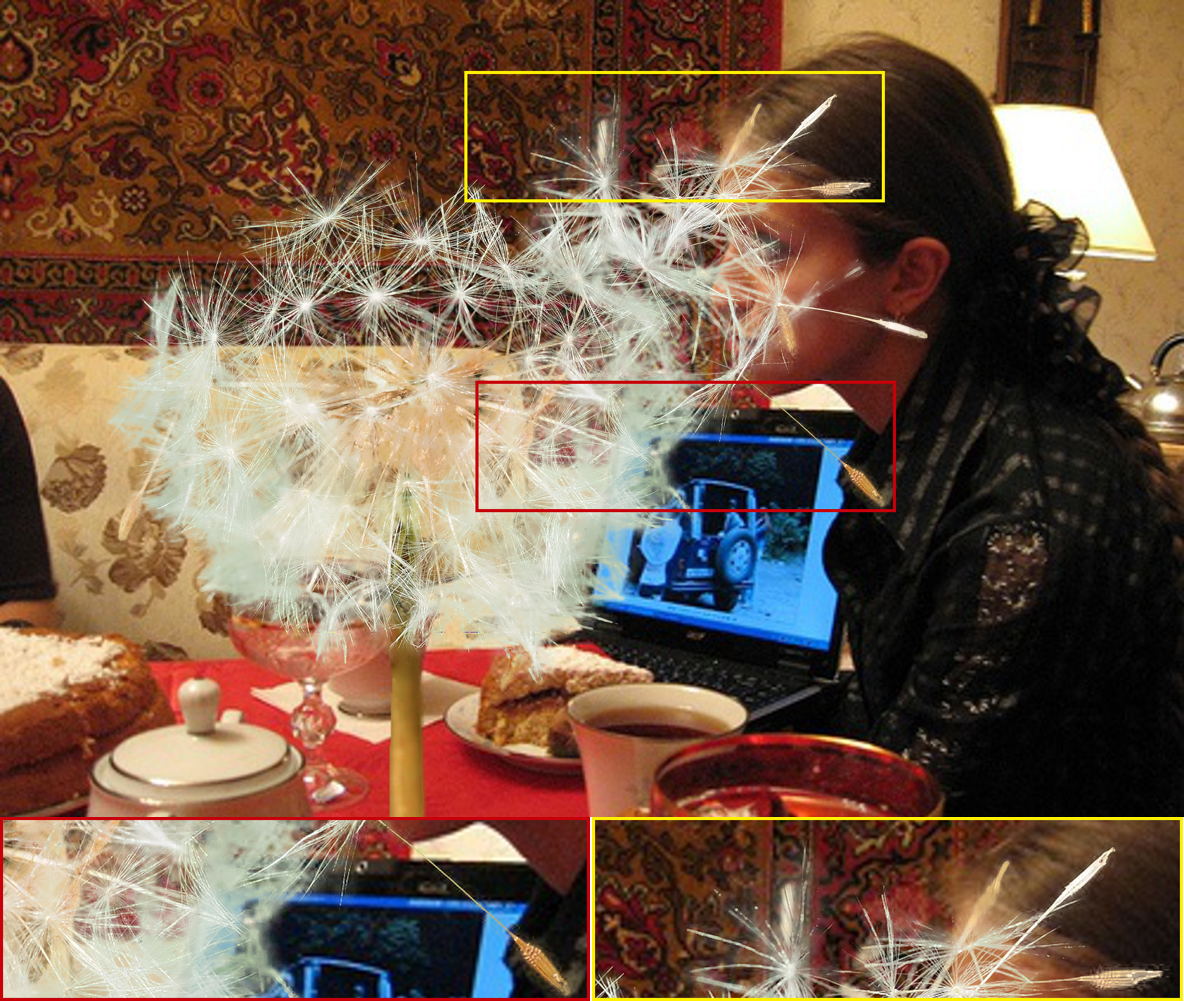} &
				\includegraphics[scale=0.04706]{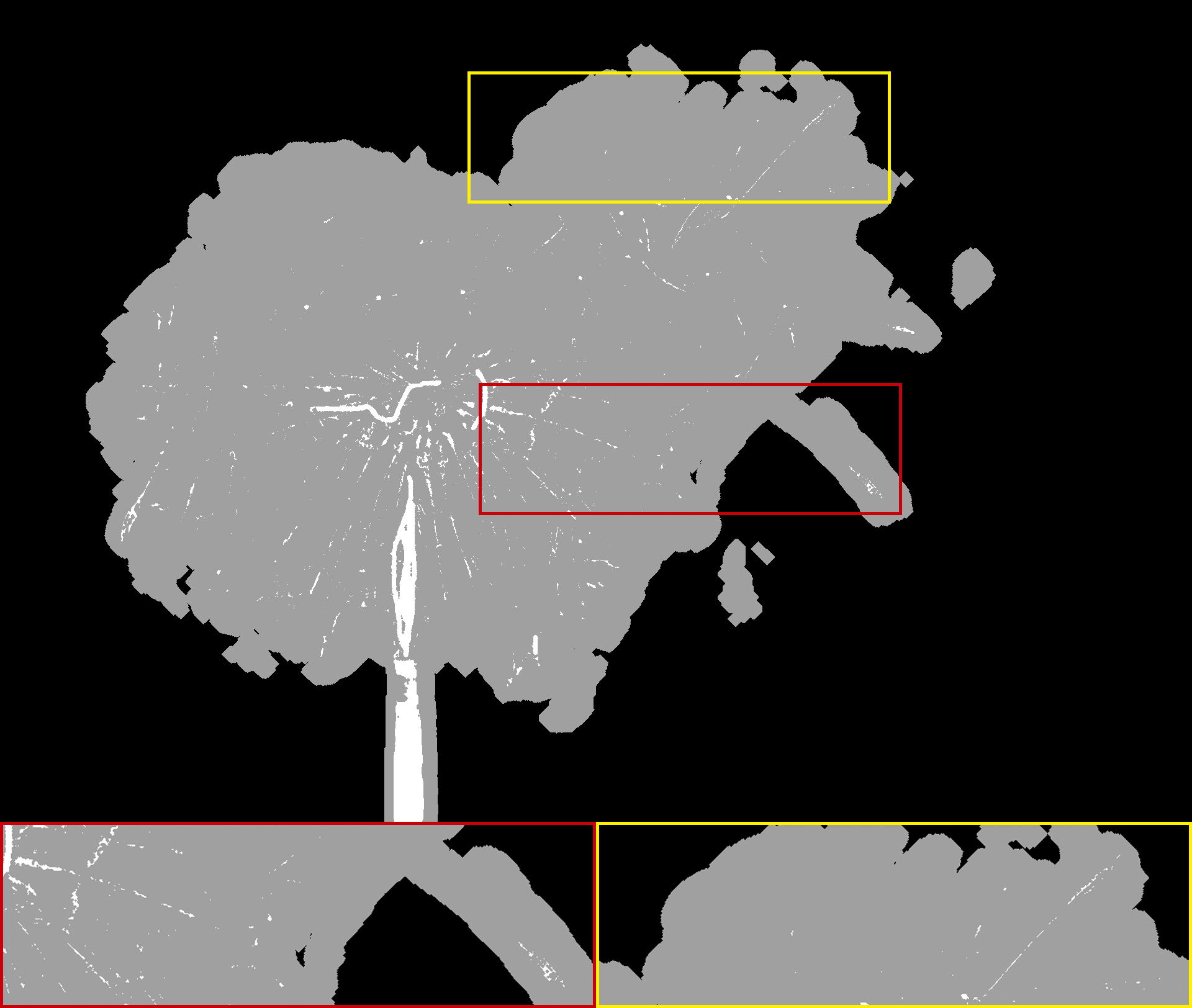} &
				\includegraphics[scale=0.158]{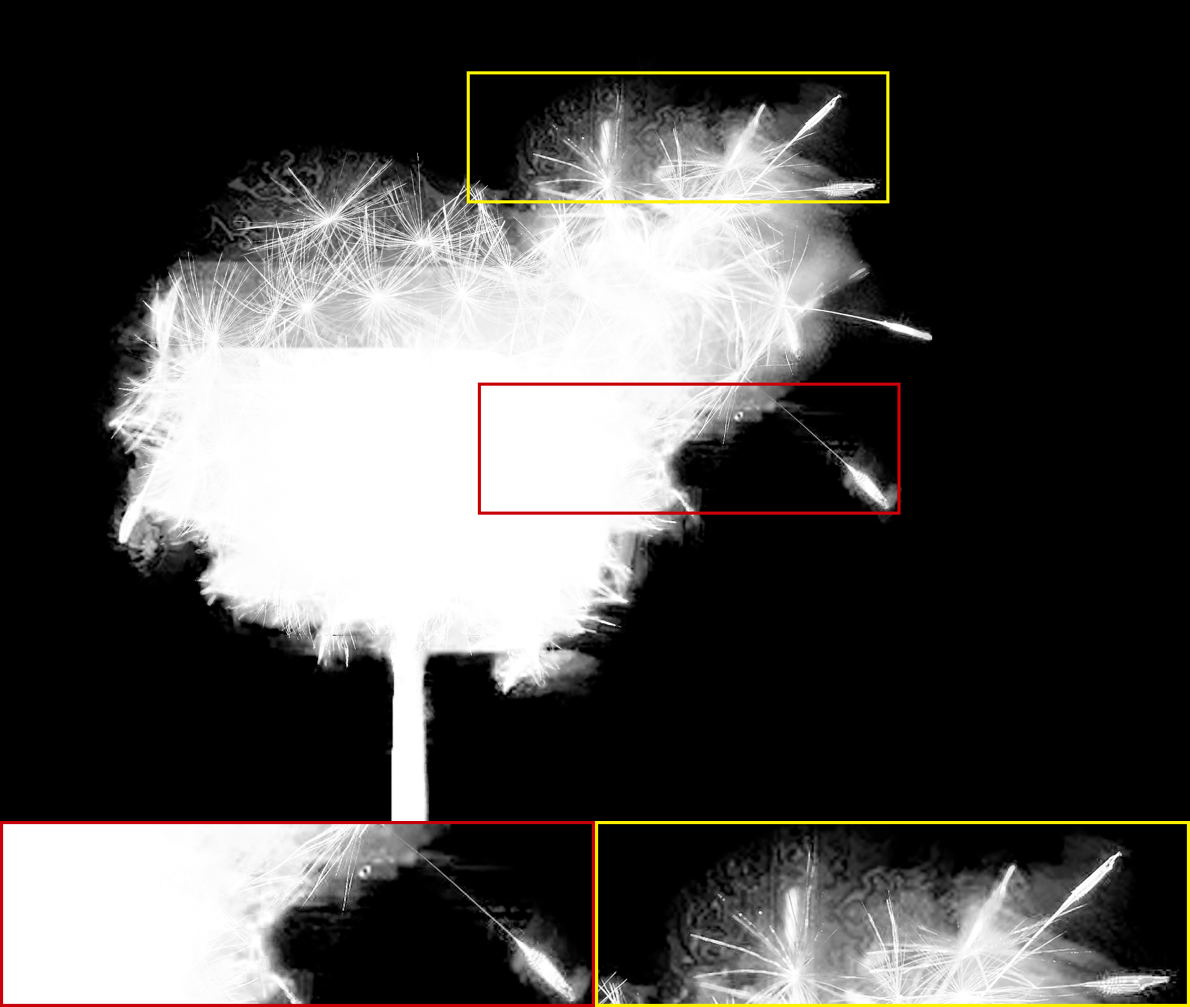} &
				\includegraphics[scale=0.1545]{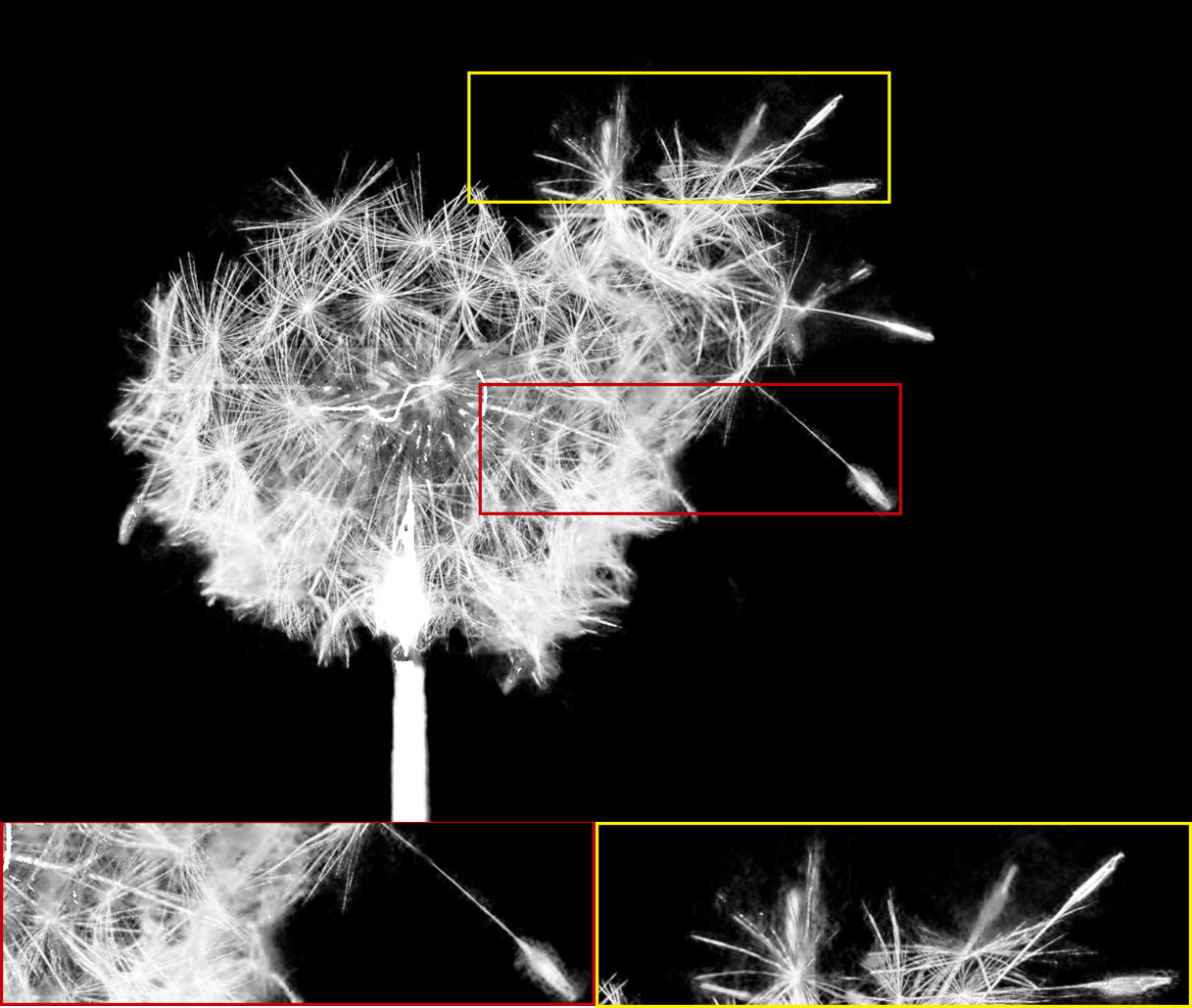} &
				\includegraphics[scale=0.166]{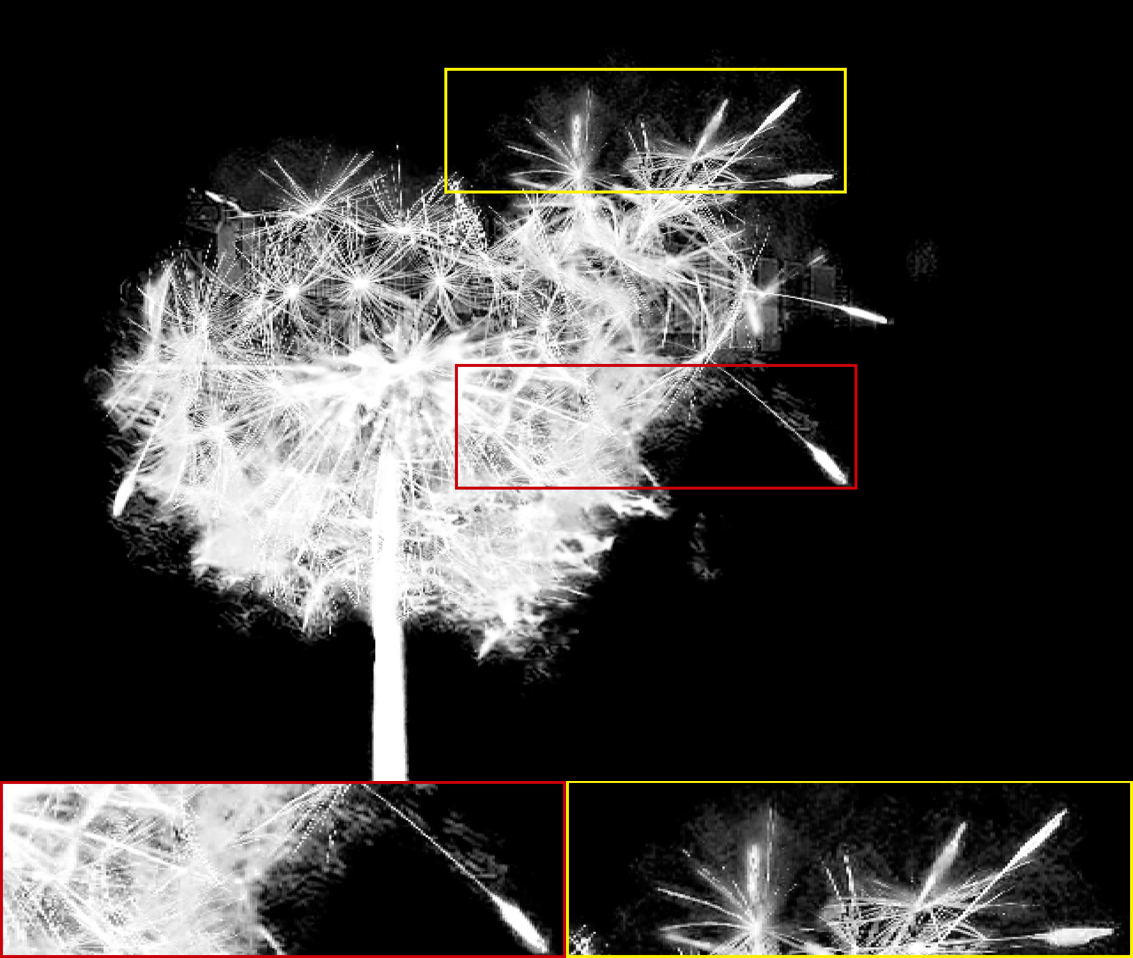} \\
				
				Input Image & Trimap & Closed Form~\cite{Levin2007A} & DIM~\cite{Xu2017Deep} & SampleNet~\cite{Tang_2019_CVPR} \\
				
				\includegraphics[scale=0.19067]{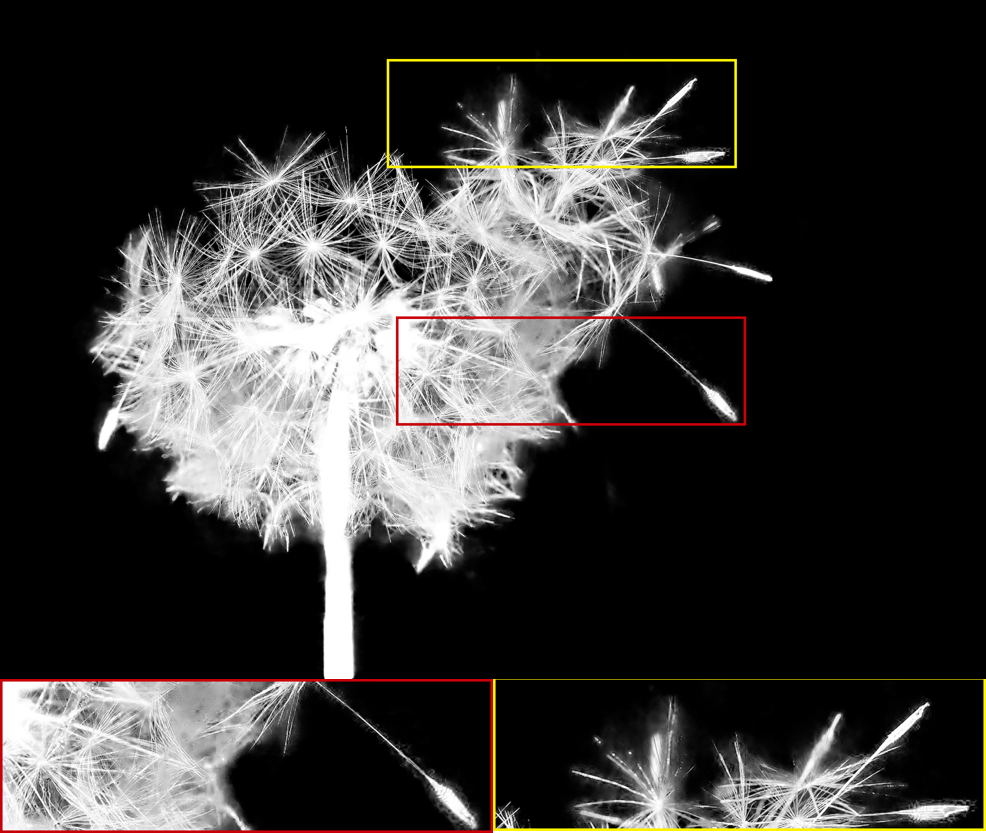} &
				\includegraphics[scale=0.18688]{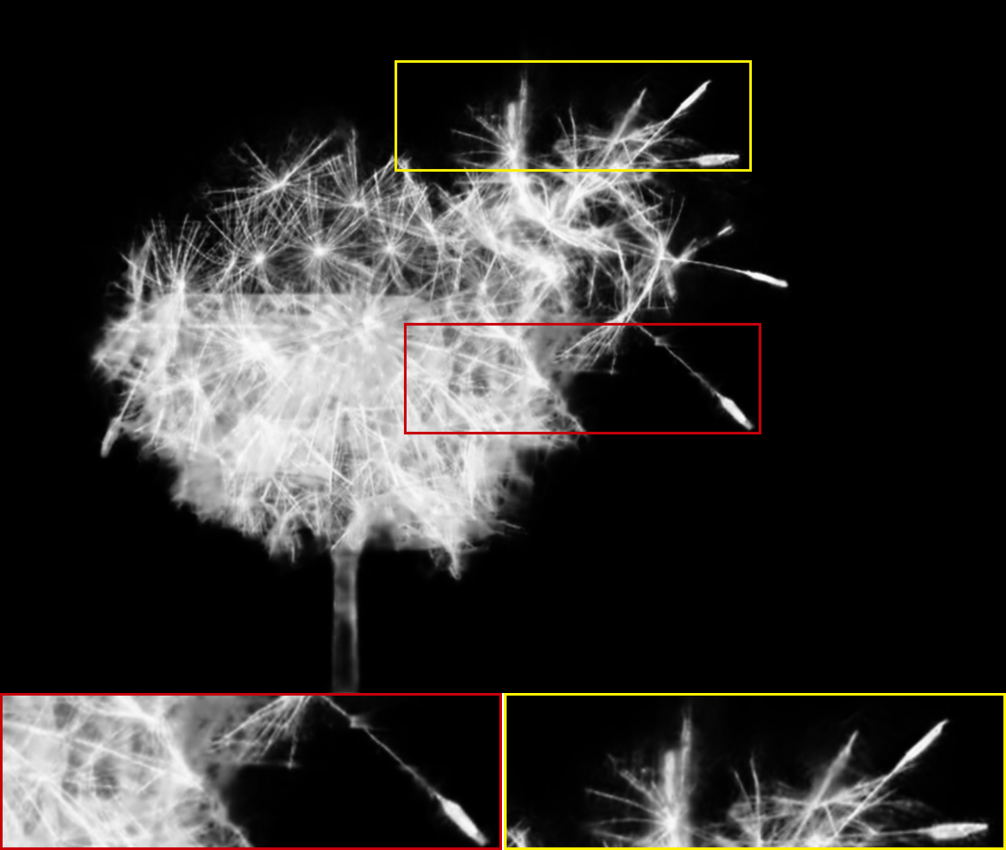} &
				\includegraphics[scale=0.27688]{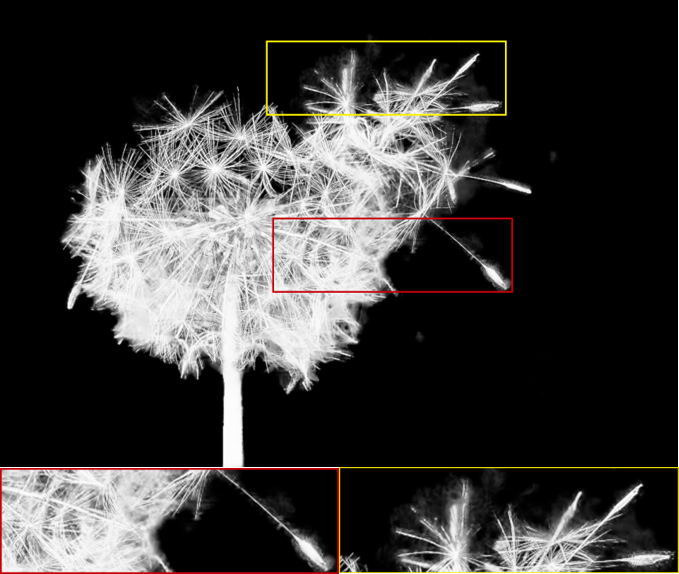} &
				\includegraphics[scale=0.2017]{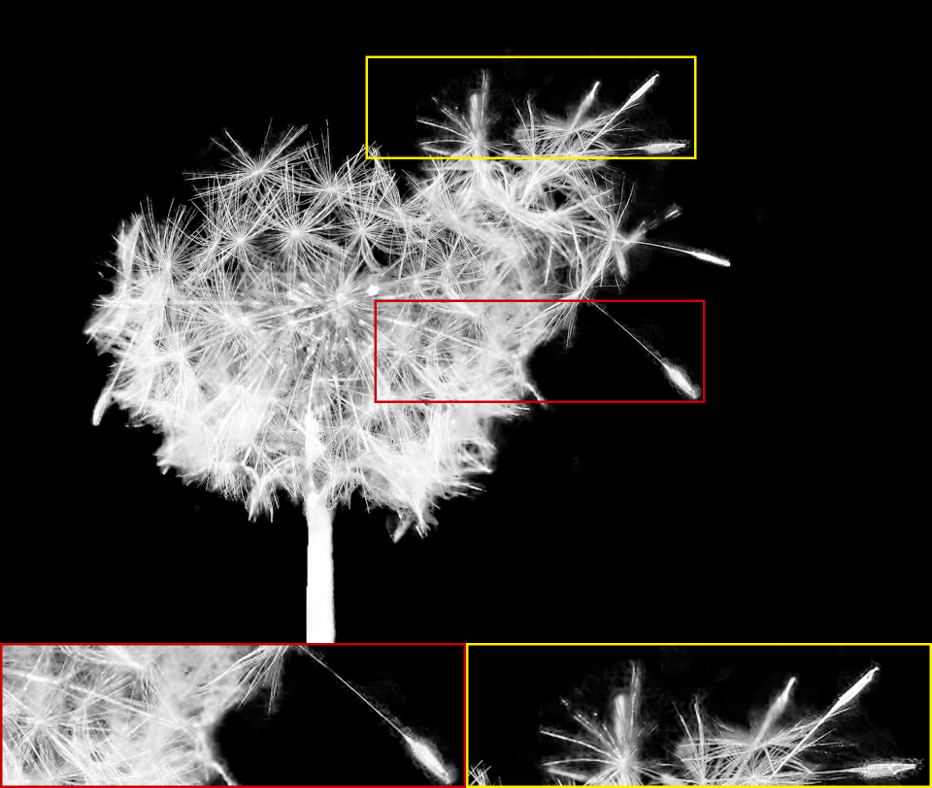} &
				\includegraphics[scale=0.19184]{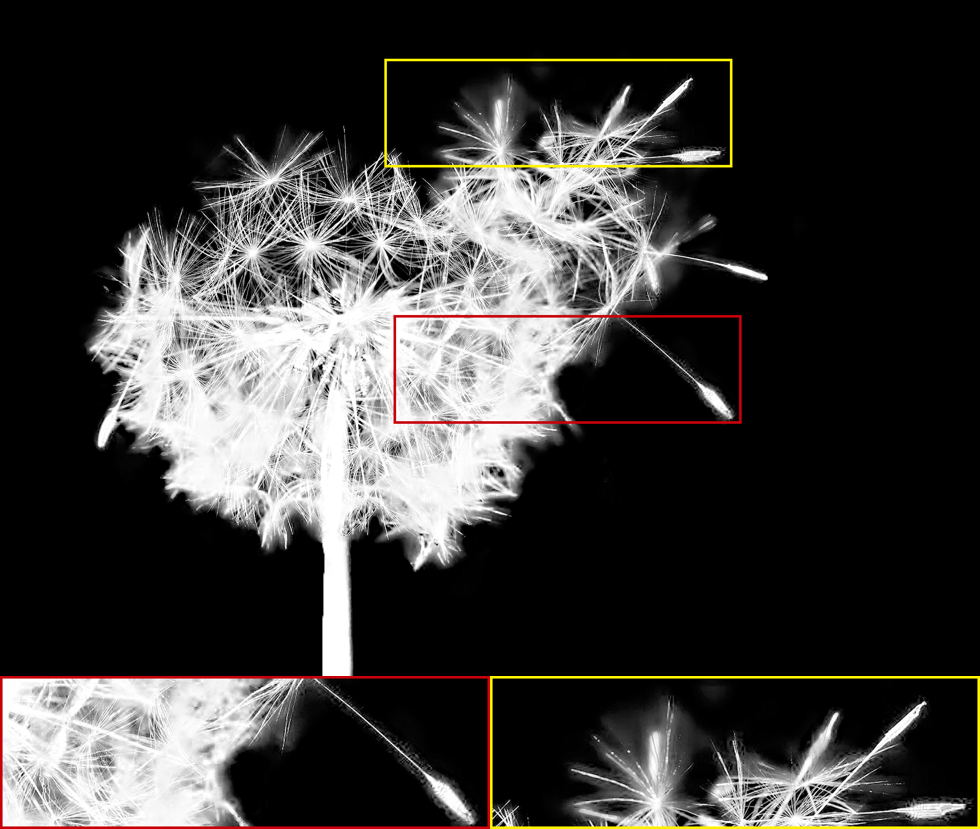} \\
				
				Context-Aware~\cite{hou2019context} & Late Fusion~\cite{Zhang2019CVPR} & HAttMatting~\cite{Qiao_2020_CVPR} & Ours & Ground Truth \\
				
				\includegraphics[scale=0.1775]{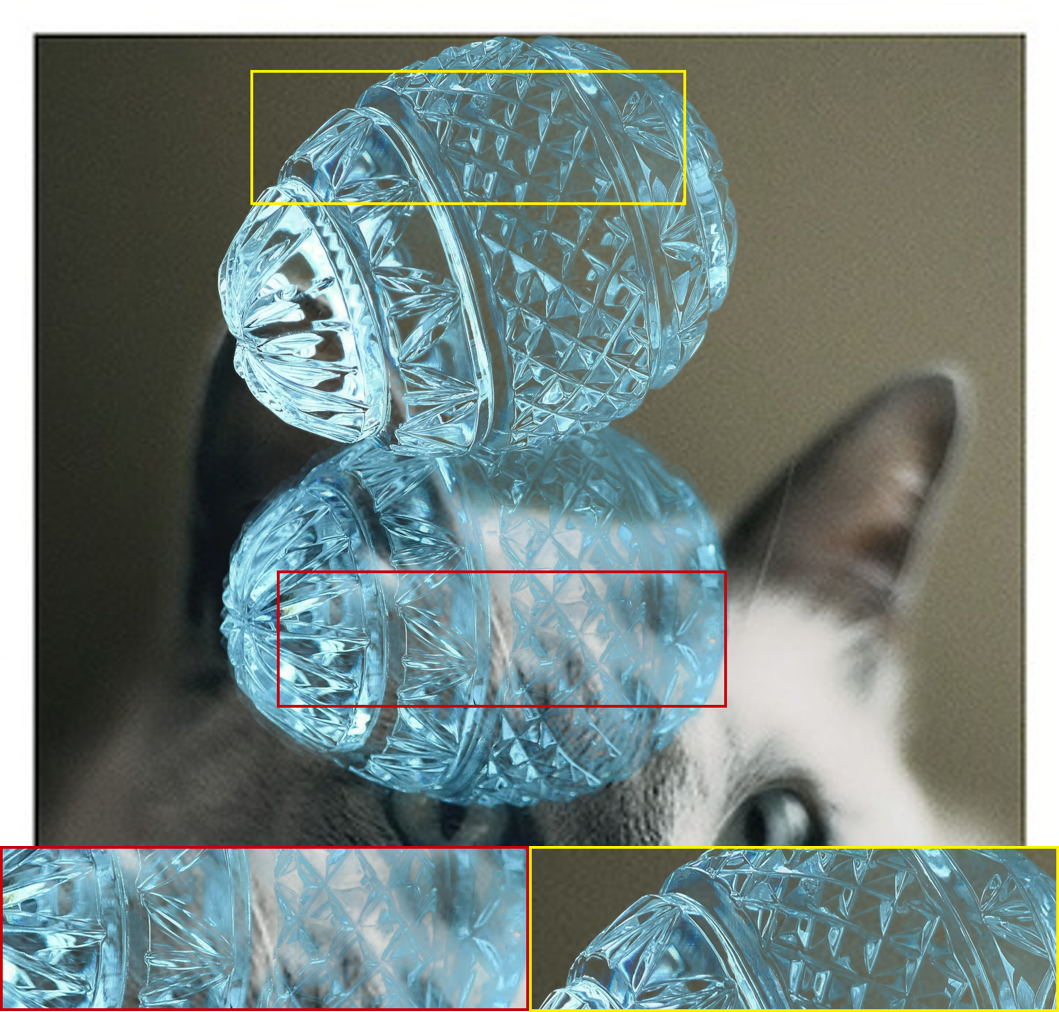} &
				\includegraphics[scale=0.04706]{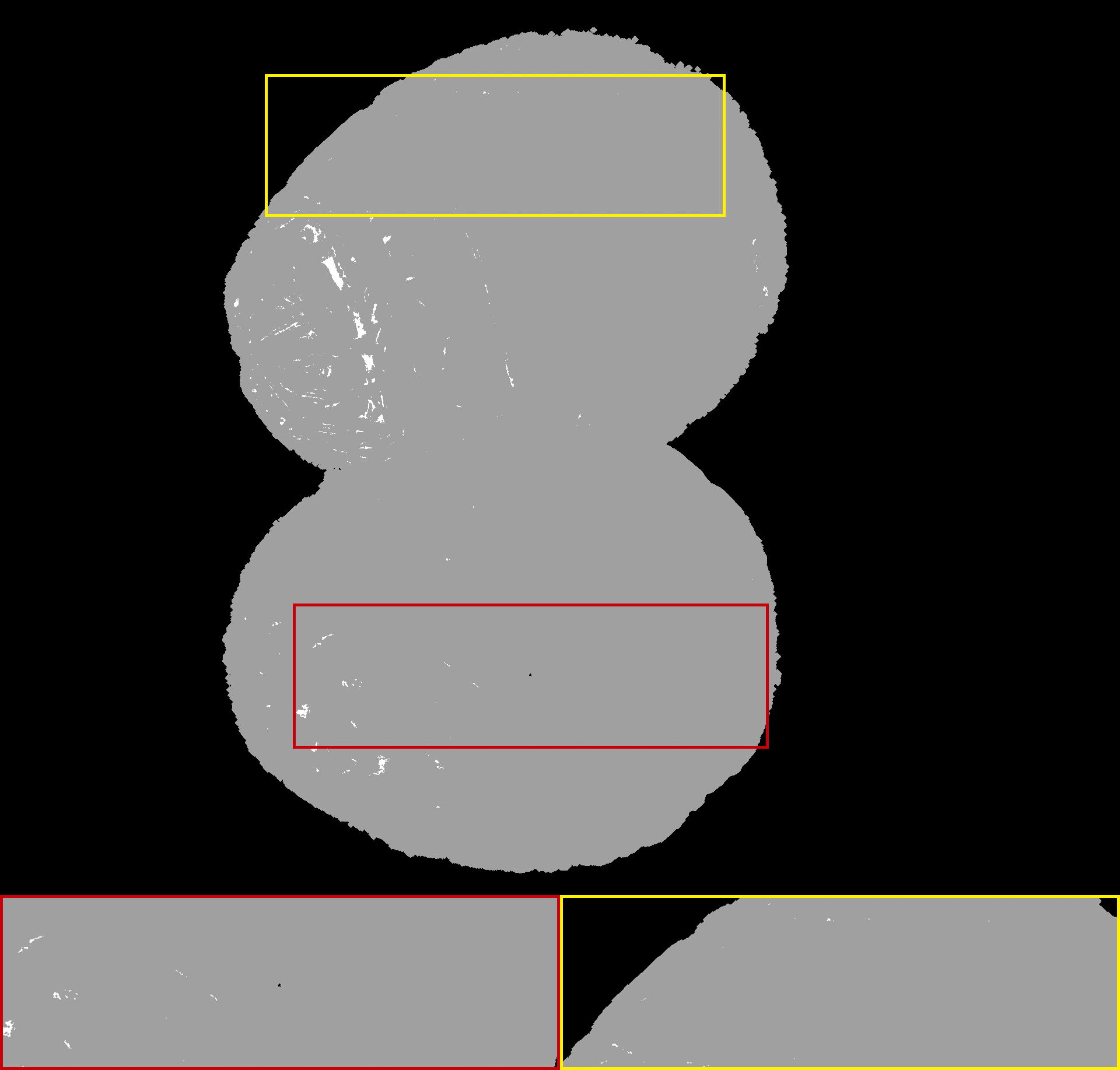} &
				\includegraphics[scale=0.1934]{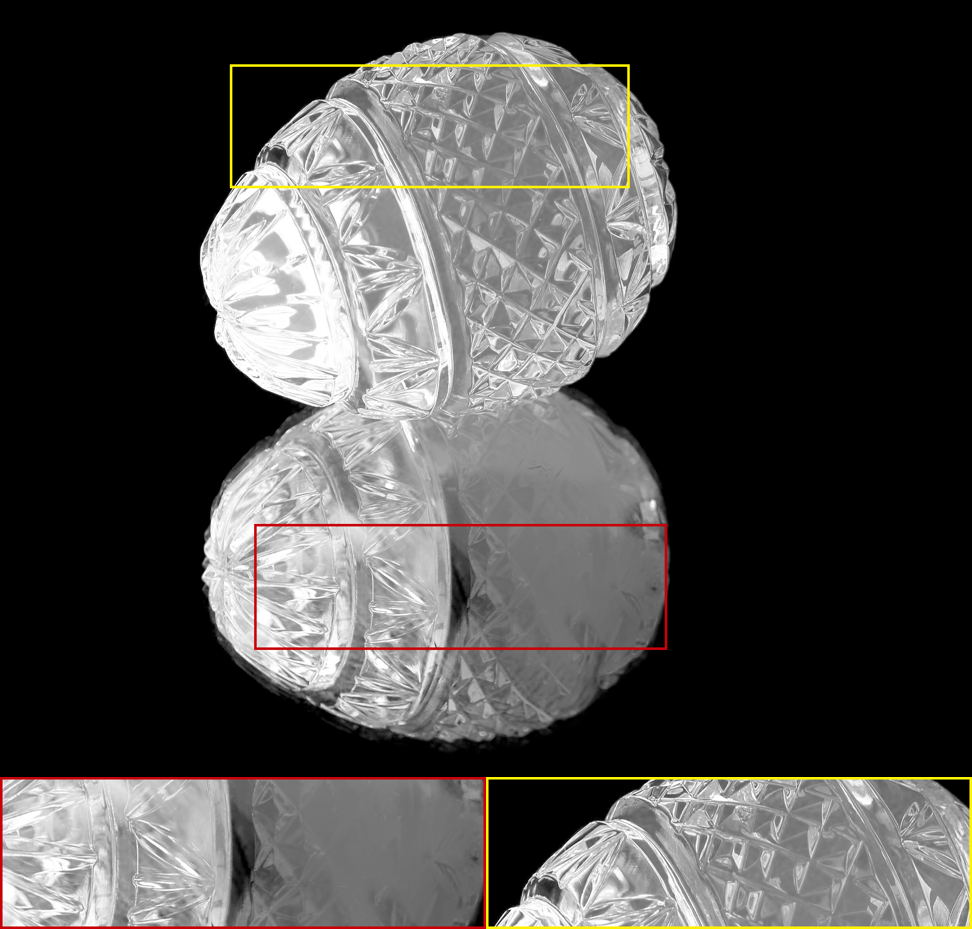} &
				\includegraphics[scale=0.18146]{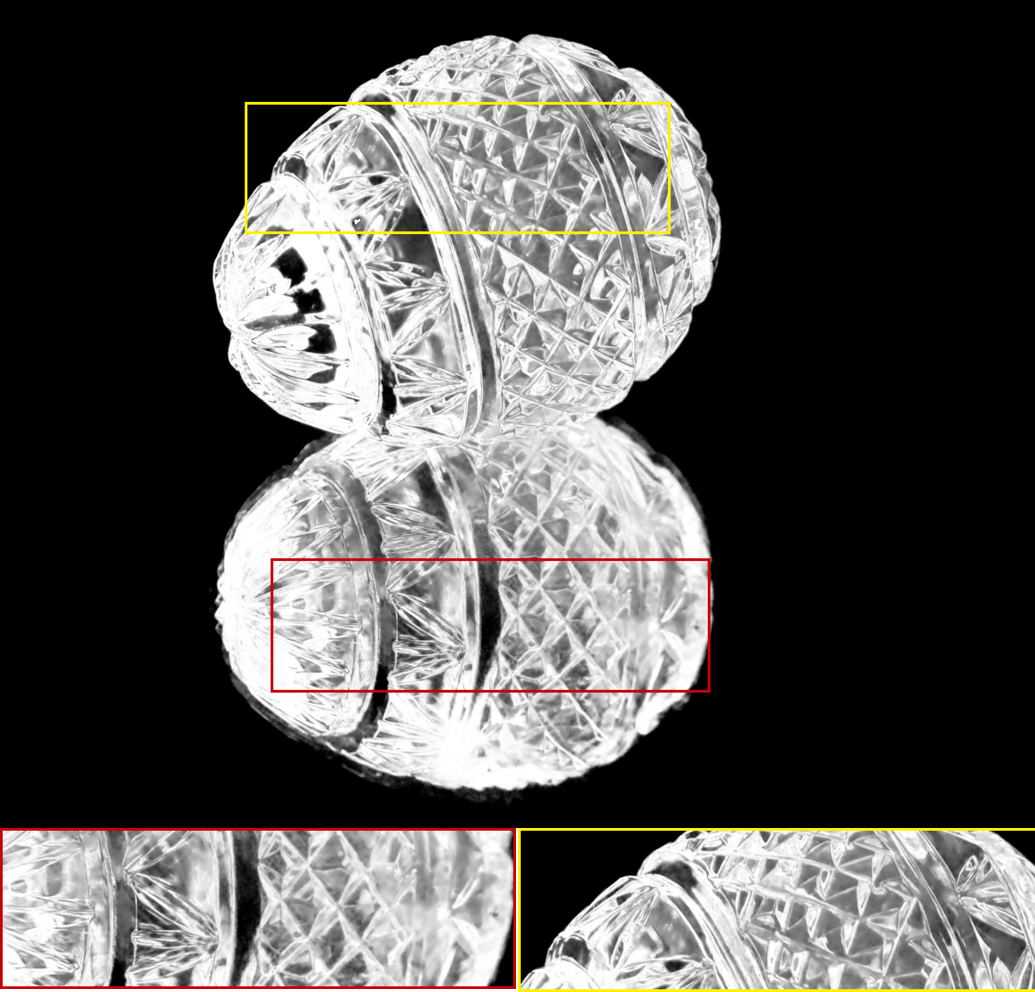} &
				\includegraphics[scale=0.19356]{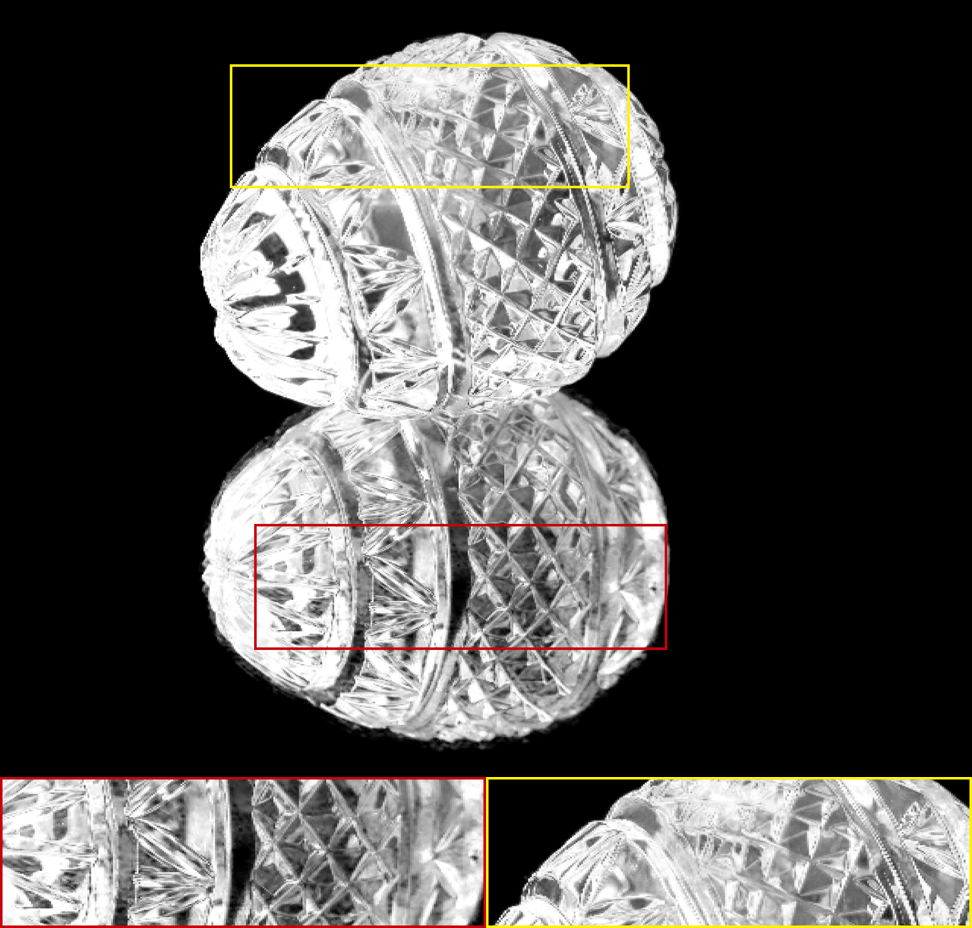} \\
				
				Input Image & Trimap & Closed Form~\cite{Levin2007A} & DIM~\cite{Xu2017Deep} & SampleNet~\cite{Tang_2019_CVPR} \\
				
				\includegraphics[scale=0.1782]{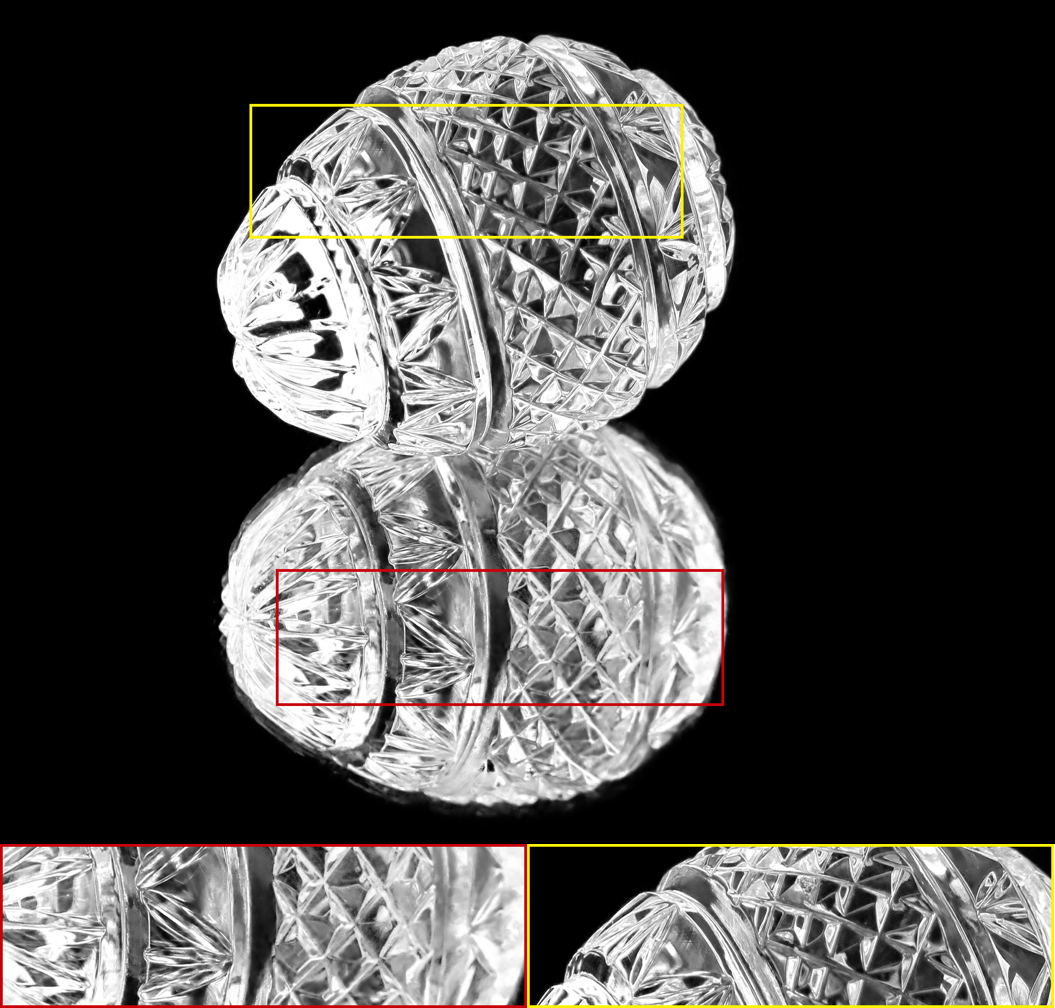} &
				\includegraphics[scale=0.04706]{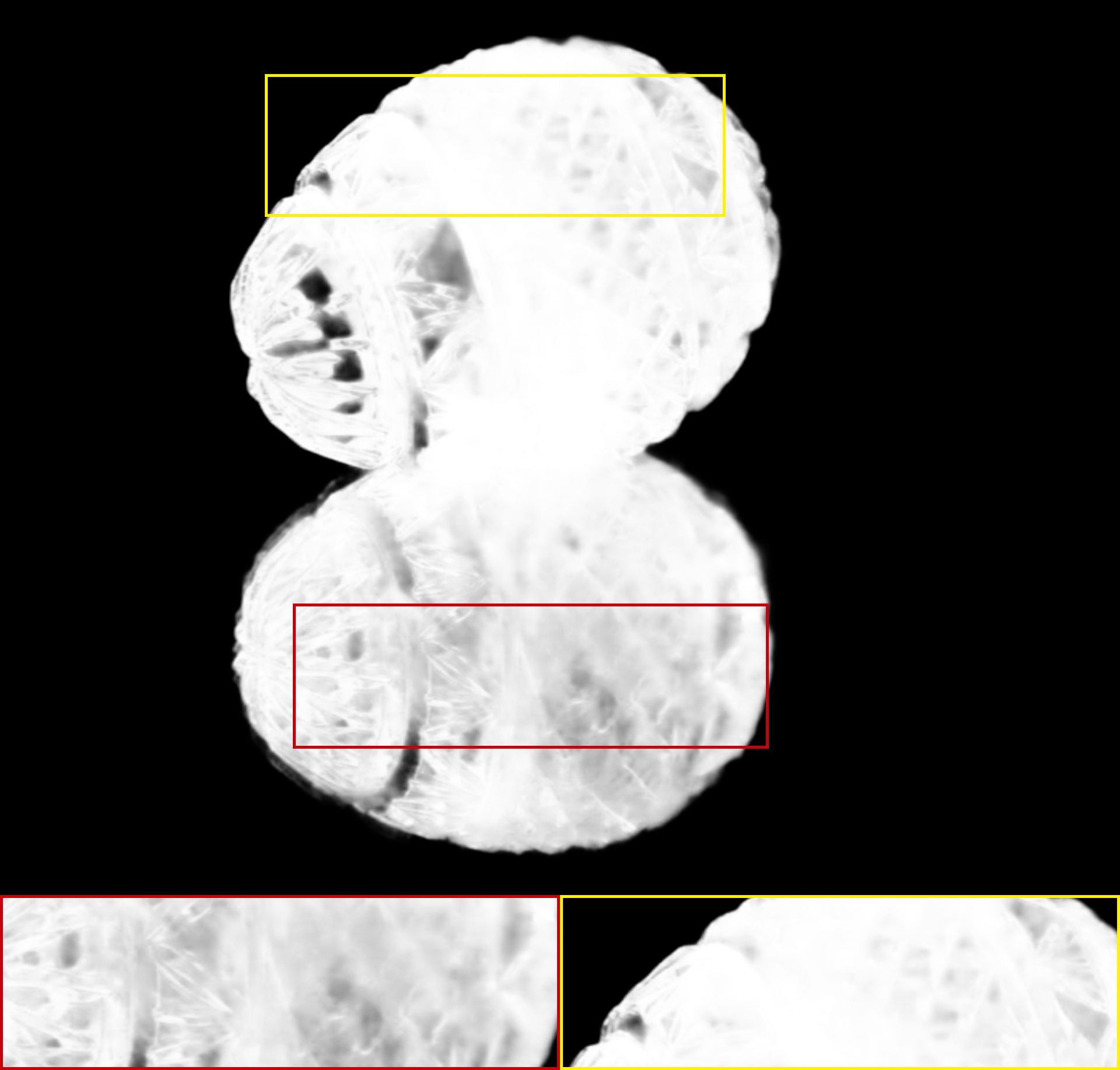} &
				\includegraphics[scale=0.1901]{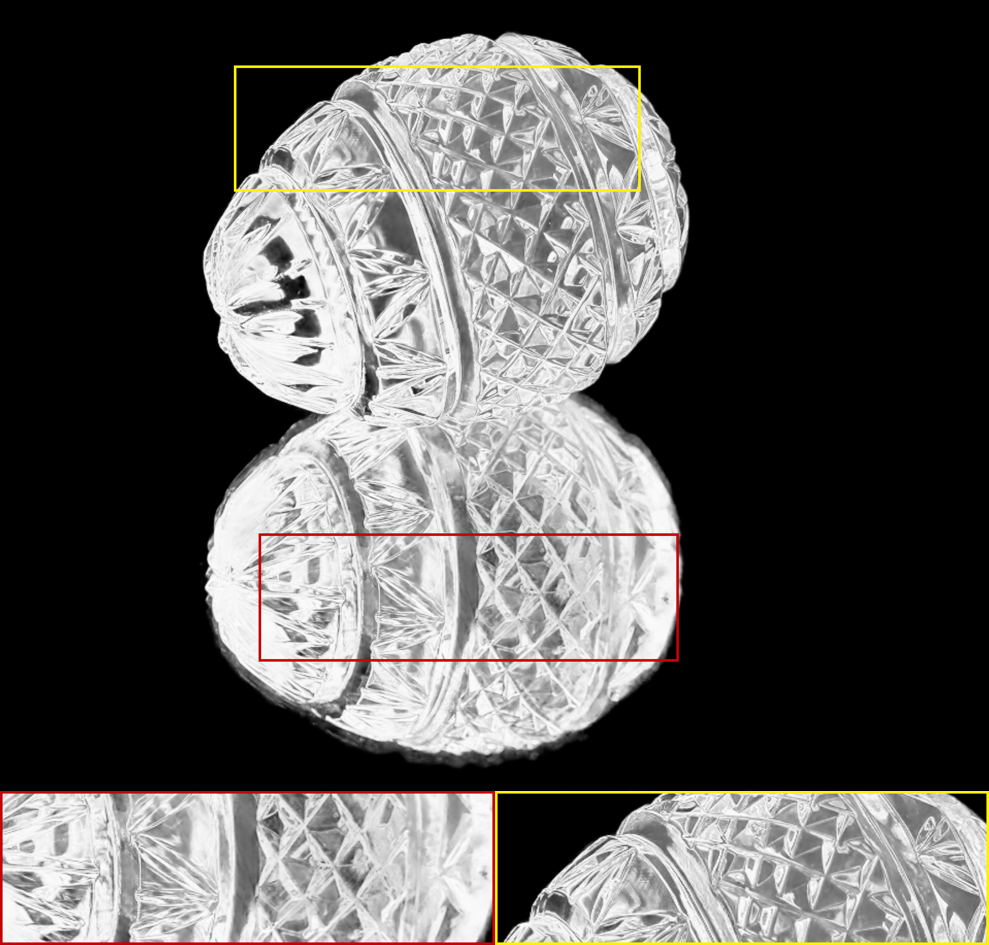} &
				\includegraphics[scale=0.1996]{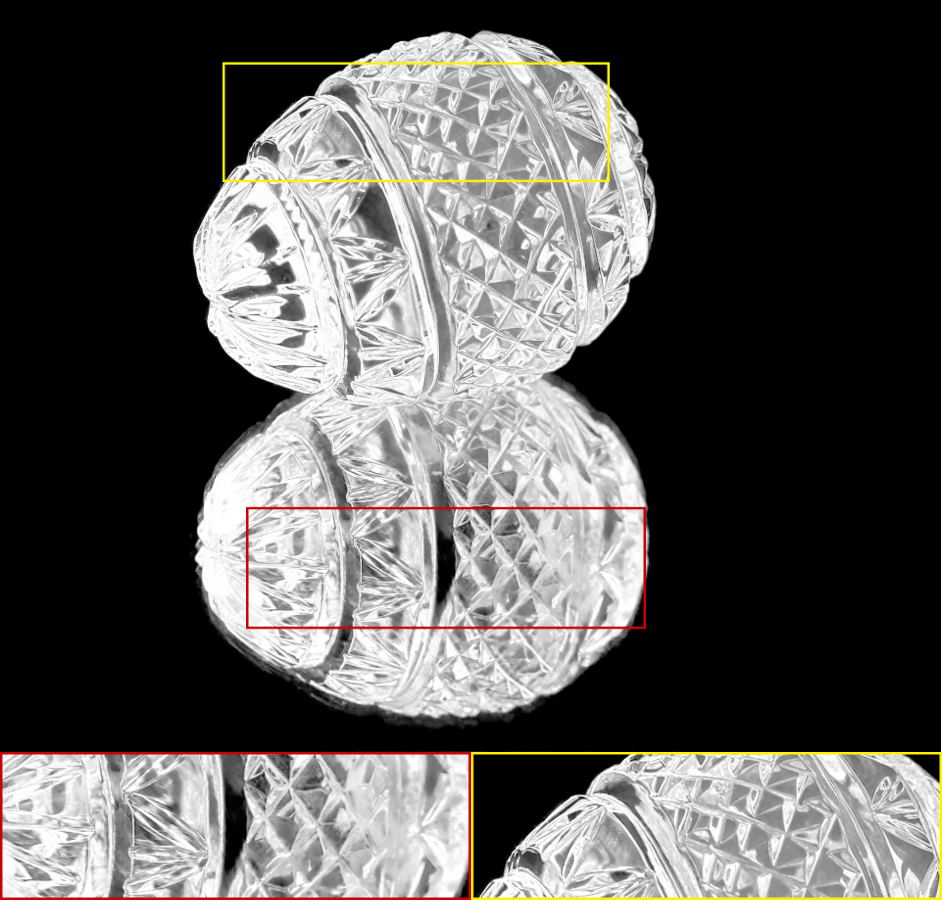} &
				\includegraphics[scale=0.1843]{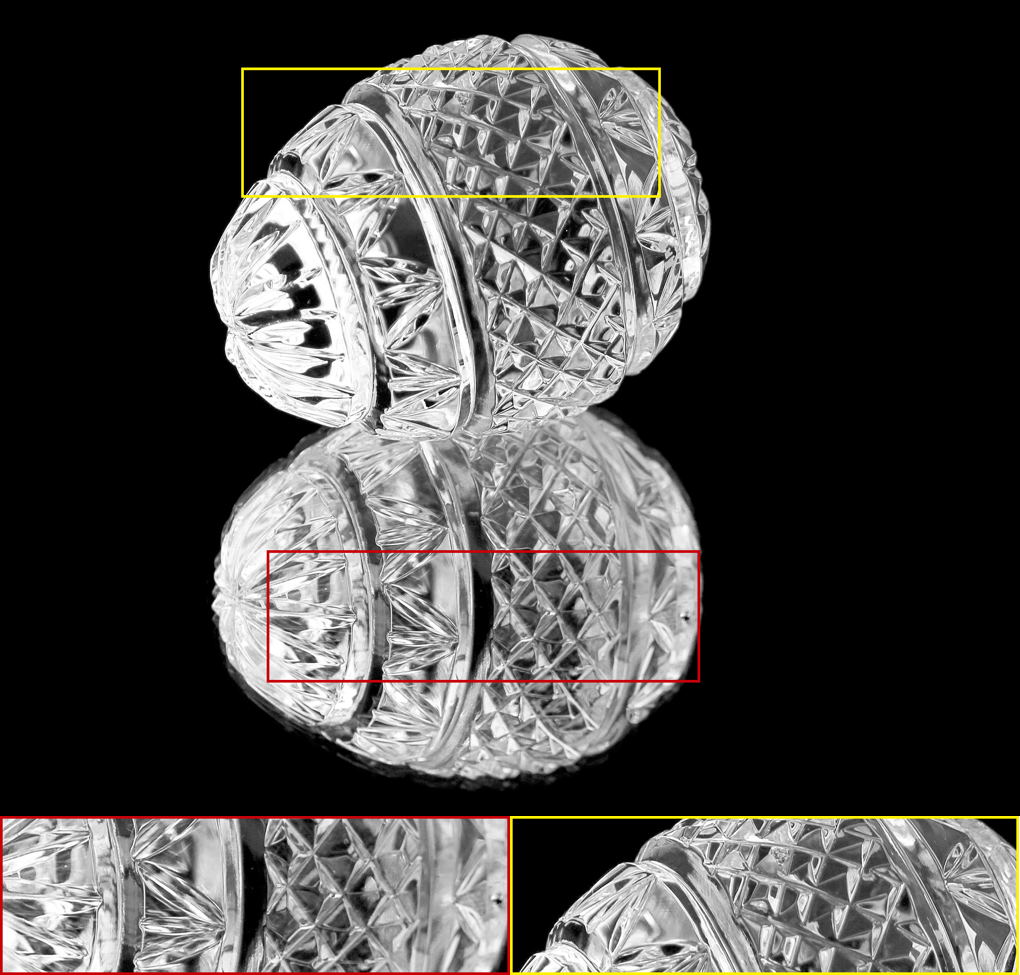} \\
				
				Context-Aware~\cite{hou2019context} & Late Fusion~\cite{Zhang2019CVPR} & HAttMatting~\cite{Qiao_2020_CVPR} & Ours & Ground Truth \\
				
				\includegraphics[scale=0.2307]{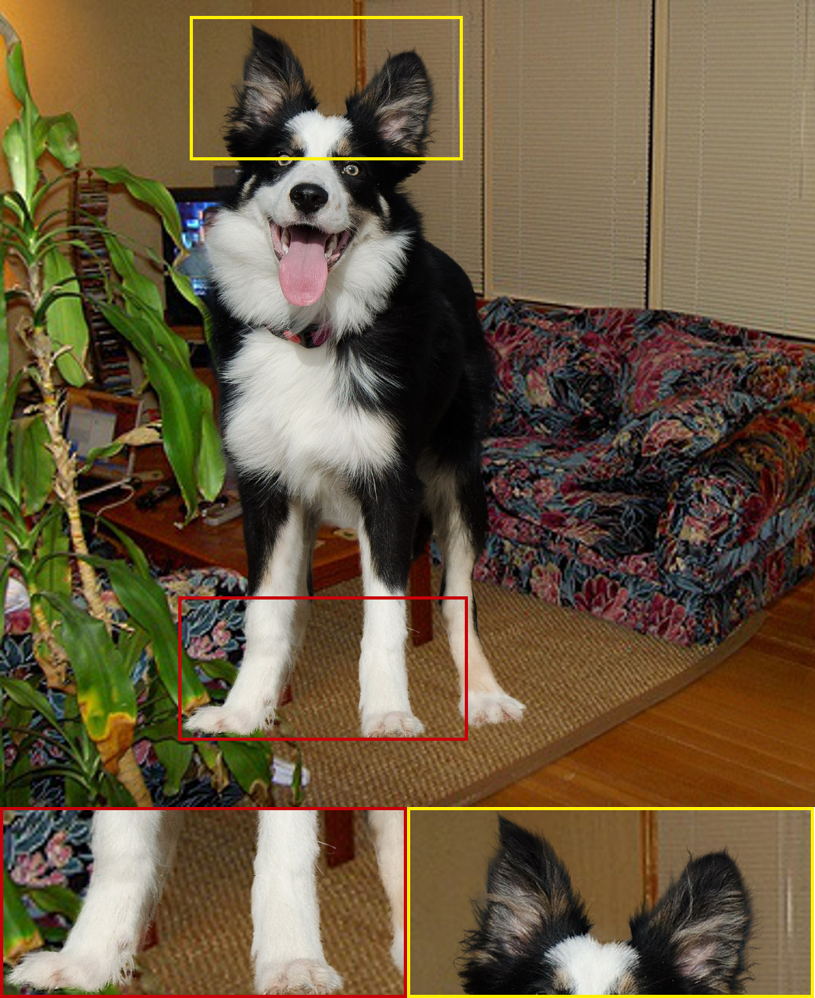} &
				\includegraphics[scale=0.07052]{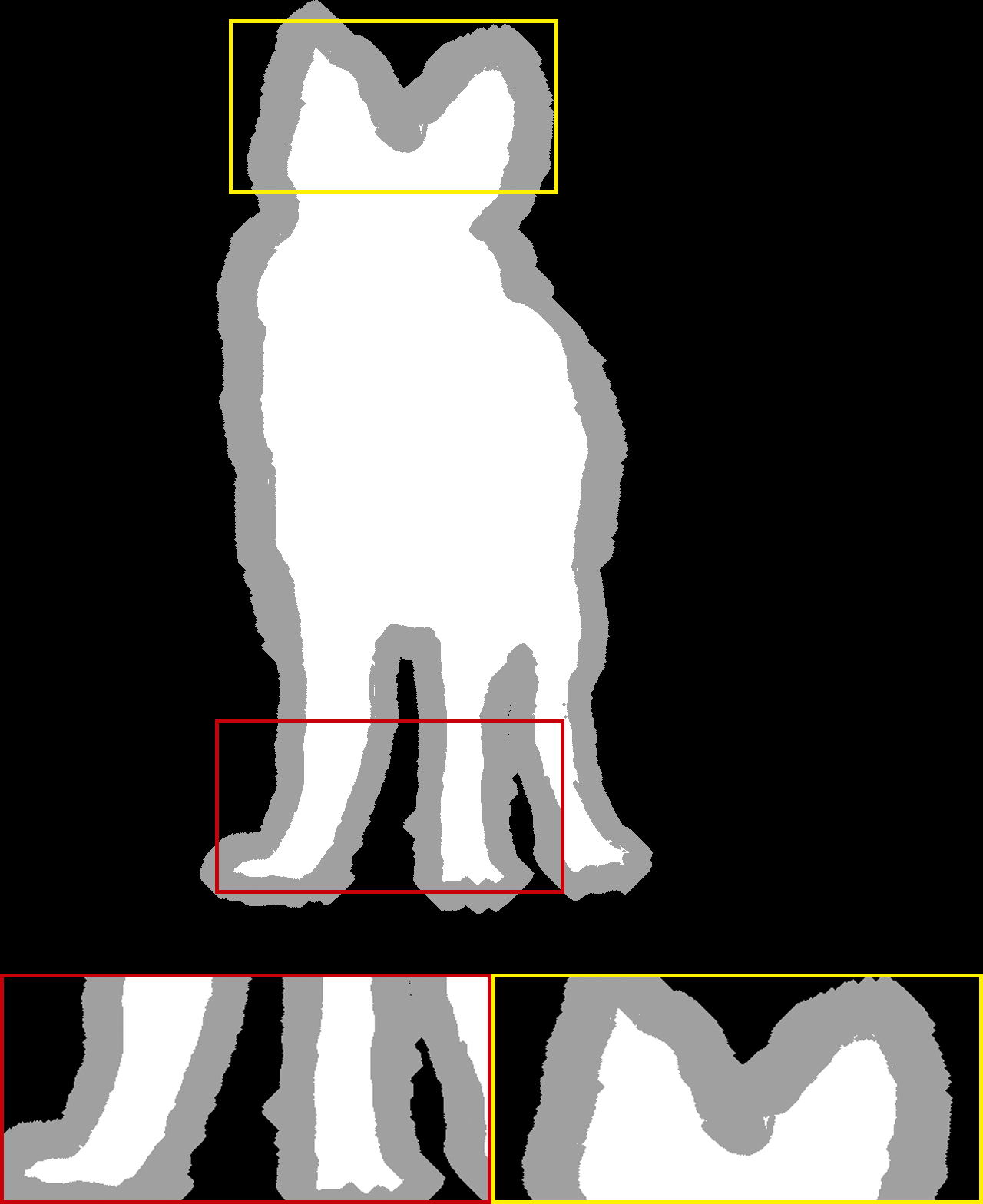} &
				\includegraphics[scale=0.21734]{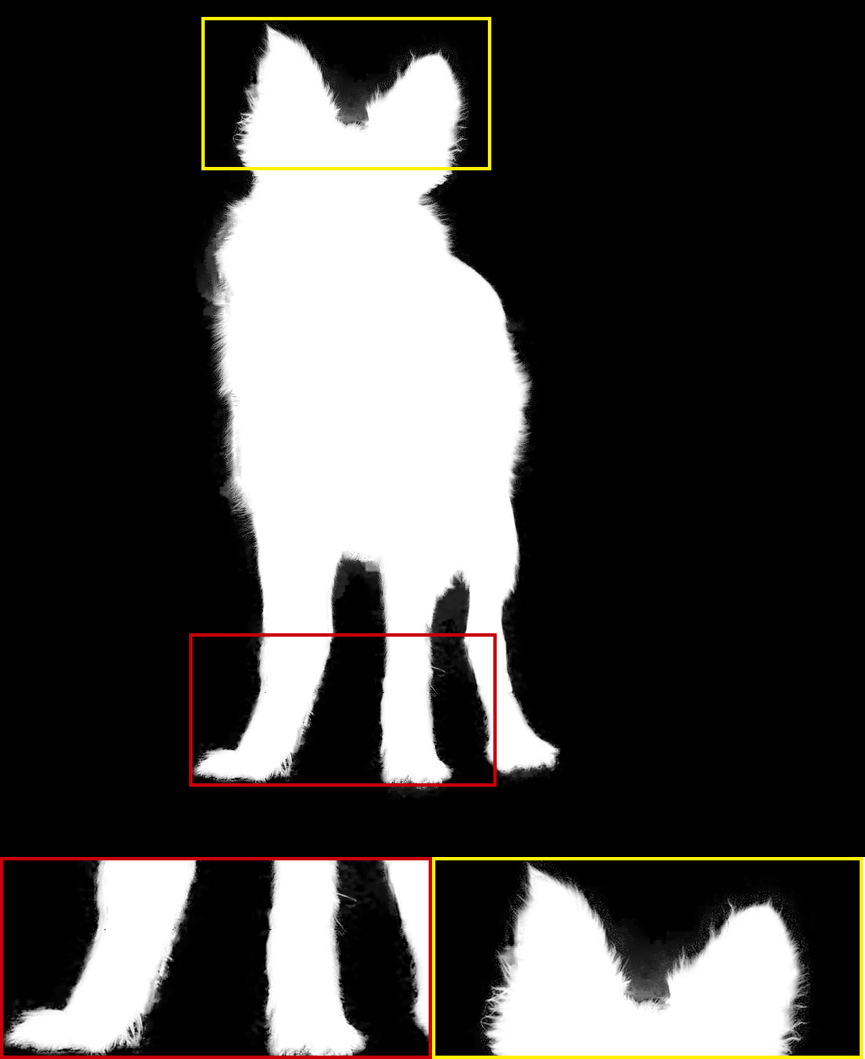} &
				\includegraphics[scale=0.23858]{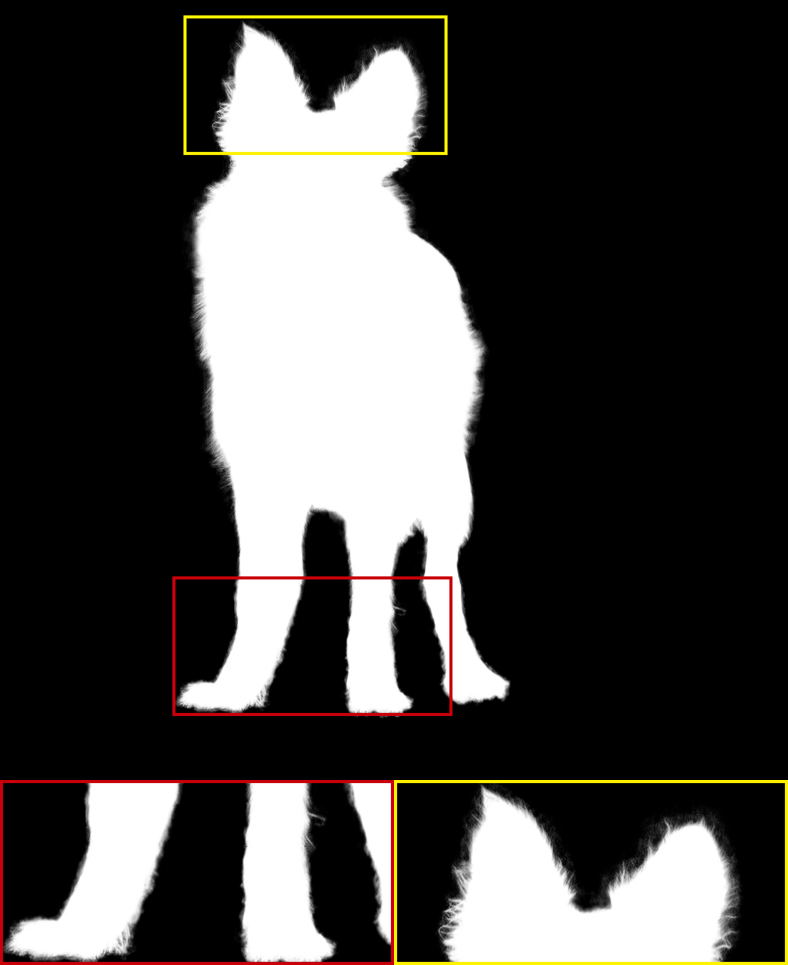} &
				\includegraphics[scale=0.26257]{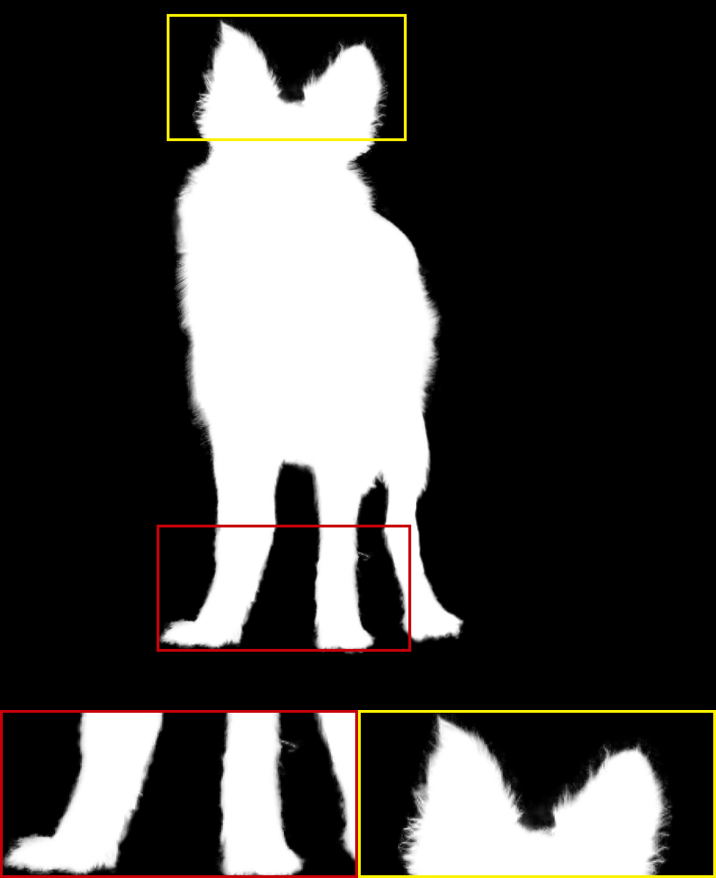} \\
				
				Input Image & Trimap & Closed Form~\cite{Levin2007A} & DIM~\cite{Xu2017Deep} & SampleNet~\cite{Tang_2019_CVPR} \\
				
				\includegraphics[scale=0.23182]{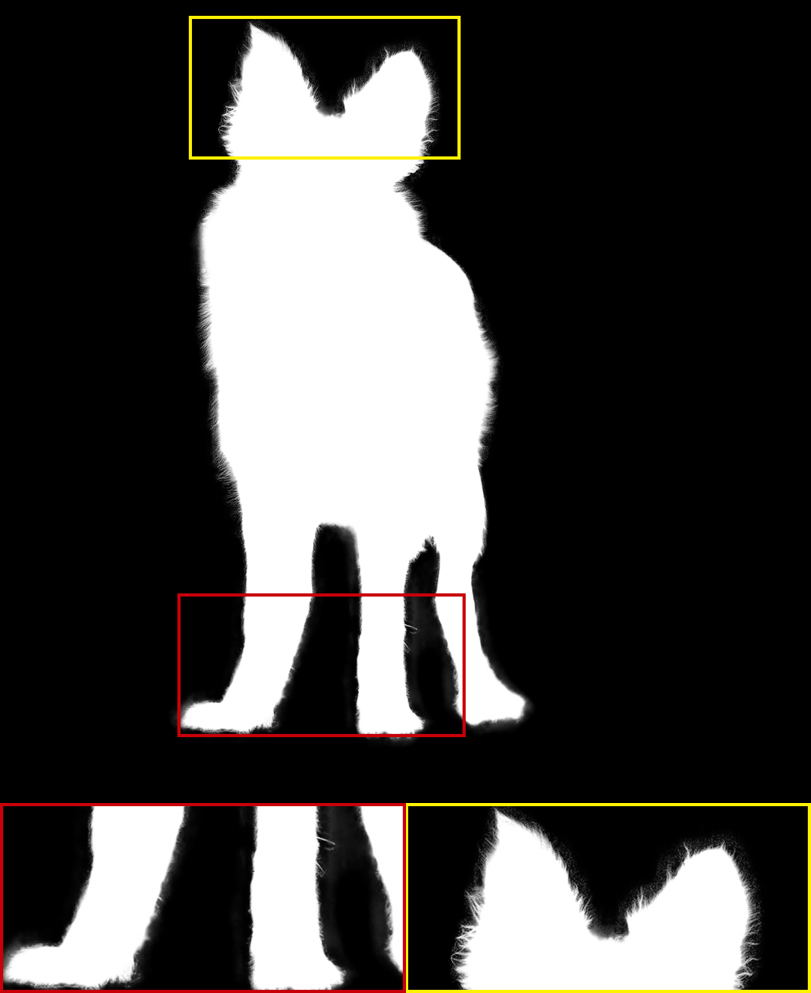} &
				\includegraphics[scale=0.07048]{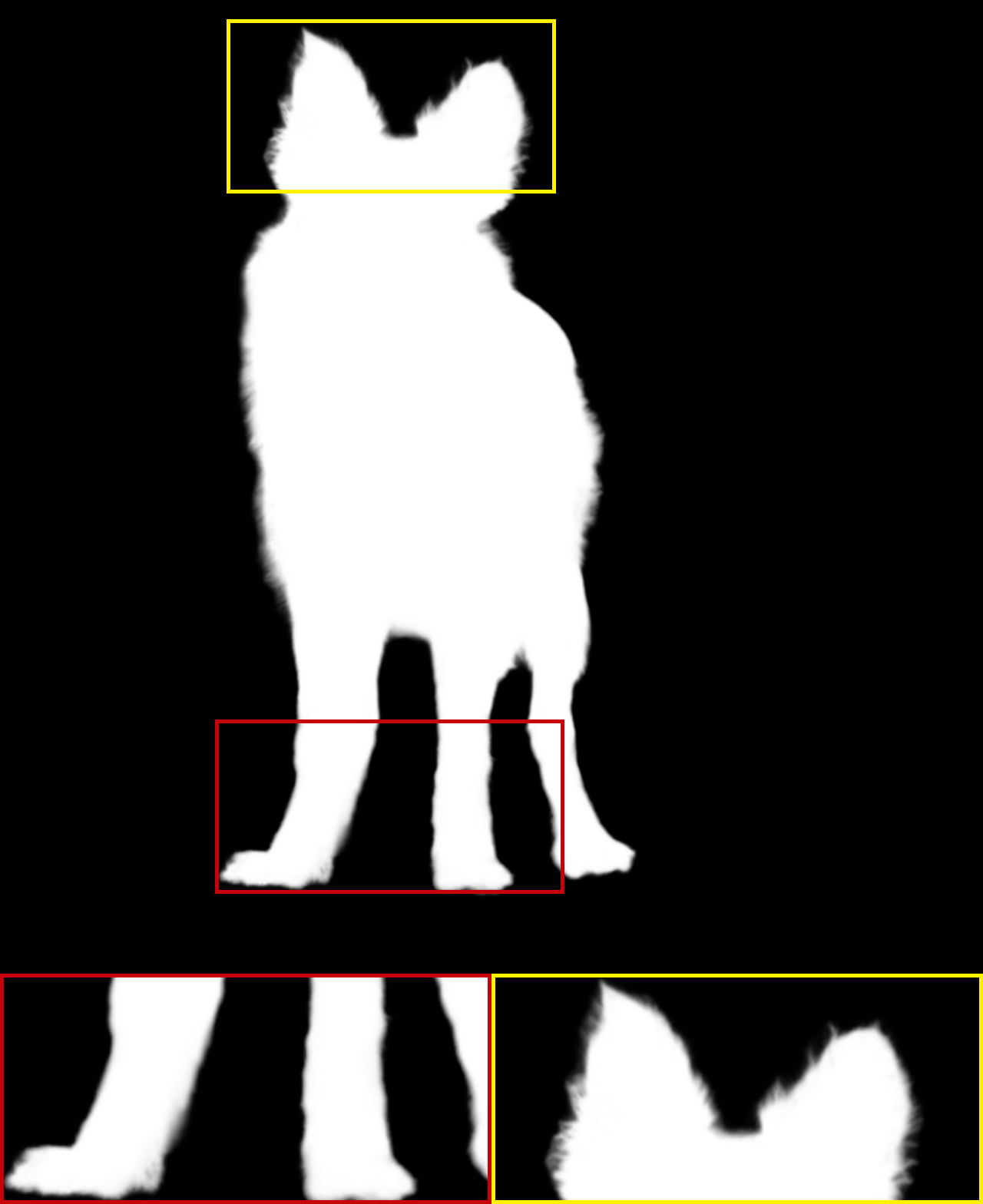} &
				\includegraphics[scale=0.31756]{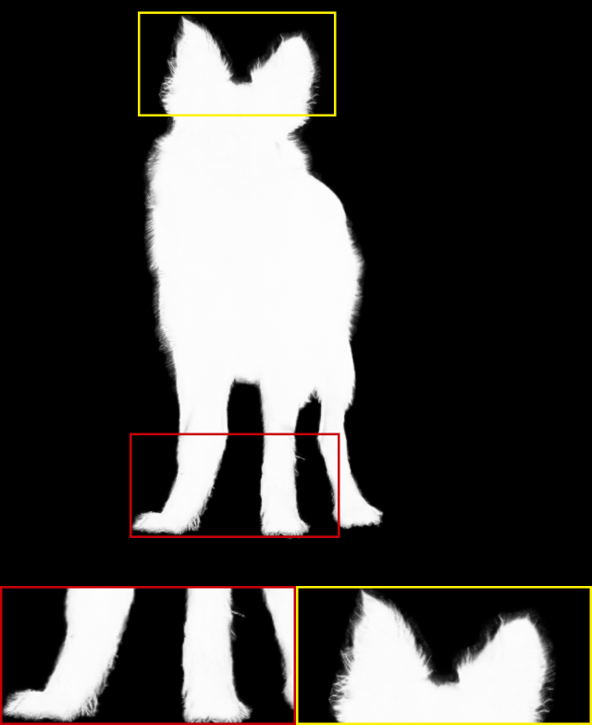} &
				\includegraphics[scale=0.2423]{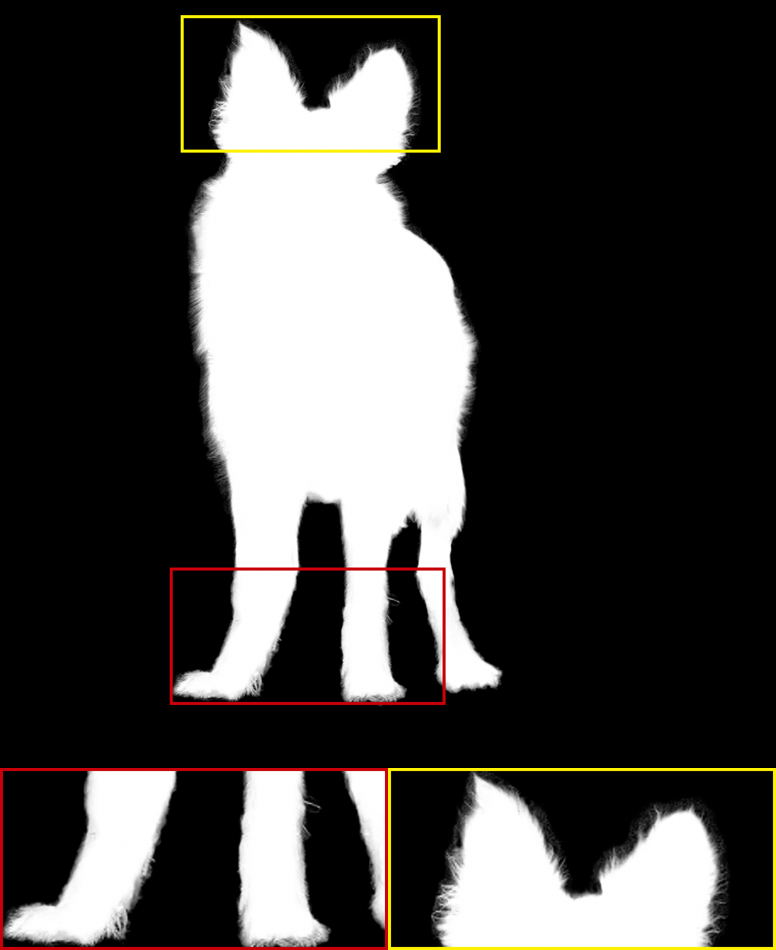} &
				\includegraphics[scale=0.2259]{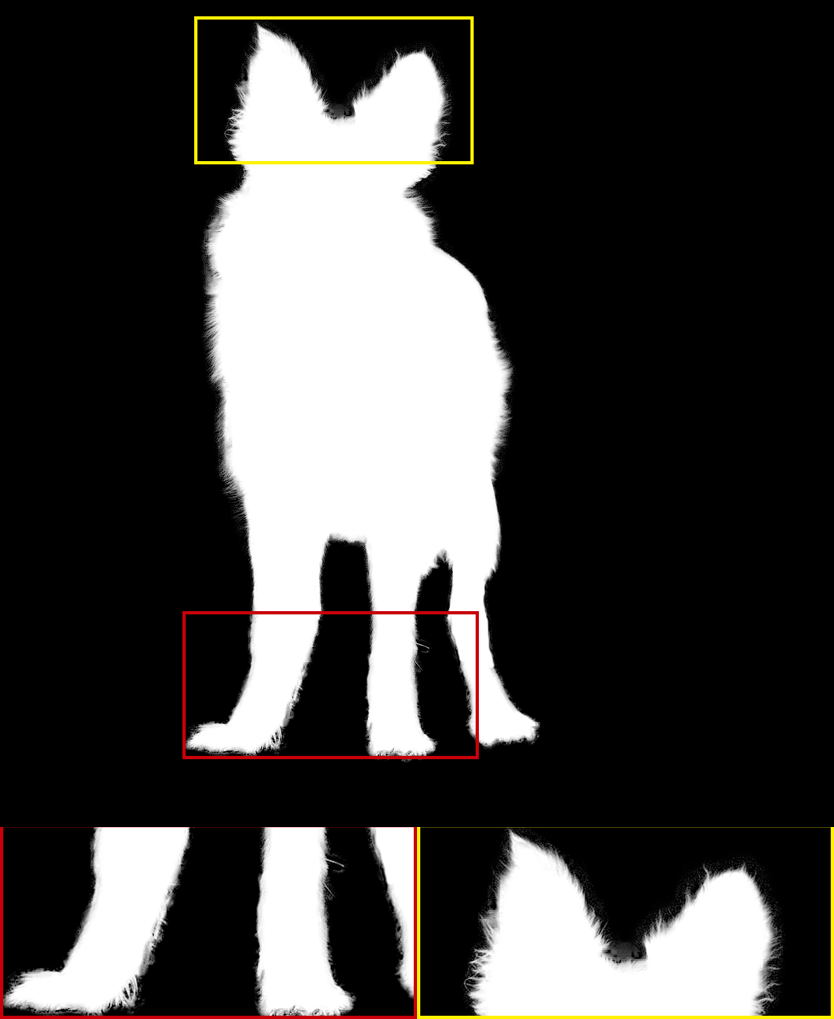} \\
				
				Context-Aware~\cite{hou2019context} & Late Fusion~\cite{Zhang2019CVPR} & HAttMatting~\cite{Qiao_2020_CVPR} & Ours & Ground Truth \\
		\end{tabular}}
		\caption{Comparison of the alpha mattes on the Composition-1k testing set~\cite{Xu2017Deep}. Some details and textures are zoomed in and placed under the images. }
		\label{fig:com_to_DIM}
	\end{figure*}
	
	\subsection{The Composition-1K testing dataset}
	\label{ssec:com_on_dim}
	
	Here we train the proposed MSIA-matte based on the training images in~\cite{Xu2017Deep} and evaluate the optimized model on the Composition-1K testing set. We also compare our model with state-of-the-art methods, including Shared Matting~\cite{gastal2010shared}, Learning-Based~\cite{zheng2009learning}, Comprehensive Sampling~\cite{Shagrian2013}, Global Matting~\cite{Rhemann2011A}, ClosedForm~\cite{Levin2007A}, KNN Matting~\cite{Chen2013KNN}, Information Flow~\cite{Aksoy2017Designing}, DCNN~\cite{Cho2016Natural}, DIM~\cite{Xu2017Deep}, AlphaGAN~\cite{lutz2018alphagan}, SampleNet~\cite{Tang_2019_CVPR}, Context-aware~\cite{hou2019context}, IndexNet~\cite{hao2019indexnet}, Late Fusion~\cite{Zhang2019CVPR} and HAttMatting~\cite{Qiao_2020_CVPR}. Table~\ref{tab:quantitative_adobe} summarizes the results of all the methods in calculating the four metrics on the full alpha images. The results in Table~\ref{tab:quantitative_adobe} are from our re-implementations or relevant researches~\cite{Qiao_2020_CVPR}.
	\begin{table}[t]
		\small
		\centering
		\setlength{\tabcolsep}{1.2mm}{
			\begin{tabular}{l|ccccc}
				\hline
				Methods & SAD$\downarrow$ & MSE$\downarrow$  & Gradient$\downarrow$  & Connectivity$\downarrow$  \\
				\hline
				Shared Matting~\cite{gastal2010shared} & 125.37 & 0.029 & 144.28 & 123.53 \\
				Learning Based~\cite{zheng2009learning} & 95.04 & 0.018 & 76.63 & 98.92 \\
				Comprehensive~\cite{Shagrian2013} & None & None & None & None \\
				Global Matting~\cite{Rhemann2011A} & 156.88 & 0.042 & 112.28 & 155.08 \\
				ClosedForm~\cite{Levin2007A} & 124.68 & 0.025 & 115.31 & 106.06 \\
				KNN Matting~\cite{Chen2013KNN} & 126.24 & 0.025 & 117.17 & 131.05 \\
				DCNN~\cite{Cho2016Natural} & 115.82 & 0.023 & 107.36 & 111.23 \\
				Information Flow~\cite{Aksoy2017Designing} & 70.36 & 0.013 & 42.79 & 70.66 \\
				DIM~\cite{Xu2017Deep} & 48.87 & 0.008 & 31.04 & 50.36 \\
				AlphaGAN~\cite{lutz2018alphagan} & 90.94 & 0.018 & 93.92 & 95.29 \\
				SampleNet~\cite{Tang_2019_CVPR} & 48.03 & 0.008 & 35.19 & 56.55 \\
				Context Aware~\cite{hou2019context} & \color{red}{38.73} & \color{red}{0.004} & \color{red}{26.13} & \color{red}{35.89} \\
				IndexNet~\cite{hao2019indexnet} & 44.52 & 0.005 & 29.88 & 42.37 \\
				\hline
				\hline
				\rowcolor{mygray}
				Late Fusion~\cite{Zhang2019CVPR} & 58.34 & 0.011 & 41.63 & 59.74\\
				\rowcolor{mygray}
				HAttMatting~\cite{Qiao_2020_CVPR} & \textbf{44.01} & \textbf{0.007} & 29.26  & 46.41\\
				\rowcolor{mygray}
				Ours & 47.86 & \textbf{0.007} & \textbf{28.61} & \textbf{43.39}\\
				\hline
		\end{tabular}}
		\caption{The quantitative results on the Composition-1K testing set. The methods in gray (the Late Fusion~\cite{Zhang2019CVPR}, HAttMatting~\cite{Qiao_2020_CVPR}, and our method) can produce alpha mattes with single RGB images, while others require trimaps to confine the transition regions. }
		\label{tab:quantitative_adobe}
	\end{table}
	
	\begin{figure*}[t]
		\arrayrulecolor{tabcolor}
		\centering
		\setlength{\tabcolsep}{1pt}\small{
			\begin{tabular}{c|c|c|c|c}
				\includegraphics[scale=0.20856]{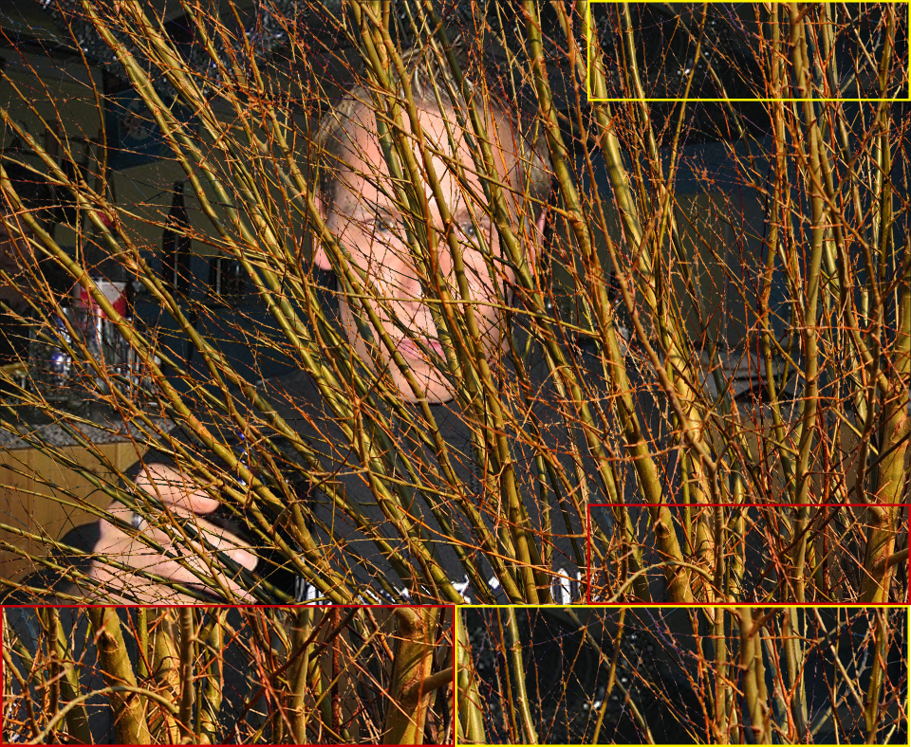} &
				\includegraphics[scale=0.1578]{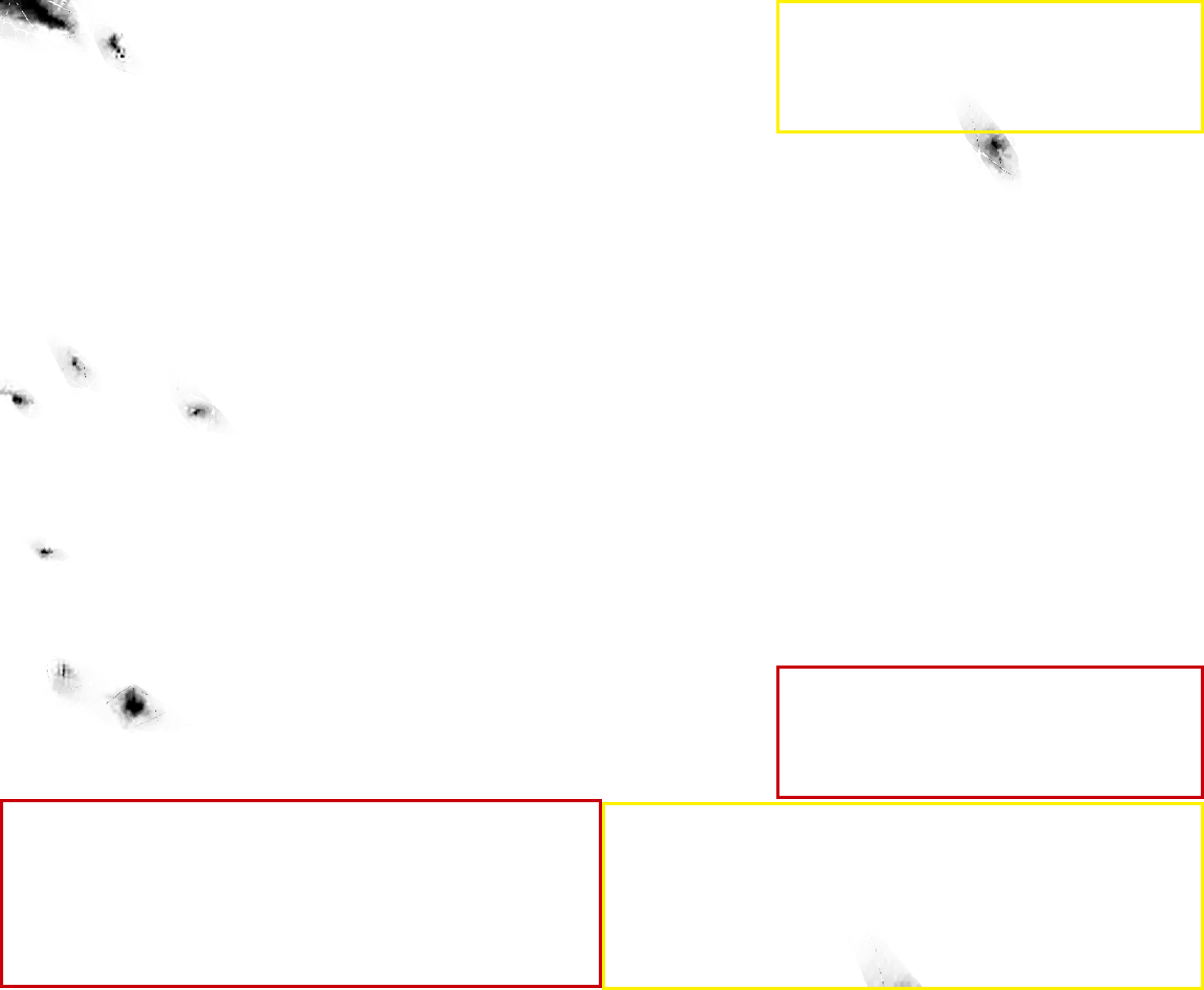} &
				\includegraphics[scale=0.15846]{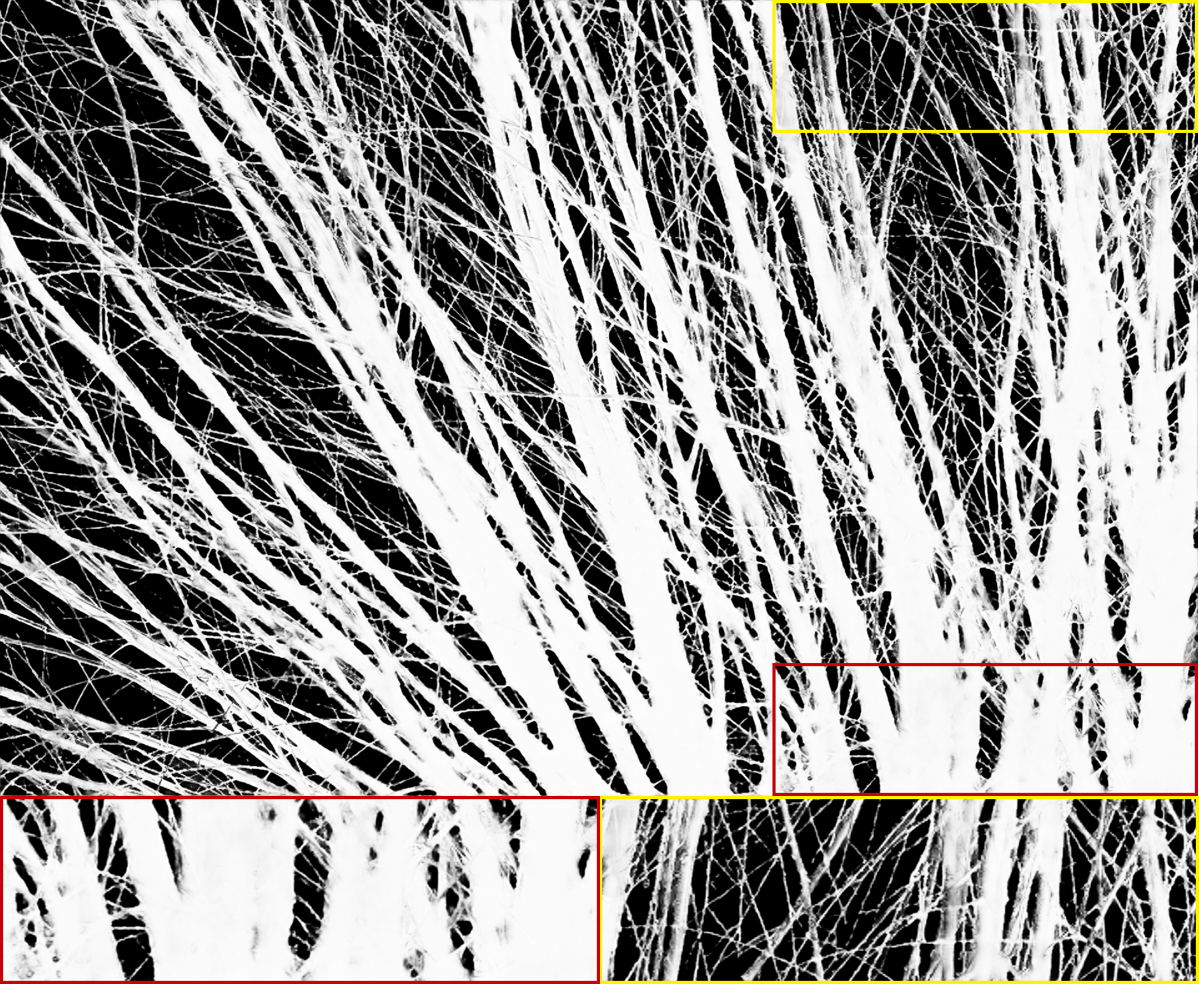} &
				\includegraphics[scale=0.1664]{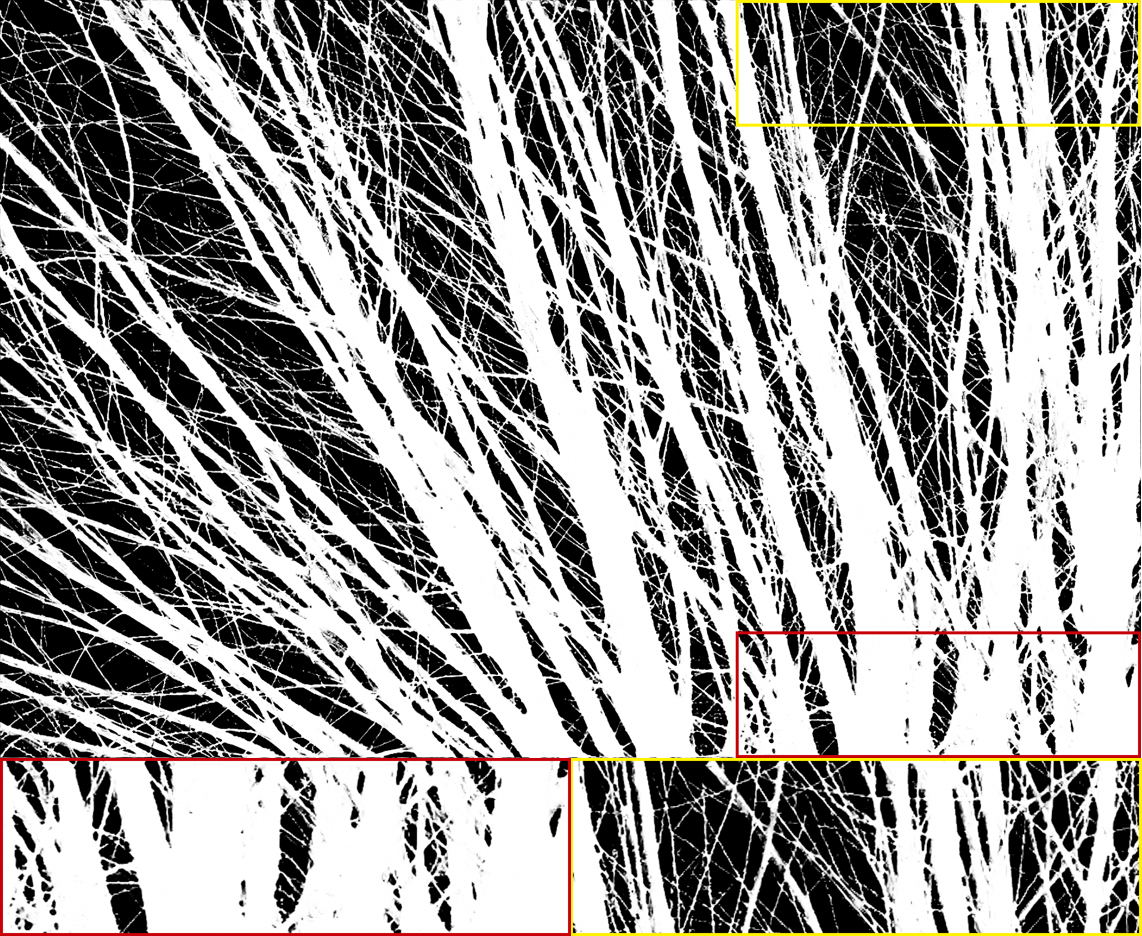} &
				\includegraphics[scale=0.16623]{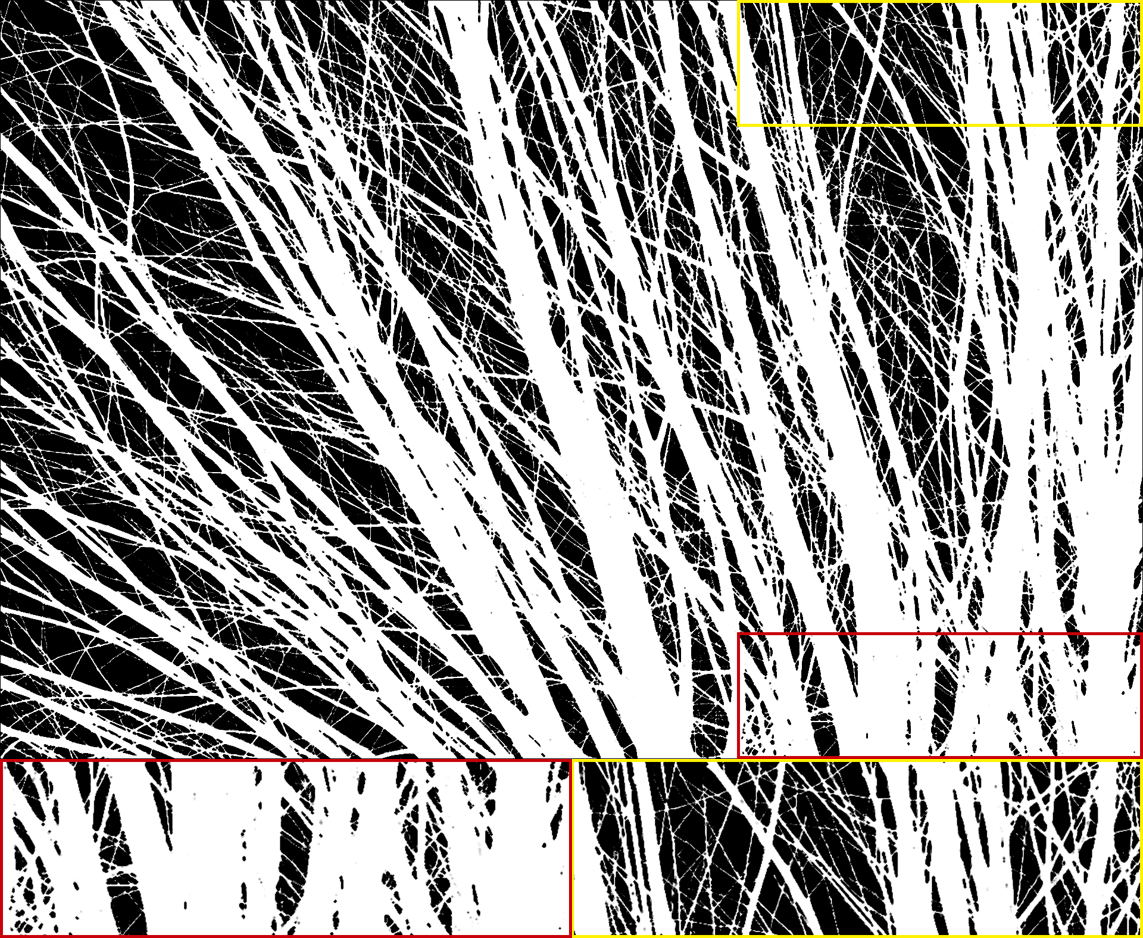} \\
				
				\includegraphics[scale=0.24297]{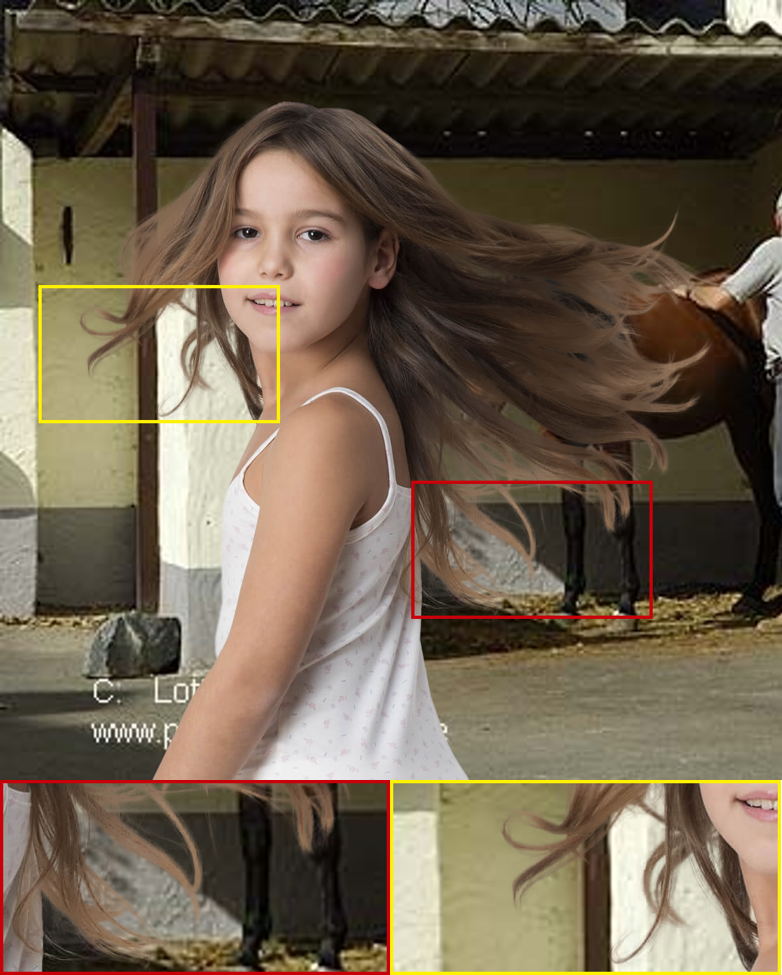} &
				\includegraphics[scale=0.24611]{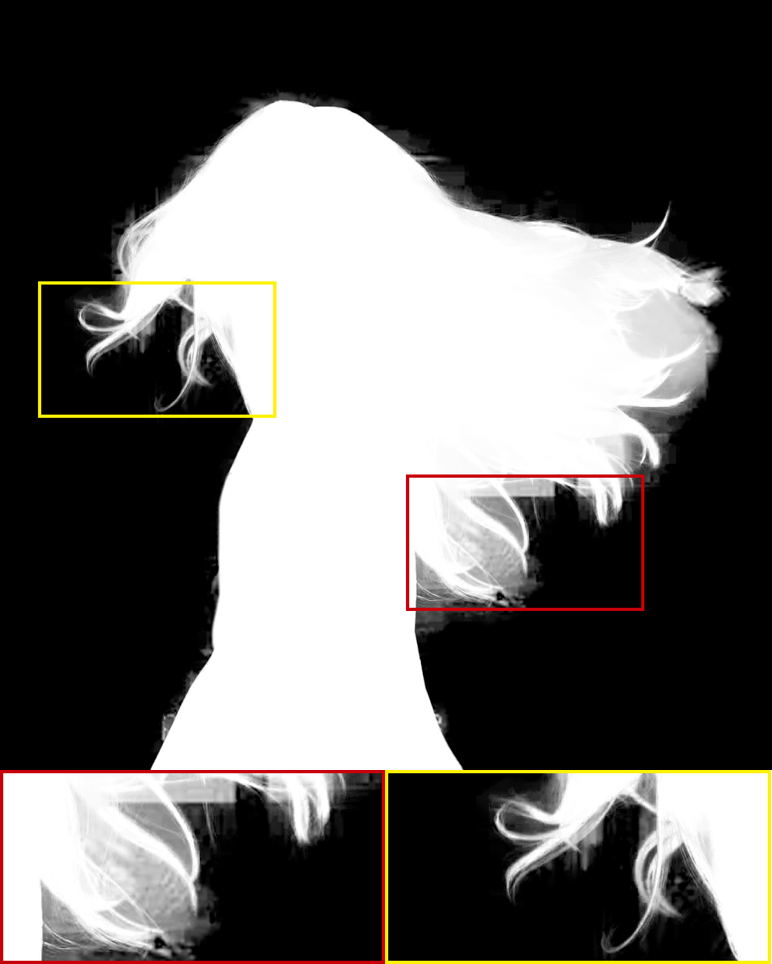} &
				\includegraphics[scale=0.22353]{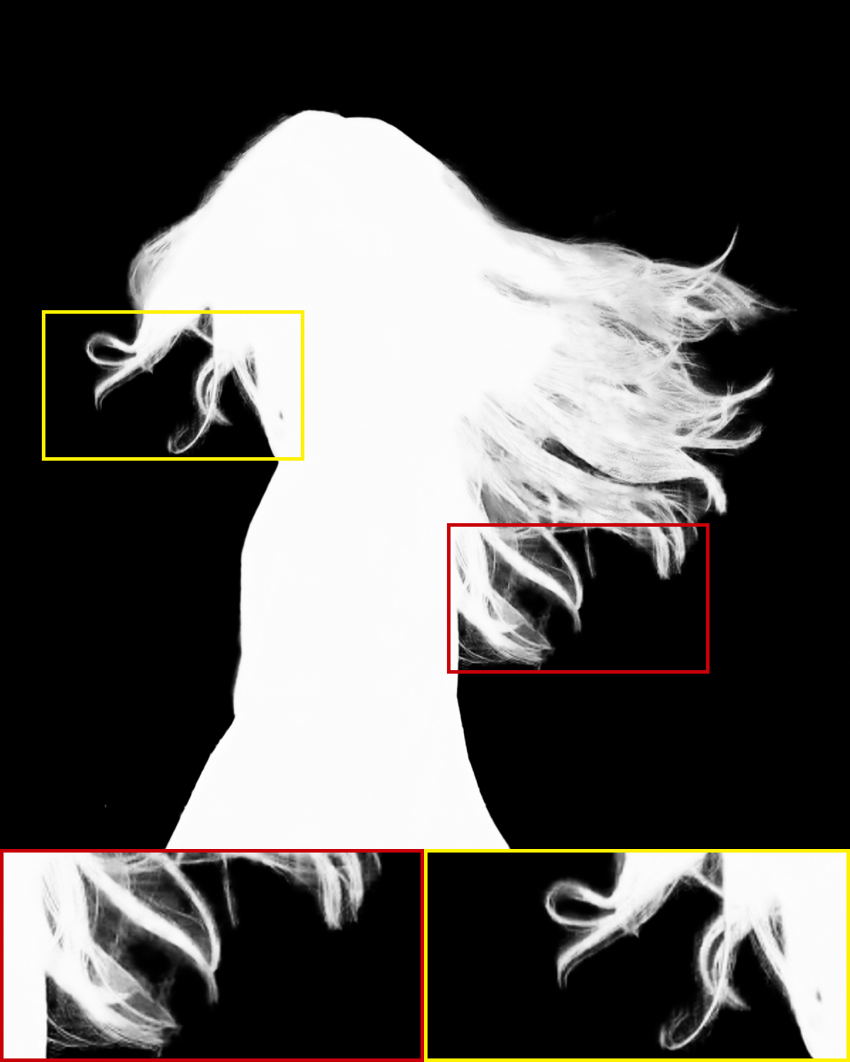} &
				\includegraphics[scale=0.21205]{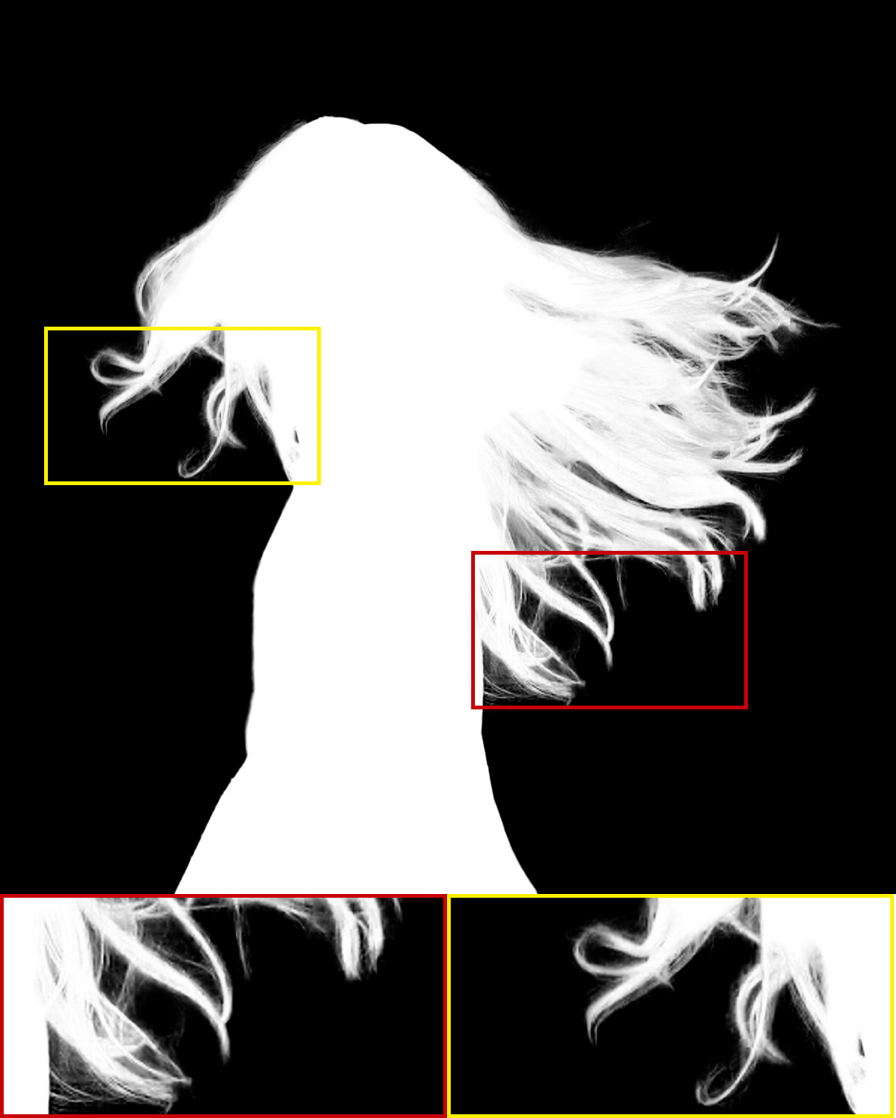} &
				\includegraphics[scale=0.2381]{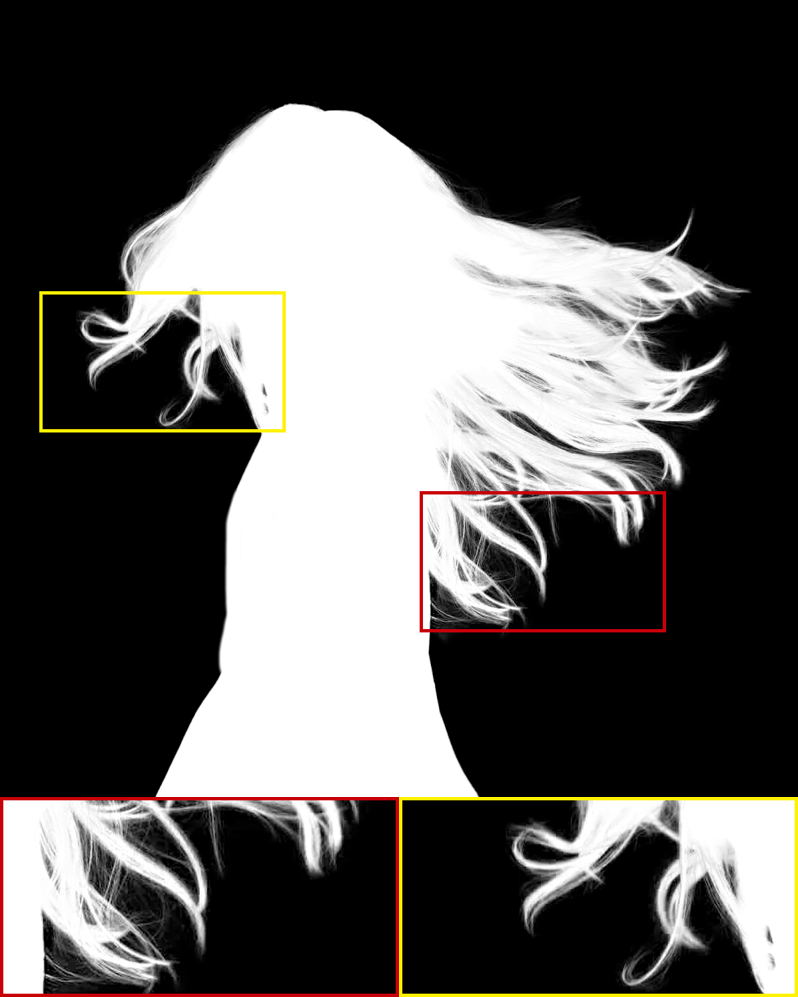} \\
				
				\includegraphics[scale=0.17273]{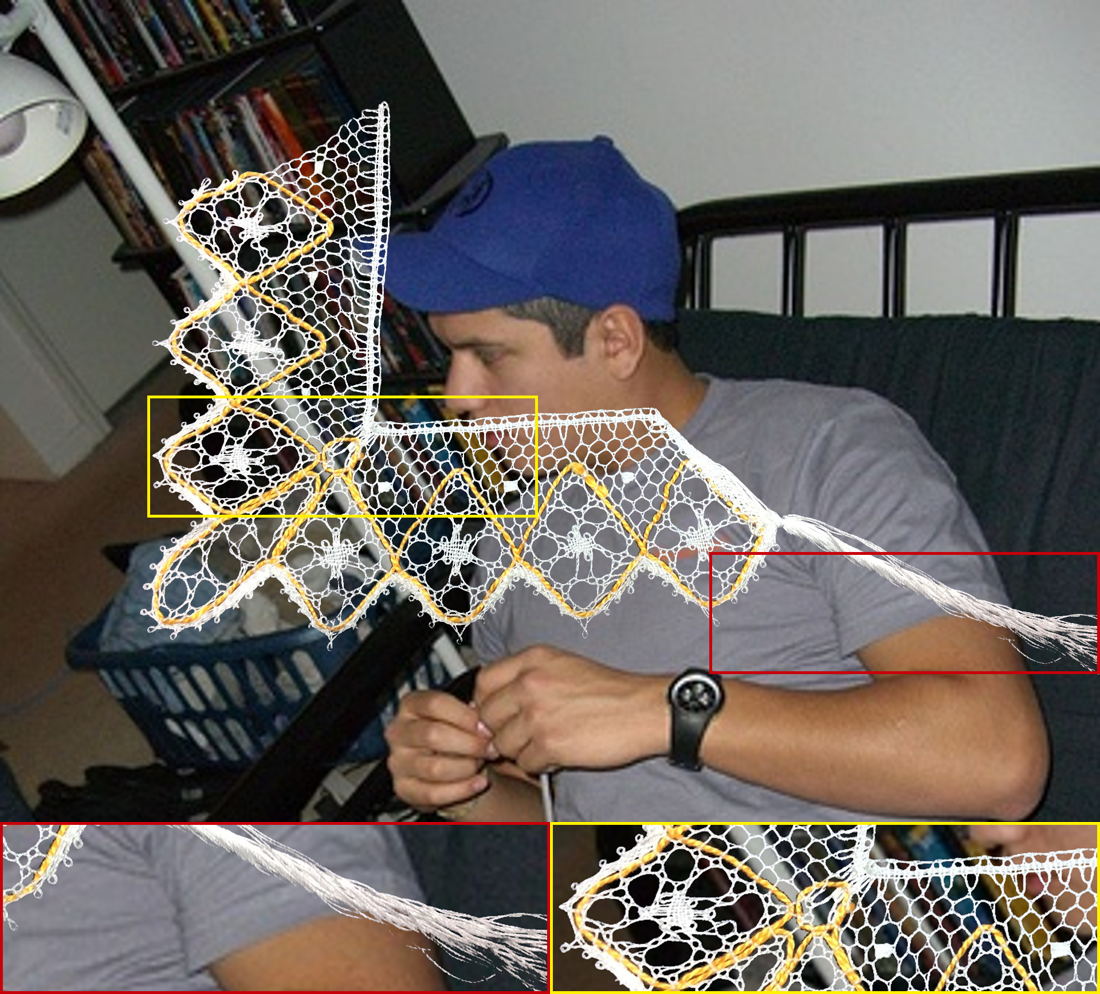} &
				\includegraphics[scale=0.1704]{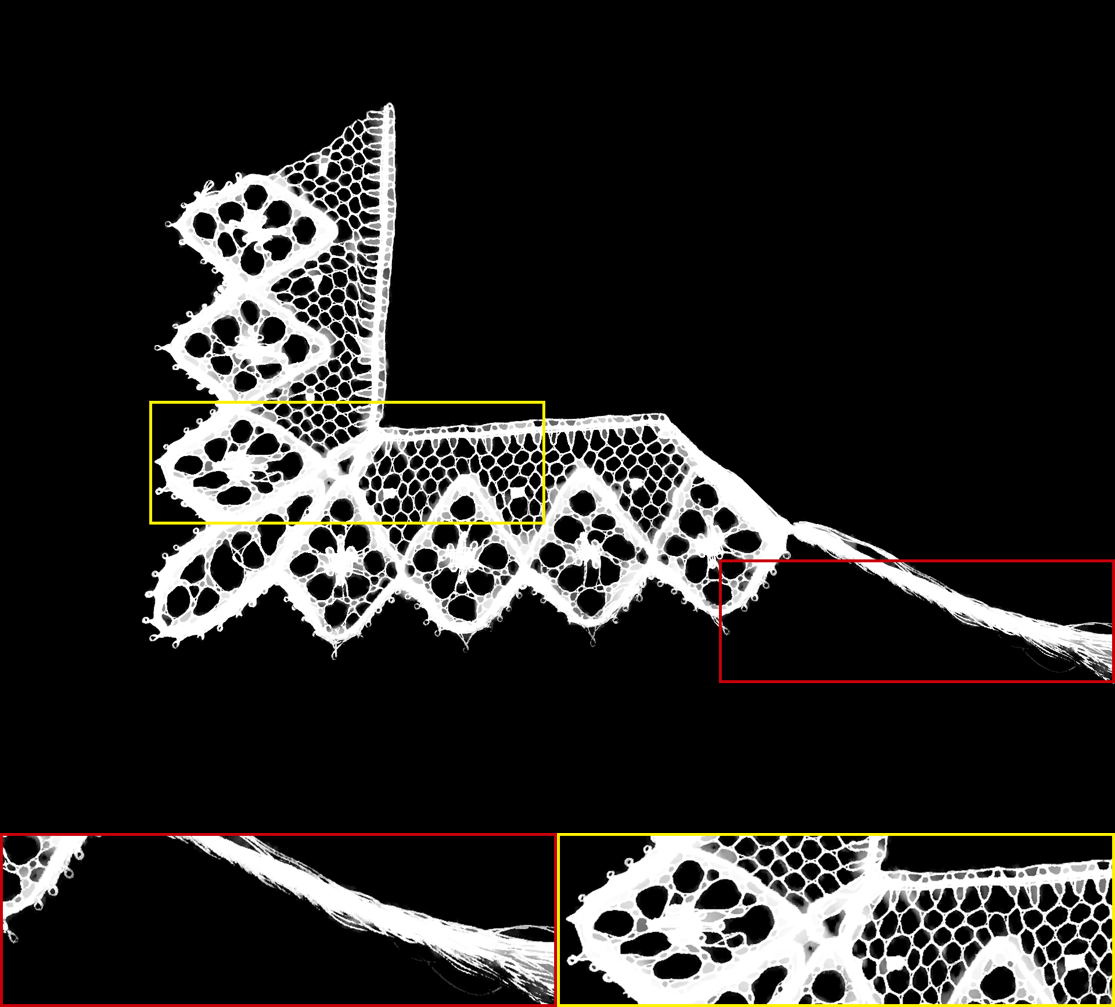} &
				\includegraphics[scale=0.10536]{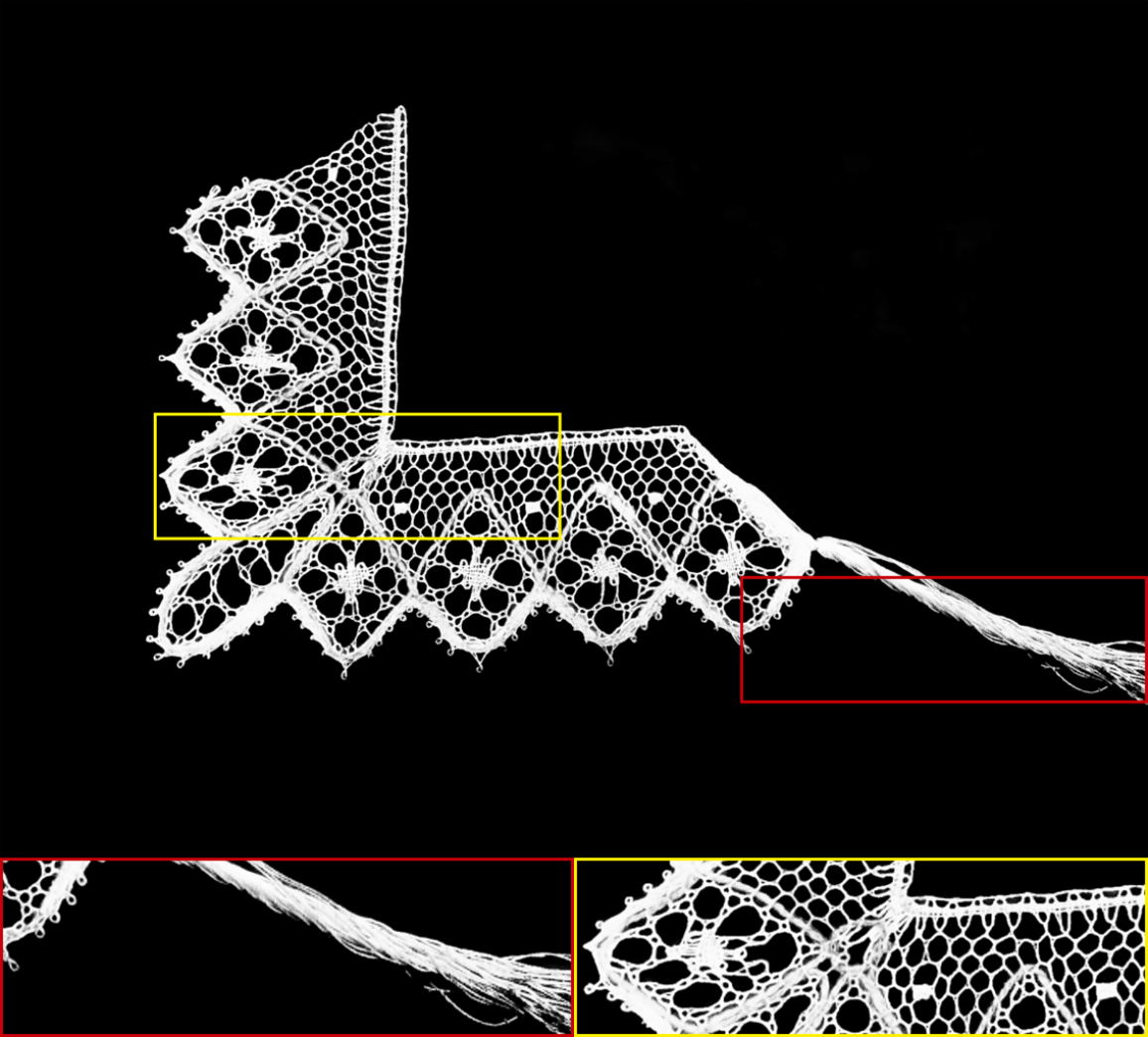} &
				\includegraphics[scale=0.1926]{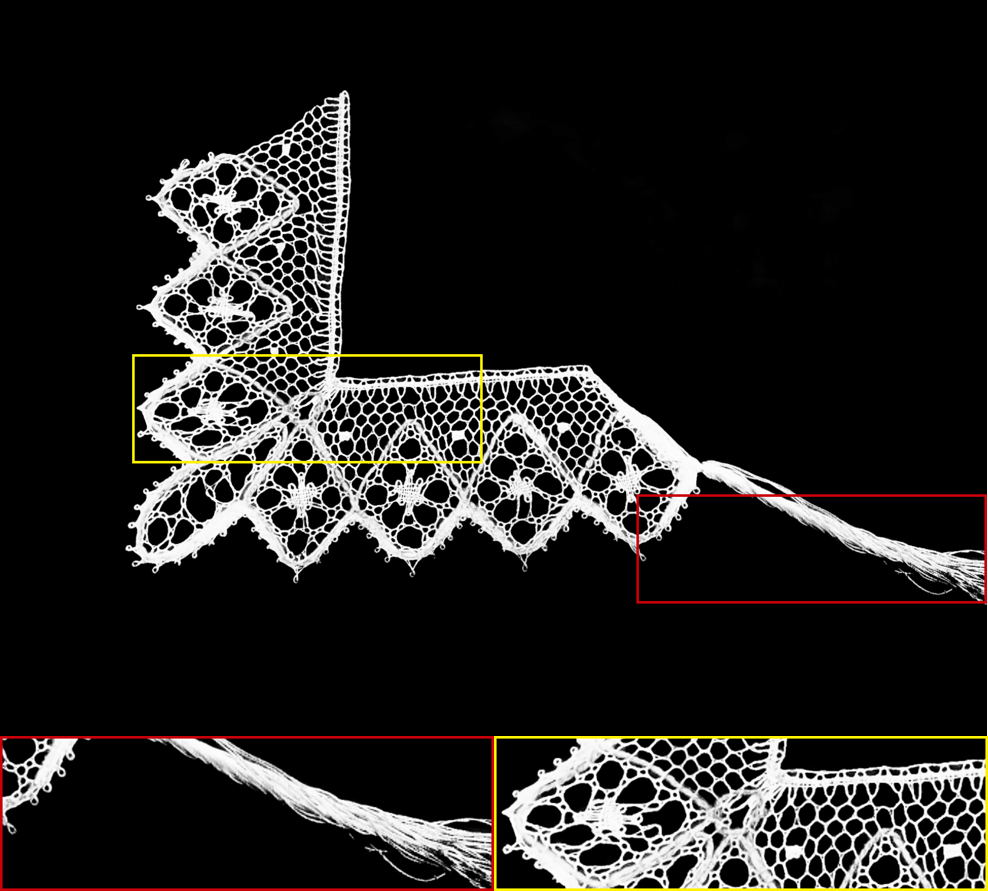} &
				\includegraphics[scale=0.1883]{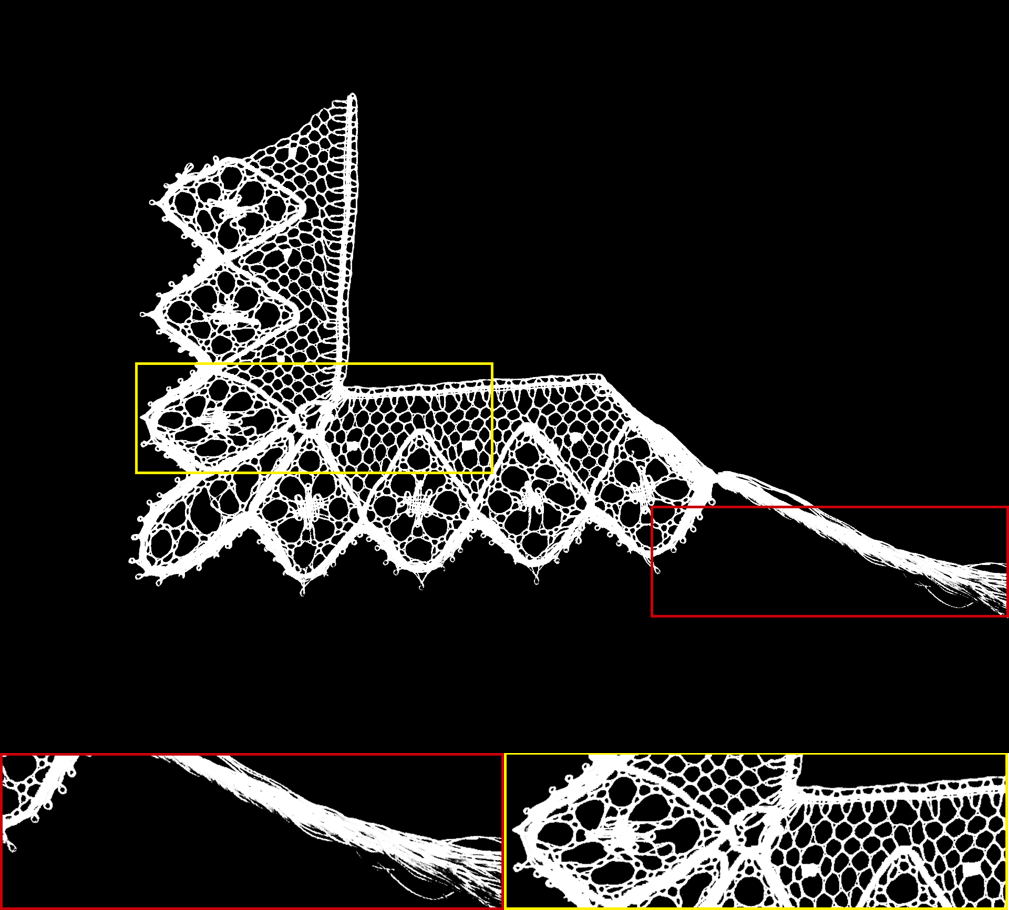} \\
				
				\includegraphics[scale=0.22144]{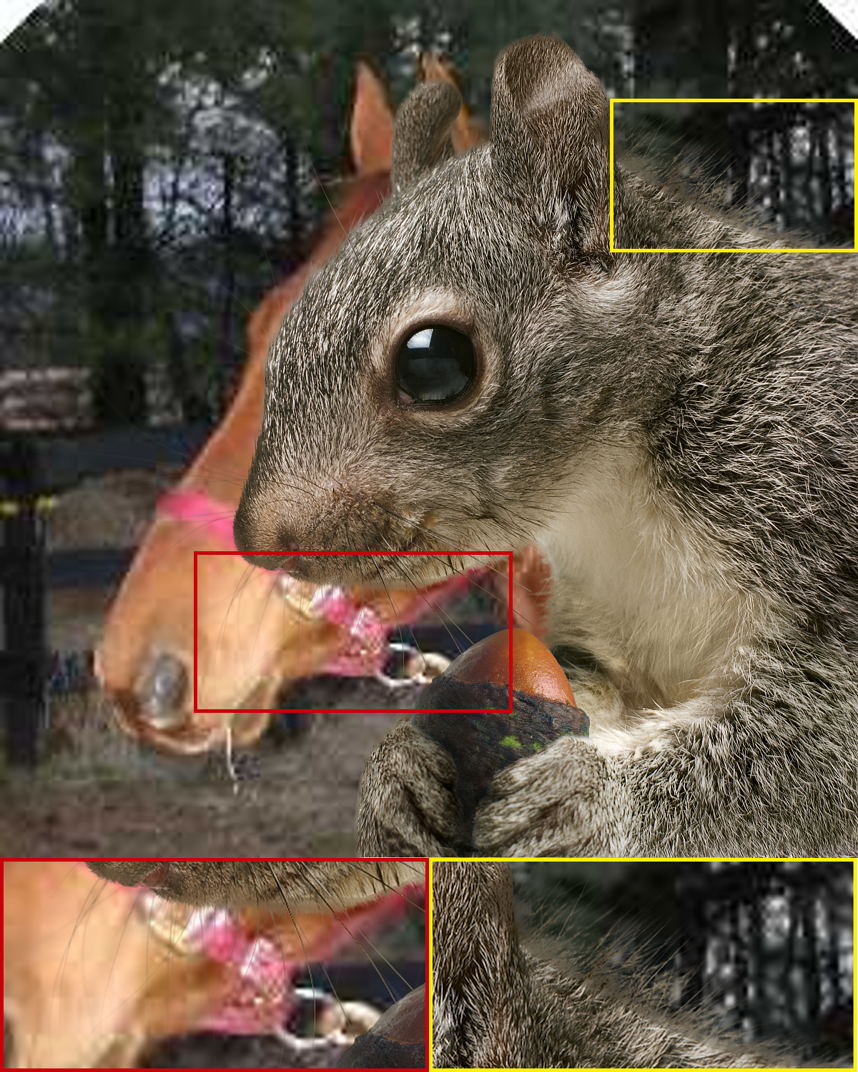} &
				\includegraphics[scale=0.07606]{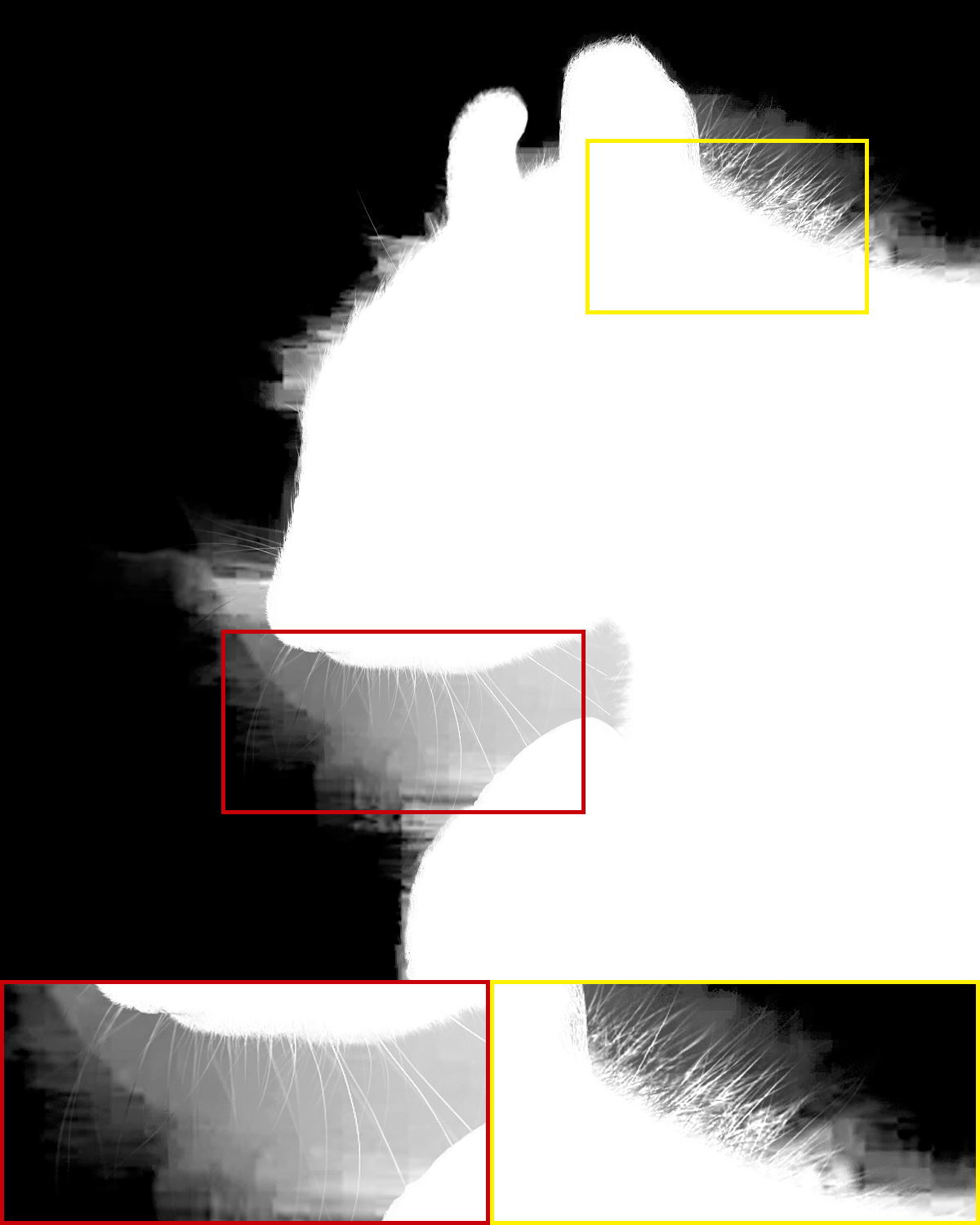} &
				\includegraphics[scale=0.2331]{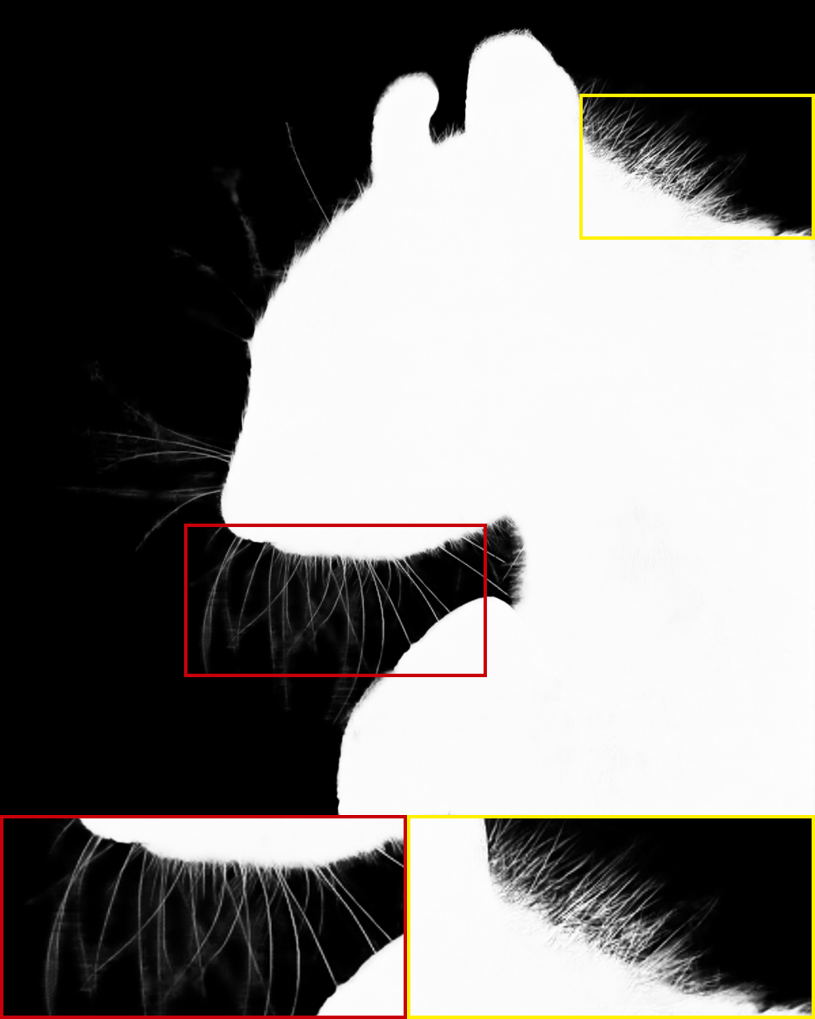} &
				\includegraphics[scale=0.2553]{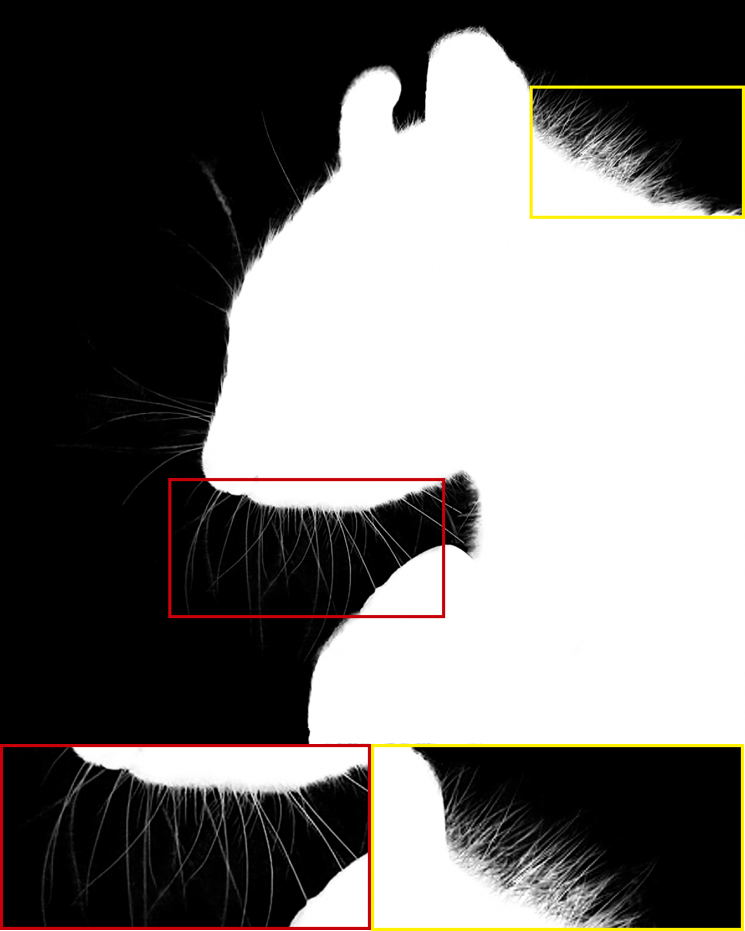} &
				\includegraphics[scale=0.20948]{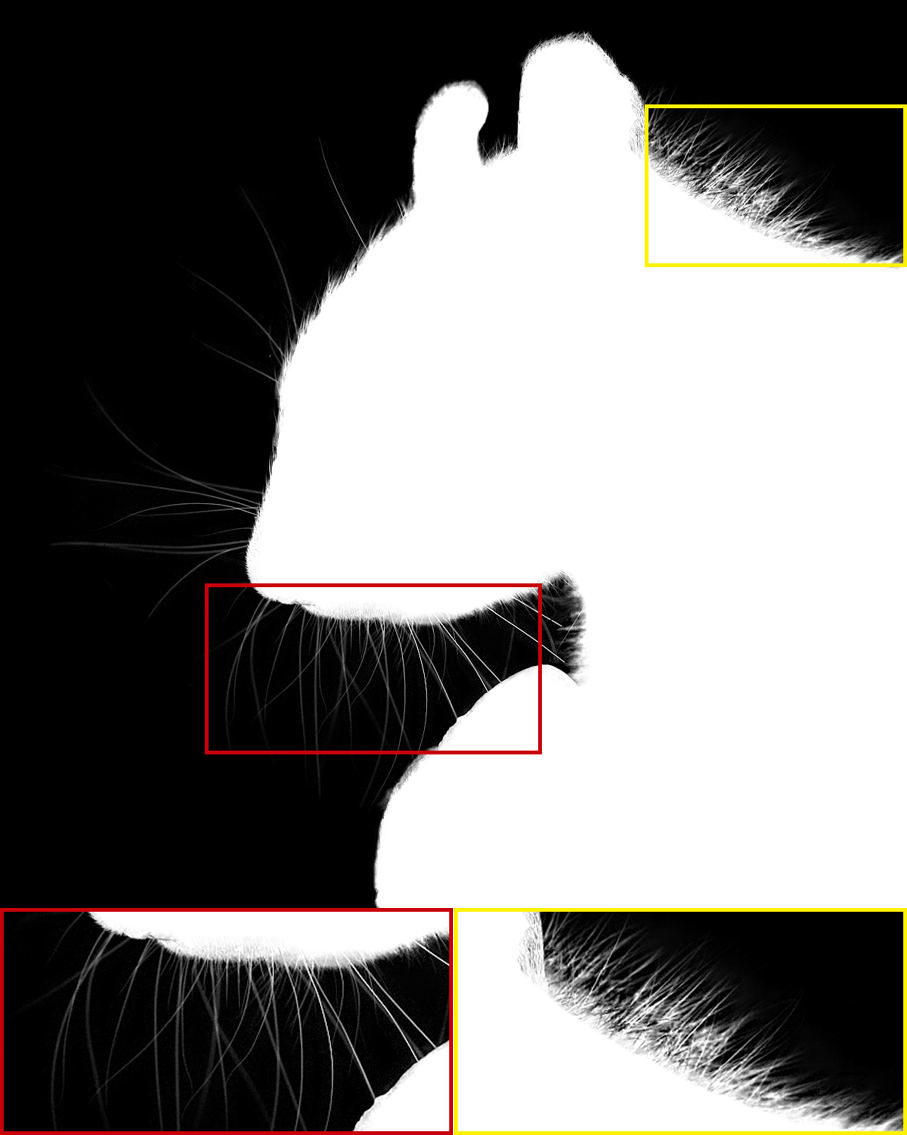} \\
				
				Input Image & Closed Form~\cite{Levin2007A} & HAttMatting~\cite{Qiao_2020_CVPR} & Ours & Ground Truth \\
		\end{tabular}}
		\caption{Comparison of the alpha mattes on the Distinctions-646 testing set~\cite{Qiao_2020_CVPR}. Some details and texture are zoomed in and placed under the images. }
		\label{fig:com_to_HAtt}
	\end{figure*}
	
	It is noted that only the Late Fusion~\cite{Zhang2019CVPR}, HAttMatting~\cite{Qiao_2020_CVPR}, and our MSIA-matte can achieve alpha mattes without trimaps in Table~\ref{tab:quantitative_adobe}. The quality of trimap has a substantial impact on the final alpha mattes, which has been proved in~\cite{Zhang2019CVPR}. Compared with trimap-based methods, we can obtain more striking results than most of them. Only the latest Context-aware~\cite{hou2019context} and IndexNet~\cite{hao2019indexnet} are better than our model. For trimap-free methods, we are second in the SAD metric to HAttMatting~\cite{Qiao_2020_CVPR} and get the same results on the MSE metric. The proposed MSIA-matte can achieve the best performance on the gradient and connectivity metrics. According to~\cite{rhemann2009perceptually}, the SAD and MSE metrics can represent pixel-wise accuracy, while the gradient and connectivity can indicate visual quality. Therefore, the comparisons with trimap-free methods suggest that our model can produce a better judgment of visual quality. This promotion can contribute to our multi-scale information assembly, which can synthetically integrate different-level foreground expressions to achieve the estimation of alpha mattes. From another side, our multi-scale foreground expression may attenuate the subject semantics of the foreground object to some extent and lead to some undersaturated pixels (e.g., some of the subject regions should be 255, but are only 252 or 253), which potentially reduces the pixel-wise accuracy. The visual results of some available methods are shown in Figure~\ref{fig:com_to_DIM}. From the first and second rows, we can see that the MSIA-matte can effectively capture subtle variations in opacity and correspond to visual quality promotion.
	
	\subsection{The Distinctions-646 dataset}
	\label{ssec:com_on_hatt}
	
	In this experiment, we train and evaluate the MSIA-matte model on the Distinctions-646 dataset~\cite{Qiao_2020_CVPR}. We also compare our model with traditional methods and some available matting networks, including Shared Matting~\cite{gastal2010shared}, Learning-Based~\cite{zheng2009learning}, Global Matting~\cite{Rhemann2011A}, ClosedForm~\cite{Levin2007A}, KNN Matting~\cite{Chen2013KNN}, Information Flow~\cite{Aksoy2017Designing}, DCNN~\cite{Cho2016Natural}, DIM~\cite{Xu2017Deep}, and HAttMatting~\cite{Qiao_2020_CVPR}. Table~\ref{tab:quantitative_hatt} reports the results of all these methods.
	
	Compared to Composition-1K~\cite{Xu2017Deep} dataset, the Distinctions-646~\cite{Qiao_2020_CVPR} has more kinds of foreground objects. Correspondingly, all metrics have a slight rise, but the entire conclusion is similar to the results of the Composition-1K testing set. Compared to existing matting methods, we can achieve state-of-the-art performance. Our model has the best performance in gradient and connectivity, which suggests that the information assembly can improve visual quality. Moreover, the visual quality of our method can also be proved in Figure~\ref{fig:com_to_HAtt}. We can adapt to continuous changes in the opacity and integrate multi-scale foreground information.
	\begin{table}[t]
		\small
		\centering
		\setlength{\tabcolsep}{1.2mm}{
			\begin{tabular}{l|ccccc}
				\hline
				Methods & SAD$\downarrow$ & MSE$\downarrow$  & Gradient$\downarrow$  & Connectivity$\downarrow$  \\
				\hline
				Shared Matting~\cite{gastal2010shared} & 119.56 & 0.026 & 129.61 & 114.37 \\
				Learning Based~\cite{zheng2009learning} & 105.04 & 0.021 & 94.16 & 110.41 \\
				Global Matting~\cite{Rhemann2011A} & 135.56 & 0.039 & 119.53 & 136.44 \\
				ClosedForm~\cite{Levin2007A} & 105.73 & 0.023 & 91.76 & 114.55 \\
				KNN Matting~\cite{Chen2013KNN} & 116.68 & 0.025 & 103.15 & 121.45 \\
				DCNN~\cite{Cho2016Natural} & 103.81 & 0.020 & 82.45 & 99.96 \\
				Information Flow~\cite{Aksoy2017Designing} & 78.89 & 0.016 & 58.72 & 80.47 \\
				DIM~\cite{Xu2017Deep} & 47.56 & 0.009 & 43.29 & 55.90 \\
				\hline
				\hline
				\rowcolor{mygray}
				HAttMatting~\cite{Qiao_2020_CVPR} & \textbf{48.98} & 0.009 & 41.57  & 49.93 \\
				\rowcolor{mygray}
				Ours & 49.51 & \textbf{0.009} & \textbf{39.70} & \textbf{46.68} \\
				\hline
		\end{tabular}}
		\caption{The quantitative results on the Distinctions-646 testing set. }
		\label{tab:quantitative_hatt}
	\end{table}
	
	\subsection{Ablation study}
	\label{ssec:ablation_study}
	\begin{table}[t]
		\small
		\centering
		\setlength{\tabcolsep}{1.6mm}{
			\begin{tabular}{ccc|cccc}
				\hline
				IniST & SedST & AI & SAD$\downarrow$ & MSE$\downarrow$  & Gradient$\downarrow$ & Connectivity$\downarrow$  \\
				\hline
				&   &   & 69.02 & 0.013 & 39.91 & 67.62 \\
				\checkmark &   &   & 64.97 & 0.012 & 35.65 & 61.73 \\
				\checkmark & \checkmark &   & 52.35 & 0.010 & 34.75 & 53.40 \\
				\checkmark & \checkmark & \checkmark & \textbf{47.86} & \textbf{0.007} & \textbf{28.61} & \textbf{43.39} \\
				\hline
				\hline
				&   &   & 64.73 & 0.016 & 71.34 & 72.48 \\
				\checkmark &   &   & 60.28 & 0.014 & 60.56 & 63.14 \\
				\checkmark & \checkmark &   & 57.47 & 0.012 & 45.34 & 60.03\\
				\checkmark & \checkmark & \checkmark & \textbf{49.51} & \textbf{0.009} & \textbf{39.70} & \textbf{46.68} \\
				\hline
		\end{tabular}}
		\caption{The quantitative results to compare the benefits of the proposed components. The ``IniST'', ``SedST'', and ``AI'' represent our initial and second phases of superficial traces branch and adaptive integration, respectively. The first four rows of results are evaluated on the Adobe Composition-1K testing dataset~\cite{Xu2017Deep}, and the last four results are based on the Distinctions-646~\cite{Qiao_2020_CVPR}. }
		\label{tab:quantitative_ablation}
	\end{table}
	
	We propose the superficial traces branch to approach low-level CNN features, and then combine different-level superficial traces with high-level semantics to integrate multi-scale foreground expression. In this section, we remove some modules to prove their benefits for the promotion of final alpha mattes. For concise description, here we denote the initial two convolutional layers of the superficial traces branch as IniST, and the later four cascaded convolutional layers are represented as SedST. For the baseline comparison, we remove IniST, SedST, and Adaptive Integration (AI), and use direct concatenation between low-level CNN features and upsampled ASPP features (the upsampling factor is 4) to produce alpha mattes. And then, we add IniST to the baseline and downsample the features from IniST as secondary superficial traces to perform following information assembly with direct concatenation. Finally, we add IniST and SedST to the baseline and only replace adaptive integration (AI) with direct concatenation to execute the decoder stage. The quantitative result of the above three models is summarized in Table~\ref{tab:quantitative_ablation}.
	
	The first four rows of Table~\ref{tab:quantitative_ablation} are evaluated on the Adobe Composition-1K testing set, while the last four results are conducted on the Distinctions-646 dataset. Compared to the baseline network, the IniST can bring an obvious improvement for the alpha mattes. This comparison can prove the ability of the IniST to filter low-level CNN features, which are relatively primitive for information assembly. Correspondingly, the SedST can further condense potential middle-levels semantics (such as hands, leaves) and preserve some primary textures and details, and the promotion of the results are also reflected on the four metrics (the SAD, MSE, Gradient and Connectivity dropped by 12.62, 0.002, 0.9 and 8.33, respectively). The subtle promotion on the MSE and Gradient two metrics are mainly due to the versatility of the SedST: almost all the images contain fine-grained textures and details, but not such middle-level foreground expression. Therefore, we import our adaptive integration to achieve information assembly effectively. The four metrics are further improved, as reported in the bold results of Table~\ref{tab:quantitative_ablation}.
	
	\subsection{Results on natural images}
	\label{ssec:results_on_natural}
	\begin{figure}[t]
		\centering
		\setlength{\tabcolsep}{1pt}\small{
			\begin{tabular}{ccc}
				\includegraphics[scale=0.16276]{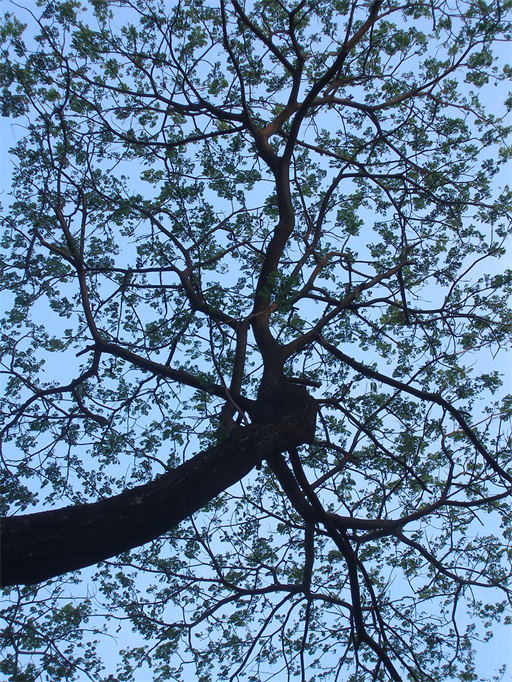} &
				\includegraphics[scale=0.14453]{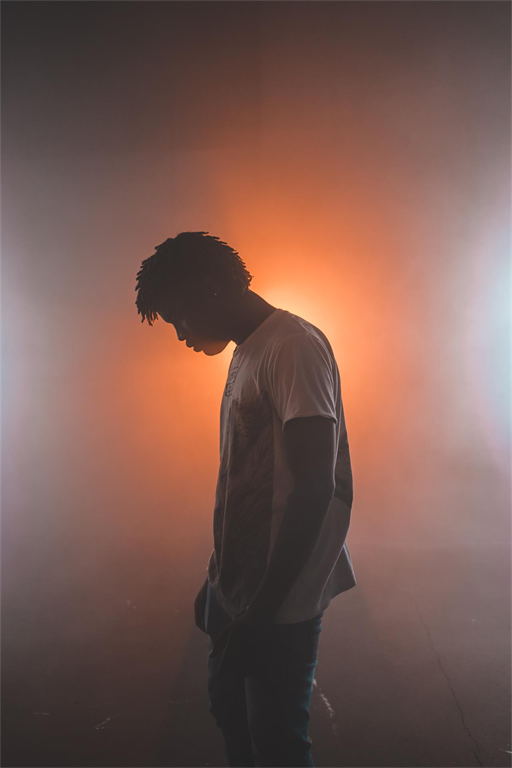} &
				\includegraphics[scale=0.2168]{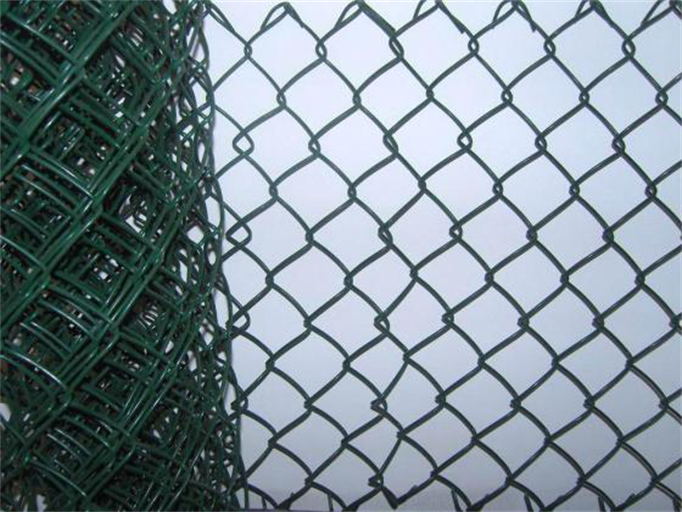} \\
				
				\includegraphics[scale=0.122]{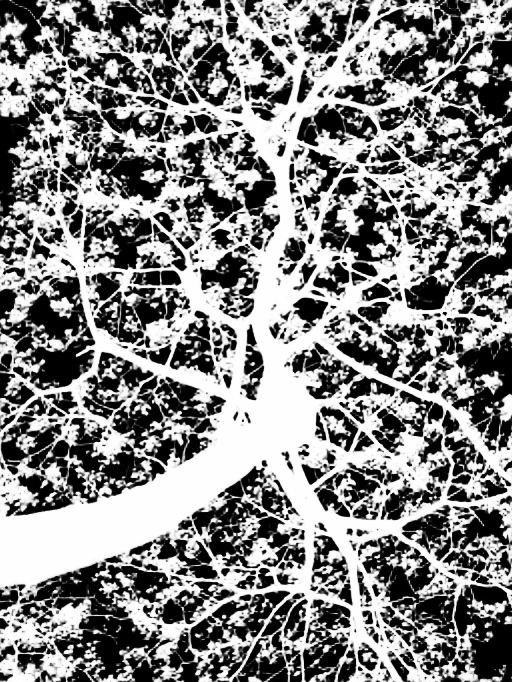} &
				\includegraphics[scale=0.1083]{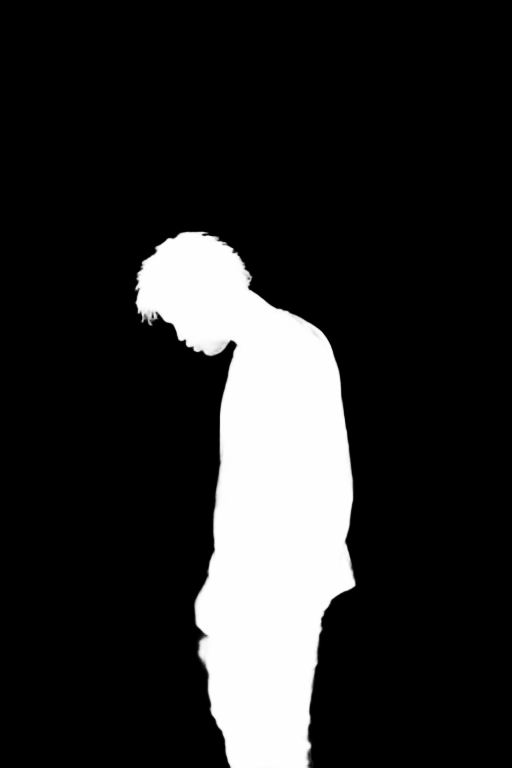} &
				\includegraphics[scale=0.1625]{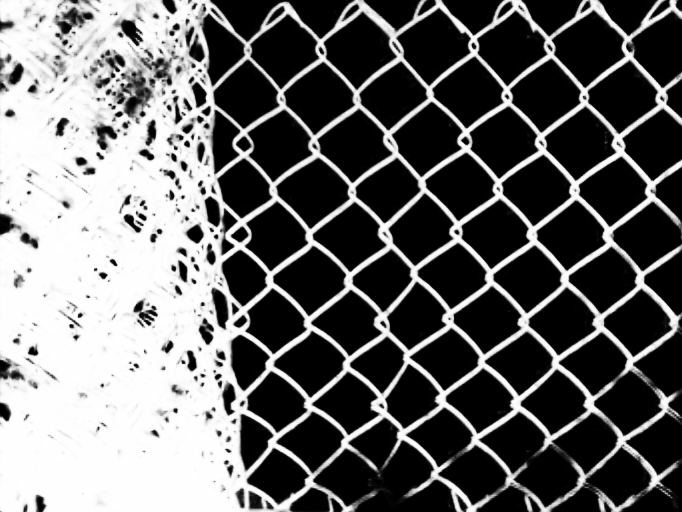} \\
		\end{tabular}}
		\caption{The results of natural images. The first row is the input images, and the second row is the alpha mattes produced by our method. }
		\label{fig:visual_nature}
	\end{figure}
	The results of our model on natural images are shown in Fig.~\ref{fig:visual_nature}. Our MSIA-matte can effectively deal with different-level foreground information (the thick trunk and the thin branches of the first column, the various opacity of the net). Besides, we can even adapt to the background of continuous changes and predict a high-quality alpha matte (the boy in the second column). 
	
	\begin{figure}[t]
		\centering
		\setlength{\tabcolsep}{1pt}\small{
			\begin{tabular}{ccc}
				\includegraphics[scale=0.179]{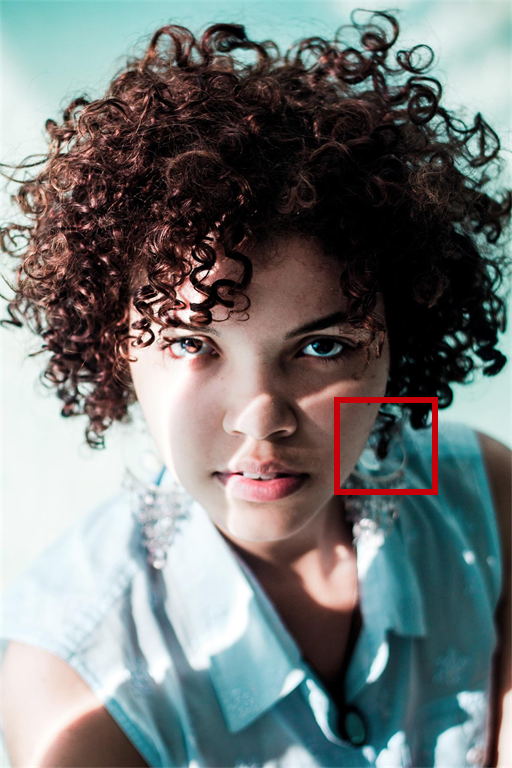} &
				\includegraphics[scale=0.2688]{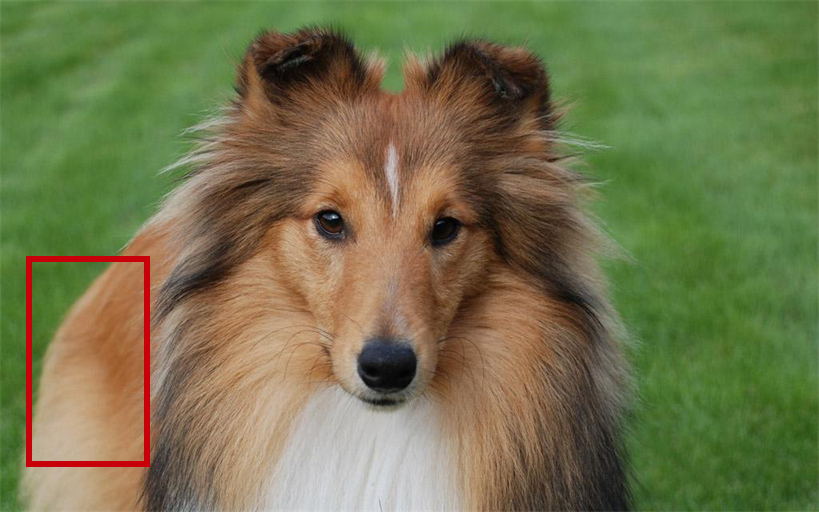} \\
				
				\includegraphics[scale=0.134]{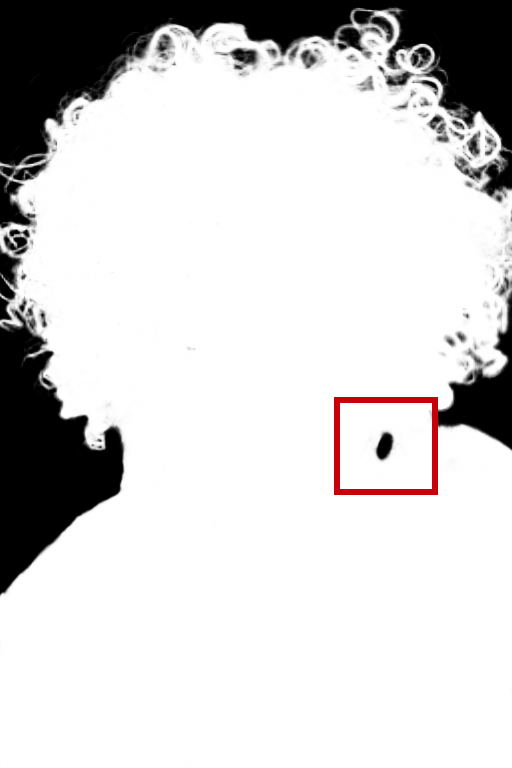} &
				\includegraphics[scale=0.2014]{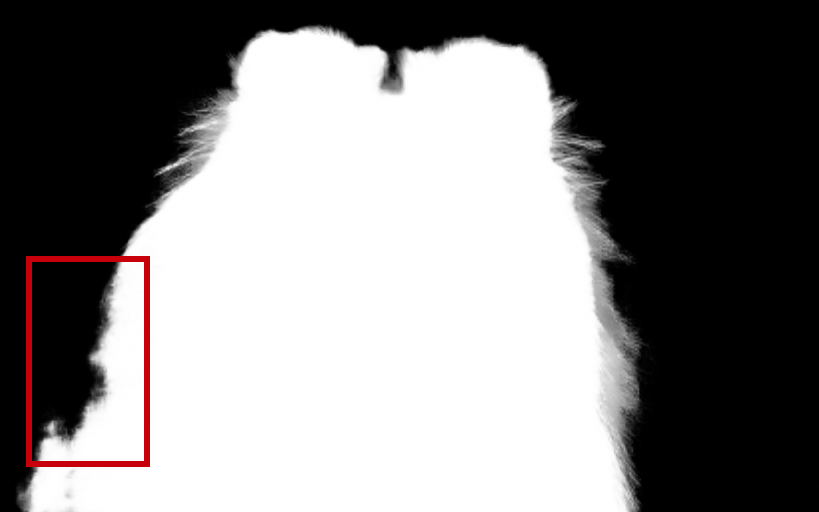} \\
		\end{tabular}}
		\caption{The failure case of our model. The first row is the input images, and the second row is the alpha mattes produced by our method. Based on our analysis in section~\ref{ssec:com_on_dim}, the proposed multi-scale information assembly can possibly attenuate the impact of the advanced semantics and produce some subtle region missing in natural images (the red box). }
		\label{fig:failure_case}
	\end{figure}
	
	\subsection{Limitations analysis}
	\label{ssec:limitations_analysis}
	There are also some failure cases of our method, as shown in Figure~\ref{fig:failure_case}. We utilize different-level foreground expression to regress the alpha mattes, which possibly attenuate the significance of the subject semantics thus can sometimes result in subtle region missing. Besides, there are two additional possible limitations to our method: (1) just like most deep-learning-based matting networks, our model is fully trained on composite datasets, and the illumination, reflection, emptiness, etc. in natural images may have a huge influence on the alpha mattes. (2) the proposed MSIA-matte can predict high-quality alpha mattes without trimaps, but our model may lose efficacy if the input images are extremely complicated (the region of interest is not commonsensible foreground or there are multiple co-existing foreground objects), and the trimaps-based methods may work better in such situations.
	
	\section{Conclusion}
	\label{sec:conclusion}
	
	In this paper, we propose a multi-scale information assembly model (MSIA-matte) to extract and integrate different foreground expression from single RGB input images. The multi-scale information assembly strategy can effectively integrate the advanced semantics from ASPP and the superficial traces from low-level CNN features. 
	
	For future work, we intend to explore the possibility of the foreground information assembly in video matting. The temporary and spatial information may be considered to further enhance the correlation of multi-scale foreground expression between different frames.
	
	\section*{Acknowledgement}
	\label{acknowledgement}
	This work was supported in part by the National Natural Science Foundation of China under Grant 91748104, Grant 61972067, Grant 61632006, Grant U1811463, Grant U1908214, Grant 61751203, in part by the National Key Research and Development Program of China under Grant 2018AAA0102003, Grant 2018YFC0910506. 
	
	\bibliographystyle{eg-alpha-doi} 
	\bibliography{egbib}       

\newcommand{\etalchar}[1]{$^{#1}$}
\begin{thebibliography}{\uppercase{GPAM{\etalchar{*}}14}}

\bibitem[AAP17]{Aksoy2017Designing}
\textsc{{Aksoy} Y., {Aydin} T.~O., {Pollefeys} M.}:
\newblock Designing effective inter-pixel information flow for natural image
  matting.
\newblock In \emph{Proceedings of the IEEE/CVF Conference on Computer Vision
  and Pattern Recognition (CVPR'17)} (2017), pp.~228--236.

\bibitem[AOP{\etalchar{*}}18]{sss}
\textsc{Aksoy Y., Oh T.-H., Paris S., Pollefeys M., Matusik W.}:
\newblock Semantic soft segmentation.
\newblock \emph{ACM Transactions on Graphics 37}, 4 (2018).

\bibitem[CGX{\etalchar{*}}18]{Chen2018SHM}
\textsc{Chen Q., Ge T., Xu Y., Zhang Z., Yang X., Gai K.}:
\newblock Semantic human matting.
\newblock In \emph{Proceedings of the ACM International Conference on
  Multimedia (MM'18)} (2018), p.~618–626.

\bibitem[CKTK17]{Cho2016Automatic}
\textsc{Cho D., Kim S., Tai Y.~W., Kweon I.~S.}:
\newblock Automatic trimap generation and consistent matting for light-field
  images.
\newblock \emph{IEEE Transactions on Pattern Analysis and Machine Intelligence
  39}, 8 (2017), 1504--1517.

\bibitem[CLT13]{Chen2013KNN}
\textsc{{Chen} Q., {Li} D., {Tang} C.}:
\newblock Knn matting.
\newblock \emph{IEEE Transactions on Pattern Analysis and Machine Intelligence
  35}, 9 (2013), 2175--2188.

\bibitem[CPK{\etalchar{*}}18]{Chen2018DeepLab}
\textsc{{Chen} L., {Papandreou} G., {Kokkinos} I., {Murphy} K., {Yuille}
  A.~L.}:
\newblock Deeplab: Semantic image segmentation with deep convolutional nets,
  atrous convolution, and fully connected crfs.
\newblock \emph{IEEE Transactions on Pattern Analysis and Machine Intelligence
  40}, 4 (2018), 834--848.

\bibitem[CTK16]{Cho2016Natural}
\textsc{Cho D., Tai Y.~W., Kweon I.}:
\newblock Natural image matting using deep convolutional neural networks.
\newblock In \emph{Proceedings of the European Conference on Computer Vision
  (ECCV'16)} (2016), pp.~626--643.

\bibitem[CZF{\etalchar{*}}19]{cai2019disentangled}
\textsc{{Cai} S., {Zhang} X., {Fan} H., {Huang} H., {Liu} J., {Liu} J., {Liu}
  J., {Wang} J., {Sun} J.}:
\newblock Disentangled image matting.
\newblock In \emph{Proceedings of the International Conference on Computer
  Vision (ICCV'19)} (2019), pp.~8818--8827.

\bibitem[FLZ]{Feng2016A}
\textsc{Feng X., Liang X., Zhang Z.}:
\newblock A cluster sampling method for image matting via sparse coding.
\newblock In \emph{Proceedings of the European Conference on Computer Vision
  (ECCV'16)}, pp.~204--219.

\bibitem[GO10]{gastal2010shared}
\textsc{Gastal E. S.~L., Oliveira M.~M.}:
\newblock Shared sampling for real-time alpha matting.
\newblock \emph{Computer Graphics Forum 29}, 2 (2010), 575--584.

\bibitem[GPAM{\etalchar{*}}14]{Goodfellow2014Generative}
\textsc{Goodfellow I.~J., Pouget-Abadie J., Mirza M., Xu B., Warde-Farley D.,
  Ozair S., Courville A., Bengio Y.}:
\newblock Generative adversarial nets.
\newblock In \emph{Proceedings of the International Conference on Neural
  Information Processing Systems (NeurIPS'14)} (2014), p.~2672–2680.

\bibitem[GSAW05]{Grady2005Random}
\textsc{Grady L., Schiwietz T., Aharon S., Westermann R.}:
\newblock Random walks for interactive alpha-matting.
\newblock In \emph{Proceedings of the Visualization, Imaging, and Image
  Processing (VIIP'05)} (2005), pp.~423--429.

\bibitem[HL19]{hou2019context}
\textsc{{Hou} Q., {Liu} F.}:
\newblock Context-aware image matting for simultaneous foreground and alpha
  estimation.
\newblock In \emph{Proceedings of the International Conference on Computer
  Vision (ICCV'19)} (2019), pp.~4129--4138.

\bibitem[HRR{\etalchar{*}}11]{Rhemann2011A}
\textsc{{He} K., {Rhemann} C., {Rother} C., {Tang} X., {Sun} J.}:
\newblock A global sampling method for alpha matting.
\newblock In \emph{Proceedings of the IEEE/CVF Conference on Computer Vision
  and Pattern Recognition (CVPR'11)} (2011), pp.~2049--2056.

\bibitem[HZRS16]{Khe2016Resnet}
\textsc{{He} K., {Zhang} X., {Ren} S., {Sun} J.}:
\newblock Deep residual learning for image recognition.
\newblock In \emph{Proceedings of the IEEE/CVF Conference on Computer Vision
  and Pattern Recognition (CVPR'16)} (2016), pp.~770--778.

\bibitem[KEE15]{Karacan2015Image}
\textsc{{Karacan} L., {Erdem} A., {Erdem} E.}:
\newblock Image matting with kl-divergence based sparse sampling.
\newblock In \emph{Proceedings of the International Conference on Computer
  Vision (ICCV'15)} (2015), pp.~424--432.

\bibitem[LAS18]{lutz2018alphagan}
\textsc{Lutz S., Amplianitis K., Smolic A.}:
\newblock Alphagan: Generative adversarial networks for natural image matting.
\newblock 259.

\bibitem[LDSX19]{hao2019indexnet}
\textsc{{Lu} H., {Dai} Y., {Shen} C., {Xu} S.}:
\newblock Indices matter: Learning to index for deep image matting.
\newblock In \emph{Proceedings of the International Conference on Computer
  Vision (ICCV'19)} (2019), pp.~3265--3274.

\bibitem[LL20]{li2020natural}
\textsc{{Li} Y., {Lu} H.}:
\newblock Natural image matting via guided contextual attention.
\newblock In \emph{AAAI} (2020).

\bibitem[LLW07]{Levin2007A}
\textsc{Levin A., Lischinski D., Weiss Y.}:
\newblock A closed-form solution to natural image matting.
\newblock \emph{IEEE Transactions on Pattern Analysis and Machine Intelligence
  30}, 2 (2007), 228--242.

\bibitem[LRB15]{Wei2015ParseNet}
\textsc{Liu W., Rabinovich A., Berg A.~C.}:
\newblock Parsenet: Looking wider to see better.
\newblock \emph{arXiv preprint arXiv:1506.04579} (2015).

\bibitem[LRL08]{Levin2008Spectral}
\textsc{{Levin} A., {Rav-Acha} A., {Lischinski} D.}:
\newblock Spectral matting.
\newblock \emph{IEEE Transactions on Pattern Analysis and Machine Intelligence
  30}, 10 (2008), 1699--1712.

\bibitem[LSD15]{JLong2015fcnn}
\textsc{{Long} J., {Shelhamer} E., {Darrell} T.}:
\newblock Fully convolutional networks for semantic segmentation.
\newblock In \emph{Proceedings of the IEEE/CVF Conference on Computer Vision
  and Pattern Recognition (CVPR'15)} (2015), pp.~3431--3440.

\bibitem[LW11]{Lee2011Nonlocal}
\textsc{Lee P., Wu Y.}:
\newblock Nonlocal matting.
\newblock In \emph{Proceedings of the IEEE/CVF Conference on Computer Vision
  and Pattern Recognition (CVPR'11)} (2011), pp.~2193--2200.

\bibitem[QLY{\etalchar{*}}20]{Qiao_2020_CVPR}
\textsc{Qiao Y., Liu Y., Yang X., Zhou D., Xu M., Zhang Q., Wei X.}:
\newblock Attention-guided hierarchical structure aggregation for image
  matting.
\newblock In \emph{Proceedings of the IEEE/CVF Conference on Computer Vision
  and Pattern Recognition (CVPR'20)} (2020).

\bibitem[RRW{\etalchar{*}}09]{rhemann2009perceptually}
\textsc{Rhemann C., Rother C., Wang J., Gelautz M., Kohli P., Rott P.}:
\newblock A perceptually motivated online benchmark for image matting.
\newblock In \emph{Proceedings of the IEEE/CVF Conference on Computer Vision
  and Pattern Recognition (CVPR'09)} (2009), pp.~1826--1833.

\bibitem[SJTS04]{Sun2004Poisson}
\textsc{Sun J., Jia J., Tang C.~K., Shum H.~Y.}:
\newblock Poisson matting.
\newblock \emph{ACM Transactions on Graphics 23}, 3 (2004), 315--321.

\bibitem[SRPC13a]{Shahrian2013Improving}
\textsc{{Shahrian} E., {Rajan} D., {Price} B., {Cohen} S.}:
\newblock Improving image matting using comprehensive sampling sets.
\newblock In \emph{Proceedings of the IEEE/CVF Conference on Computer Vision
  and Pattern Recognition (CVPR'13)} (2013), pp.~636--643.

\bibitem[SRPC13b]{Shagrian2013}
\textsc{{Shahrian} E., {Rajan} D., {Price} B., {Cohen} S.}:
\newblock Improving image matting using comprehensive sampling sets.
\newblock In \emph{Proceedings of the IEEE/CVF Conference on Computer Vision
  and Pattern Recognition (CVPR'13)} (2013), pp.~636--643.

\bibitem[SZ15]{simonyan2014deep}
\textsc{Simonyan K., Zisserman A.}:
\newblock Very deep convolutional networks for large-scale image recognition.
\newblock In \emph{Proceedings of the International Conference on Learning
  Representations (ICLR'15)} (2015).

\bibitem[TAO{\etalchar{*}}19]{Tang_2019_CVPR}
\textsc{{Tang} J., {Aksoy} Y., {Oztireli} C., {Gross} M., {Aydin} T.~O.}:
\newblock Learning-based sampling for natural image matting.
\newblock In \emph{Proceedings of the IEEE/CVF Conference on Computer Vision
  and Pattern Recognition (CVPR'19)} (2019), pp.~3050--3058.

\bibitem[WC07]{Wang2007Optimized}
\textsc{Wang J., Cohen M.~F.}:
\newblock Optimized color sampling for robust matting.
\newblock In \emph{Proceedings of the IEEE/CVF Conference on Computer Vision
  and Pattern Recognition (CVPR'07)} (2007), pp.~1--8.

\bibitem[XGD{\etalchar{*}}17]{Xie2017Aggregated}
\textsc{{Xie} S., {Girshick} R., {Dollár} P., {Tu} Z., {He} K.}:
\newblock Aggregated residual transformations for deep neural networks.
\newblock In \emph{Proceedings of the IEEE/CVF Conference on Computer Vision
  and Pattern Recognition (CVPR'17)} (2017), pp.~5987--5995.

\bibitem[XPCH17]{Xu2017Deep}
\textsc{Xu N., Price B., Cohen S., Huang T.}:
\newblock Deep image matting.
\newblock In \emph{Proceedings of the IEEE/CVF Conference on Computer Vision
  and Pattern Recognition (CVPR'17)} (2017), pp.~311--320.

\bibitem[YCSS01]{Chuang2003A}
\textsc{{Yung-Yu Chuang}, {Curless} B., {Salesin} D.~H., {Szeliski} R.}:
\newblock A bayesian approach to digital matting.
\newblock In \emph{Proceedings of the IEEE/CVF Conference on Computer Vision
  and Pattern Recognition (CVPR'01)} (2001), pp.~II--II.

\bibitem[YXC{\etalchar{*}}18]{Yang2018Active}
\textsc{Yang X., Xu K., Chen S., He S., Yin B.~Y., Lau R.}:
\newblock Active matting.
\newblock In \emph{Proceedings of the International Conference on Neural
  Information Processing Systems (NeurIPS'18)} (2018), pp.~4590--4600.

\bibitem[ZBSS04]{Zhou2004Image}
\textsc{{Zhou Wang}, {Bovik} A.~C., {Sheikh} H.~R., {Simoncelli} E.~P.}:
\newblock Image quality assessment: from error visibility to structural
  similarity.
\newblock \emph{IEEE Transactions on Image Processing 13}, 4 (2004), 600--612.

\bibitem[ZGF{\etalchar{*}}19]{Zhang2019CVPR}
\textsc{{Zhang} Y., {Gong} L., {Fan} L., {Ren} P., {Huang} Q., {Bao} H., {Xu}
  W.}:
\newblock A late fusion cnn for digital matting.
\newblock In \emph{Proceedings of the IEEE/CVF Conference on Computer Vision
  and Pattern Recognition (CVPR'19)} (2019), pp.~7461--7470.

\bibitem[ZK09]{zheng2009learning}
\textsc{Zheng Y., Kambhamettu C.}:
\newblock Learning based digital matting.
\newblock In \emph{Proceedings of the International Conference on Computer
  Vision (ICCV'09)} (2009), pp.~889--896.

\end{thebibliography}
	
	
	
\end{document}